\documentclass{article}

\usepackage{microtype}
\usepackage{graphicx}
\usepackage{subcaption}
\usepackage{booktabs} 
\usepackage{makecell}

\usepackage{hyperref}
\usepackage{listings}
\usepackage{xcolor}
\usepackage{enumitem}
\usepackage{makecell}
\usepackage[preprint]{sty/icml2026}

\usepackage{amsmath}
\usepackage{amssymb}
\usepackage{mathtools}
\usepackage{amsthm}
\usepackage{multirow}
\usepackage[table]{xcolor}
\usepackage{enumitem}
\usepackage{bm}

\usepackage[capitalize,noabbrev]{sty/cleveref}

\theoremstyle{plain}

\theoremstyle{definition}

\theoremstyle{remark}

\usepackage[disable,textsize=tiny]{sty/todonotes}
\newif\ifnote
\notefalse

\icmltitlerunning{Semantics-Aware Generative Latent Data Augmentation for Learning in Low-Resource Domains}

\begin{document}

\twocolumn[
  \icmltitle{Semantics-Aware Generative Latent Data Augmentation \\for Learning in Low-Resource Domains} 



  \icmlsetsymbol{equal}{*}

  \begin{icmlauthorlist}
    \icmlauthor{Jaesung Bae}{yyy}
    \icmlauthor{Minje Kim}{yyy}
  \end{icmlauthorlist}

  \icmlaffiliation{yyy}{Siebel School of Computing and Data Science, University of Illinois Urbana-Champaign, Illinois, USA}

  \icmlcorrespondingauthor{Jaesung Bae}{jb82@illinois.edu}

  \icmlkeywords{Generative data augmentation, foundation model, latent space, feature space, speech emotion recognition, imbalanced image classification}

  \vskip 0.3in
]



\printAffiliationsAndNotice{}  

\begin{abstract}
    Despite strong performance in data-rich regimes, deep learning often underperforms in the data-scarce settings common in practice. While foundation models (FMs) trained on massive datasets demonstrate strong generalization by extracting general-purpose features, they can still suffer from scarce labeled data during downstream fine-tuning. To address this, we propose GeLDA, a semantics-aware generative latent data augmentation framework that leverages conditional diffusion models to synthesize samples in an FM-induced latent space. Because this space is low-dimensional and concentrates task-relevant information compared to the input space, GeLDA enables efficient, high-quality data generation. GeLDA conditions generation on auxiliary feature vectors that capture semantic relationships among classes or subdomains, facilitating data augmentation in low-resource domains. We validate GeLDA in two large-scale recognition tasks: (a) in zero-shot language-specific speech emotion recognition, GeLDA improves the Whisper-large baseline's unweighted average recall by 6.13\%; and (b) in long-tailed image classification, it achieves 74.7\% tail-class accuracy on ImageNet-LT, setting a new state-of-the-art result.
  \end{abstract}
\section{Introduction}
In many real-world applications, collecting labeled data is costly, and some categories are systematically underrepresented (e.g., in low-resource languages or a dataset with a long-tailed distribution). This yields \emph{label imbalance}, where a small number of classes dominate the training set and substantially degrade recognition performance. A common approach to mitigate this issue is to learn generalizable representations from large-scale unlabeled data with large model capacity, i.e., by using \textit{foundation models}~\cite{wav2vec, whisper, clip, bert, wavlm, CLAP, DINOv2}, and then fine-tune them for downstream tasks using lightweight adapters~\cite{hu2022lora, clip_adopter, chen2022adaptformer}. However, when the data for the downstream task is scarce or heavily imbalanced, adapter training can still overfit to dominant classes. Methods such as loss re-weighting~\cite{la_loss, focal_loss} or gradient adjustment~\cite{The_equalization_loss} can partially mitigate this issue, but the lack of data variability is a fundamental issue causing underrepresented classes. 

Data augmentation (DA) addresses the data scarcity issue by generating additional samples that follow the original data distribution. Since the empirical distribution is not representative for certain classes, a DA method needs to be carefully designed to improve the variability of the dataset in a legitimate way, i.e., the augmented empirical distribution to be a better approximation of the original. A typical approach is to manipulate raw data samples (e.g., images, text, or audio) directly in the input space \cite{ida_image_traditional1, ida_image_traditional2, mixup, cutmix,ida_importantaug,ida_cross_speaker_emotion_transfer, specaug}. Alternatively, advanced generative models, such as text-to-speech (TTS)~\cite{ida_keyword_spotting, ida_text_is_all_you_need, ida_speech_enhancement1, ida_tts_1, ida_tts_2} or image diffusion models~\cite{ida_image_1, ida_image_2, ida_image_3}, can synthesize more realistic samples, while they do not guarantee improved diversity that also matches the unseen data distribution. It is due to the high dimensionality and complexity of the raw data space, making modeling challenging. This makes input-space DA especially difficult for underrepresented data. Moreover, since raw data often includes details unnecessary for a downstream task, respecting such subtleties during generation is wasteful.

Alternatively, DA can operate in a well-structured feature space, which is typically low-dimensional and highly abstract. This space can be more efficient and effective than the input space, as DA can focus on the task-relevant core information rather than being distracted by creating the peripheral details. Moreover, semantic similarity between classes or domains is often better captured in feature spaces, enabling the DA to operate in the ``easier'' space and to better handle label imbalance. Previous latent space DA methods often rely on the simple statistical manipulation of existing data points, based on the strong assumption that well-learned feature spaces are approximately convex~\cite{cheung2021modals, latent_filling, data_aug_in_feature_space}. More recently, a few studies have explored generative model-based approaches~\cite{LeMDA, a_closer_look_at_feature_space_da, ldmlr}, but a research gap remains in understanding the role of generative models for DA. For example, their interplay with existing foundation models (FMs), effective conditioning methods, and how different levels of abstraction in the feature spaces affect the DA quality, are the research questions of this paper.

In this work, we propose a semantics-aware generative latent DA (GeLDA) framework as an efficient and effective alternative to input-space DA. It is characterized by the use of pretrained FMs and their fine-tuned versions to ensure DA is performed in a well-structured, task-relevant latent space. Diffusion models are adopted to generate latent samples, by being conditioned on \emph{augmented} label information to facilitate the connection between the high-resource and low-resource classes or domains. This makes GeLDA especially powerful in severely low-resource data setups.
We challenge the GeLDA framework in two realistic data-scarcity scenarios: (a) zero-shot (i.e., without any labeled data) speech emotion recognition (SER) for low-resource languages, and (b) long-tailed image classification on a highly imbalanced dataset. 
In the SER task, GeLDA improves unweighted average recall by 6.2\% with Whisper-large~\cite{whisper} as the FM, using a small diffusion model (21M parameters) trained on only 83 hours of data. On the ImageNet Long Tail (LT) dataset~\cite{imagenetlt}, GeLDA achieves 74.7\% tail-class accuracy, setting a new state-of-the-art (SOTA) result, while maintaining the middle- and head-class accuracies. These results suggest that GeLDA is applicable to a range of real-world tasks, mitigating the well-known difficulties in low-resource and label-imbalanced scenarios. Our contributions are summarized as follows:
\begin{itemize}[leftmargin=*, itemindent=0pt, , noitemsep, topsep=0pt]
\item GeLDA proposes a semantics-aware generative DA framework that synthesizes data in an FM-learned latent space, more efficiently and effectively than input-space DA.
\item We propose augmenting the label and subdomain information to capture the relationship between labels and conditioning the generative model on it. We show that the augmented side information successfully facilitates DA in low-resource domains.
\item Through ablations, we analyze how different abstraction choices in the latent space affect GeLDA.
\item We validate GeLDA on SER and ImageNet-LT tasks to show its merit in different modalities and few- and zero-shot task variations.
\end{itemize}
\section{Related Works}
\paragraph{Foundation models:}
An FM is trained on large-scale, diverse data, typically using self-supervised or weakly-supervised learning. It is designed to generalize to various downstream tasks~\cite{on_the_opportunities_and_risks}. For example, wav2vec 2.0~\cite{wav2vec},  WavLM~\cite{wavlm}, and BERT~\cite{bert} are trained with masked-prediction objectives for speech and text. More task-oriented speech FMs have also emerged, including Whisper~\cite{whisper} for automatic speech recognition (ASR) and emotion2vec~\cite{emotion2vec} for SER. Notably, Whisper’s encoder is known to transfer well to non-ASR tasks such as emotion recognition~\cite{whisper_ser_1, emobox} and audio tagging~\cite{whisper_audio_tagging}. In the vision-language domain, CLIP~\cite{clip} has been successful in learning a shared semantic space between image and text modalities, showcasing excellent zero-shot image classification performance~\cite{online_zero_shot_classification_with_clip, clip_vqa}. To the best of our knowledge, though, GeLDA is the first work that extensively studies the relationship between the FM-learned latent space and DA. We leverage their ability to capture the semantics in the feature space, where the DA is performed.

\paragraph{Input-space data augmentation:} Transformation techniques can alter data distributions, e.g., rotation, flipping, cropping, and color change in vision applications~\cite{ida_image_traditional1, ida_image_traditional2}, along with mixing examples and labels~\cite{mixup, cutmix}. In audio, adding noise~\cite{ida_importantaug}, pitch modulation~\cite{ida_cross_speaker_emotion_transfer}, and masking~\cite{specaug} are widely used. These methods increase diversity but are not guaranteed to effectively recover the unseen parts of the data distribution, limiting the generalizability of the learned models. Recently, generative models have been adopted for input-space DA. In speech, generative models for voice conversion and TTS can synthesize new utterances with varying conditions, such as speaker identity and text content, benefiting ASR~\cite{ida_text_is_all_you_need}, keyword spotting~\cite{ida_keyword_spotting}, TTS~\cite{ida_tts_1, ida_tts_2}, speech enhancement~\cite{ida_speech_enhancement1, ida_speech_enhamcement_2}, and emotion recognition~\cite{ida_emotion_1}. In vision, diffusion-based image generation has also gained popularity for DA~\cite{ida_image_1, ida_image_2, ida_image_3}. Despite these successes, generative input-space DA often requires large training datasets and models, making it impractical for low-resource domains.

\paragraph{Latent-space data augmentation: }
An intuitive way to perform DA in the latent space is to use a linear transformation or add a random perturbation to the existing latent vectors \cite{a_closer_look_at_feature_space_da, data_aug_via_latent_space, data_aug_in_feature_space, latent_filling, cheung2021modals}, assuming the data distribution in the latent space is smooth. To further ensure a convex, simple latent space, adversarial loss functions can be used \cite{data_aug_via_latent_space, cheung2021modals}. Based on the simplicity in the latent space, prior studies used mixing or recombining latent representations across samples \cite{feature_space_transfer_for_da, feature_space_augmenation_for_lt_data}. Other works utilize consistency loss to improve generation quality for image recognition~\cite{LARE} and TTS \cite{latent_filling}. Recently, LCReg leverages shared features between high- and low-resource classes~\cite{LCReg}, but their method explicitly computes and stores these shared features, and then performs statistics-based sampling for DA. Generative model-based approaches, including variational autoencoders (VAEs) \cite{LeMDA,a_closer_look_at_feature_space_da} and diffusion models \cite{ldmlr}, have also been studied. However, they remain less explored than input-space methods. To the best of our knowledge, no prior work has investigated how the choice of latent space properties affects the effectiveness of DA. To this end, the proposed GeLDA framework leverages FM-learned and fine-tuned latent representations as well as versatile label and subdomain conditioning via classifier-free guidance (CFG) in diffusion models. It enables latent-space DA in challenging, data-scarce settings.
\section{Problem Definition}
\label{sec:problem_definition}
Our target applications are recognition tasks under data imbalance. We first describe the FM-based recognition framework and then formalize the data imbalance scenarios.

\begin{figure}[t]
     \centering
     \begin{subfigure}[b]{0.20\columnwidth}
         \centering
         \includegraphics[width=\textwidth]{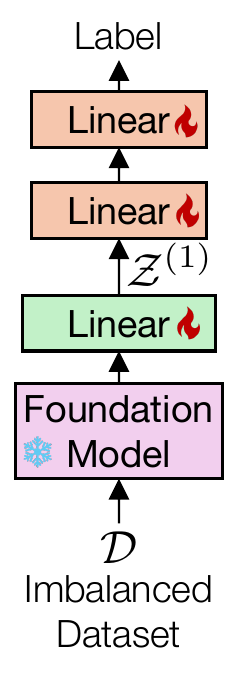}\vspace{-0.1in}
         \caption{}
         \label{fig:model_pretrain}
     \end{subfigure}
     \quad
     \begin{subfigure}[b]{0.214\columnwidth}
         \centering
         \includegraphics[width=\textwidth]{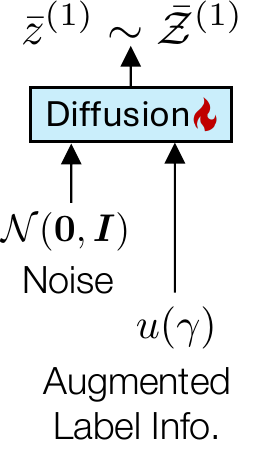}\vspace{-0.1in}
         \caption{}
         \label{fig:model_diffusion}
     \end{subfigure}
     \quad
     \begin{subfigure}[b]{0.373\columnwidth}
         \centering
         \includegraphics[width=\textwidth]{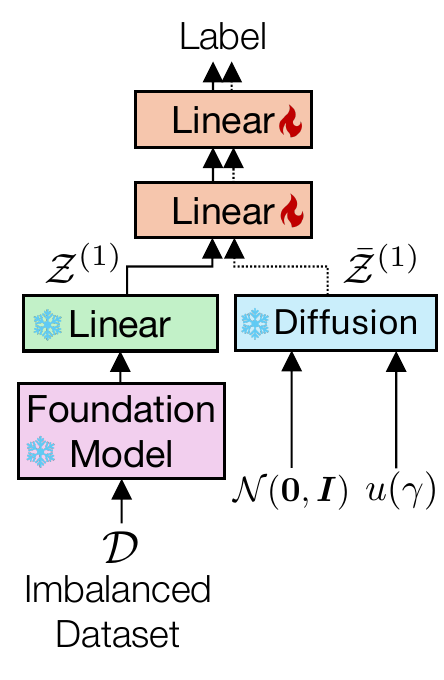}\vspace{-0.1in}
         \caption{}
         \label{fig:model_finetune}
     \end{subfigure}
        \caption{
        Example of the GeLDA framework with an $L=3$ task adapter in $\mathcal{Z}^{(1)}$. (a) Repurposing the FM into a downstream recognition model via adapter layers. (b) Training a latent diffusion model to synthesize features in $\mathcal{Z}^{(1)}$, conditioned on augmented label information $u(\gamma)$ to transfer cues to the low-resource class $\gamma$ from related high-resource classes. (c) Fine-tuning the model using synthesized and ground-truth samples in $\mathcal{Z}^{(1)}$.}
        \vspace{-0.4cm}
        \label{fig:model}
\end{figure}

\paragraph{Recognition with foundation models:}
A recognition model maps an input $x \in \mathcal{X}$ to a label $y \in \mathcal{Y} = \{c|1, 2, \ldots, C\}$. Modern recognition systems commonly employ a pretrained FM $\mathcal{F} : \mathbb{R}^D \rightarrow \mathbb{R}^M$ to transform $x$ into a latent representation $z\in\mathcal{Z}$ (typically $M < D$). While this latent space $\mathcal{Z}$ is generally well structured with abstraction, it may still contain information irrelevant to a specific downstream task. To adapt $\mathcal{Z}$ further to the downstream task, we train a lightweight task adapter $\mathcal{H}$ on top of $\mathcal{F}$, often implemented as $L$ stacked linear layers:
\begin{equation}
\hat{y} = \text{Softmax} \circ \mathcal{H}^{(L)} \circ \cdots \circ \mathcal{H}^{(1)} \circ \mathcal{F}(x),
\end{equation}
where each layer $\mathcal{H}^{(l)}$ induces a progressively more task-specific latent space $\mathcal{Z}^{(l)}$, with $\mathcal{Z}^{(0)}$ being the FM's output. Fig. \ref{fig:model_pretrain} illustrates the common recognition framework. 

\paragraph{Imbalanced datasets:}
Training data often suffers from \emph{label imbalance}. Let $\mathcal{D} = \bigsqcup_{c=1}^C \mathcal{D}_c$ be the training set with $\mathcal{D}_c = \{(x_i, y_i)| y_i=c\}$. The dataset is imbalanced when the sample distribution over $\mathcal{Y}$ is skewed, such that some classes are substantially underrepresented, e.g., there exists a class-specific subset $\mathcal{D}_\gamma$ with a very low cardinality. In such a case, \textit{few-shot} learning aims to maximize the use of samples in $\mathcal{D}_\gamma$, while \textit{zero-shot} corresponds to $|\mathcal{D}_\gamma|=0$. The imbalance is detrimental when the empirical distribution $\hat{p}(x|y=\gamma)$ induced from $\mathcal{D}_\gamma$ poorly approximates $p(x|y=\gamma)$. In Sec. \ref{sec:exp_cv}, we show decreasing performance of the image classification models in the \textit{tail} classes. As a remedy, GeLDA relates a low-resource class to its semantically related high-resource classes during the synthesis.

\paragraph{Subdomains with severer label imbalance:}
If the task can be divided into subdomains (e.g., language-agnostic SER broken down into language-specific SER), label imbalance can get severe. With a shared label space $\mathcal{Y}$, we write $\mathcal{D} = \bigcup_{k=1}^K \bigsqcup_{c=1}^C \mathcal{D}_c^{(k)}$, where $\mathcal{D}_c^{(k)}$ denotes class-$c$ samples from subdomain $k$. Even if class $c$ has enough samples across all the subdomains, it may be scarce in the target subdomain $\kappa$, i.e., $|\mathcal{D}_c| \gg  |\mathcal{D}_c^{(\kappa)}|$ (e.g., missing \textsc{surprise} samples in the German corpus). We assume DA can leverage the similarity between a low-resource class in $\kappa$ and a related high-resource subdomain $k'$ (e.g., \textsc{surprise} in the English set). We thus propose conditioning the generative DA model on discriminative subdomain information (e.g., the target language) to implicitly guide generation via related subdomains that have more data.

\begin{figure*}[ht]
     \centering
     \begin{subfigure}[b]{0.184\textwidth}
         \centering
         \includegraphics[width=\textwidth]{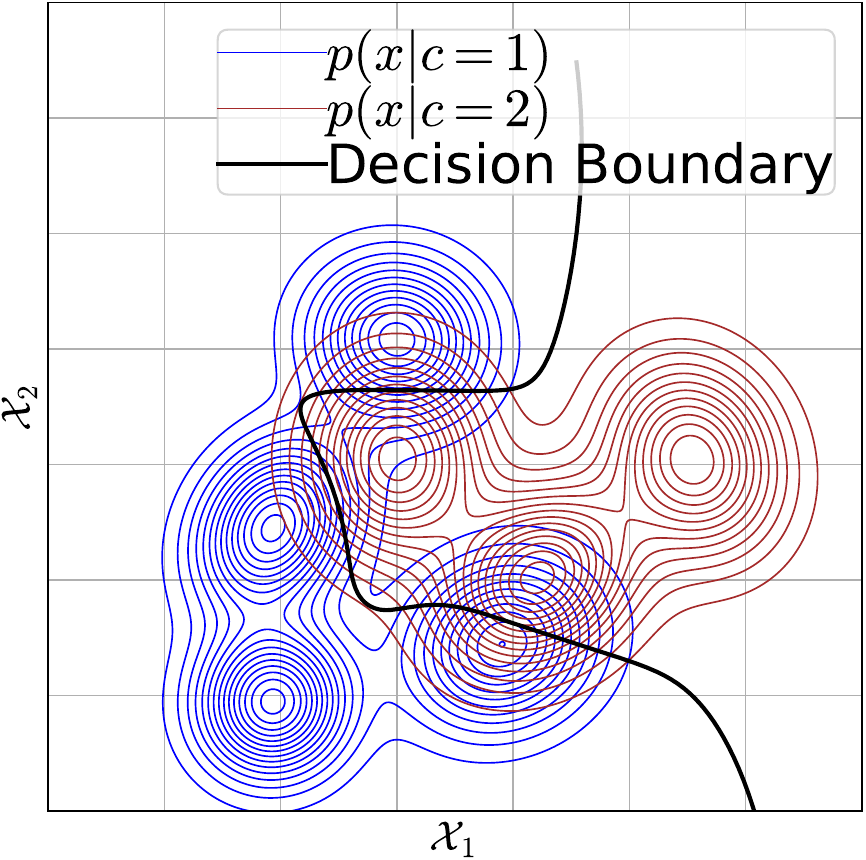}\vspace{-0.05in}
         \caption{}
         \label{fig:input_dist}
     \end{subfigure}
     \qquad
     \begin{subfigure}[b]{0.184\textwidth}
         \centering
         \includegraphics[width=\textwidth]{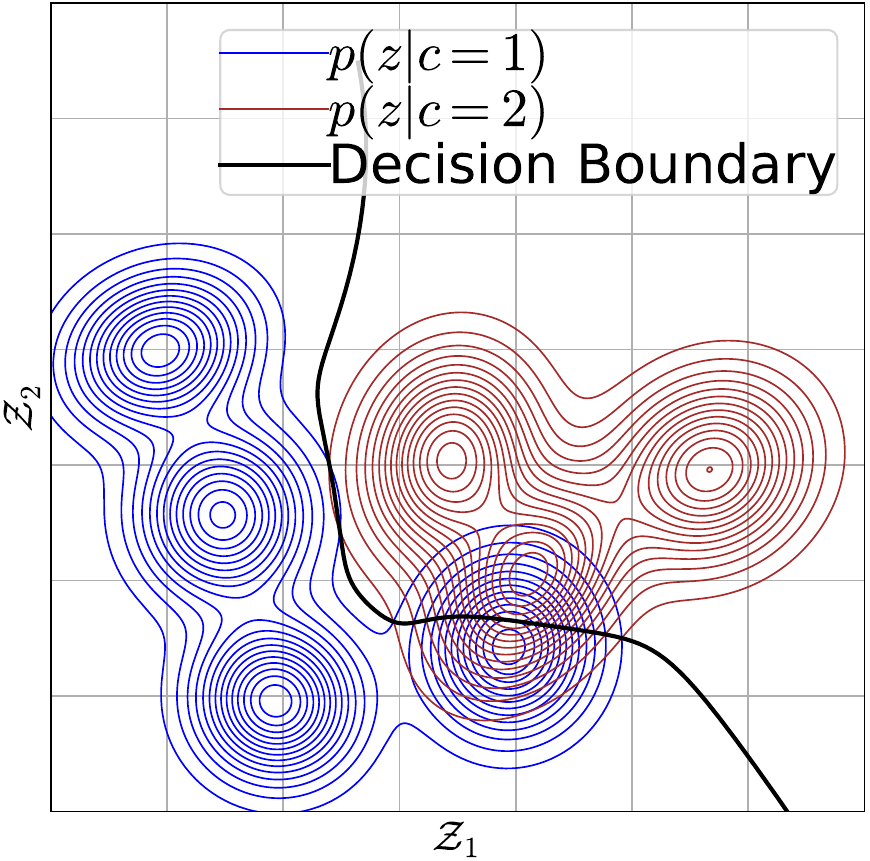}\vspace{-0.05in}
         \caption{}
         \label{fig:feature_dist}
     \end{subfigure}
     \qquad
     \begin{subfigure}[b]{0.176\textwidth}
         \centering
         \includegraphics[width=\textwidth]{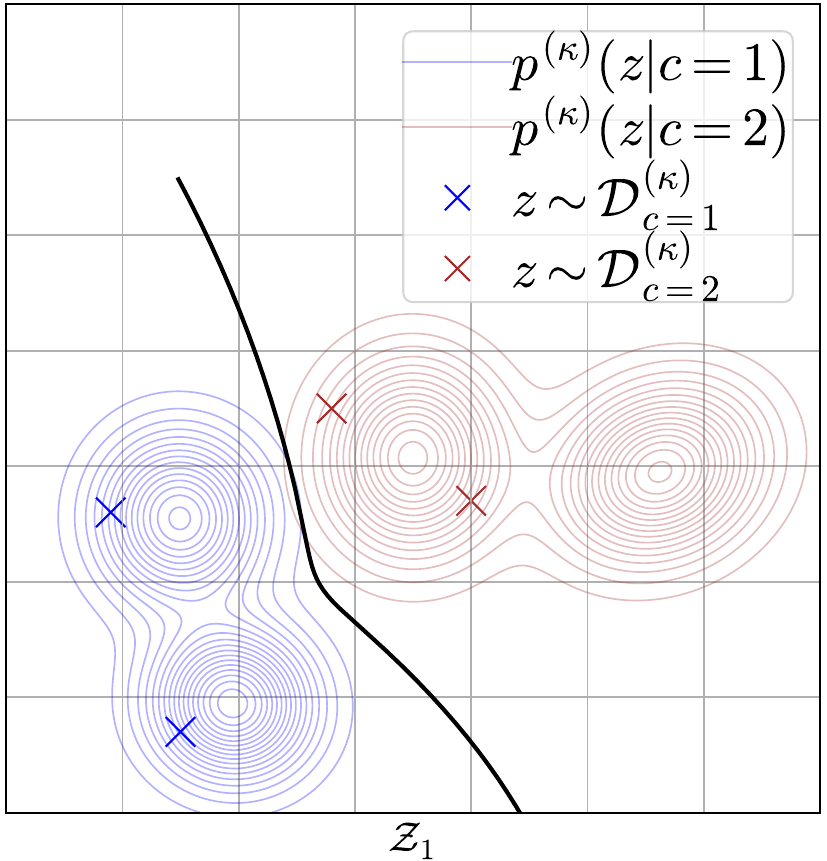}\vspace{-0.05in}
         \caption{}
         \label{fig:low-resource-samples}
     \end{subfigure}
     \qquad
     \begin{subfigure}[b]{0.176\textwidth}
         \centering
         \includegraphics[width=\textwidth]{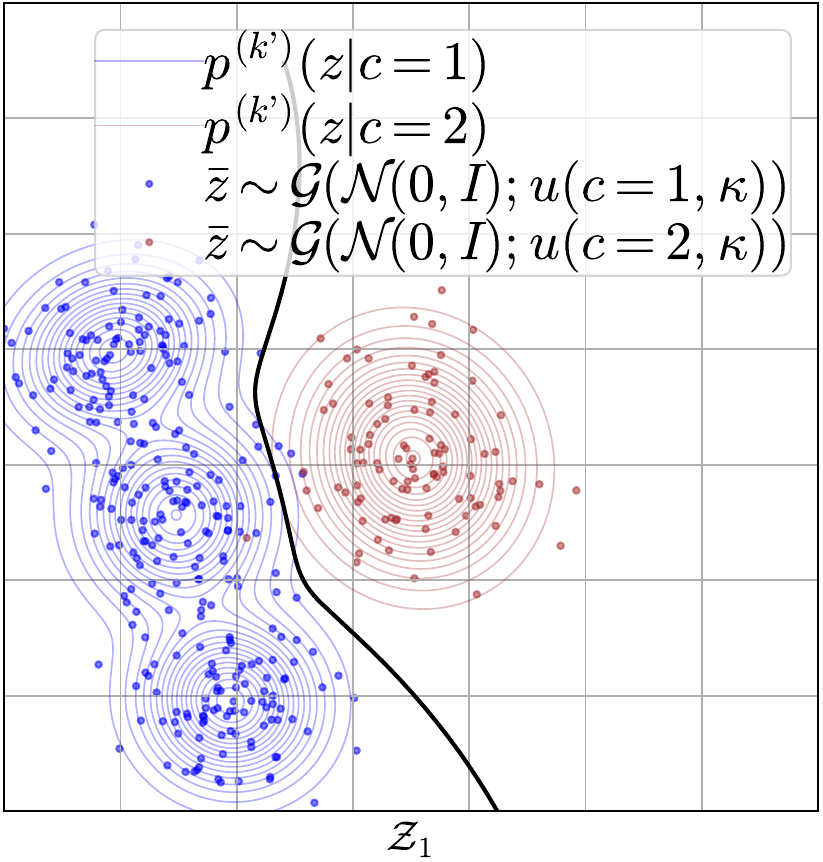}\vspace{-0.05in}
         \caption{}
         \label{fig:augmented}
     \end{subfigure}
        \caption{Illustration of the proposed GeLDA in the feature space. (a) Original data distribution $\mathcal{X}$ with a complex decision boundary. (b) More separable class distributions in the learned feature space $\mathcal{Z}$. (c) Few samples from the low-resource subset $\mathcal{D}^{(\kappa)}$. Contours are from the unknown ground-truth distribution of subdomain $\kappa$. (d) GeLDA's feature-space data augmentation results $\bar{z}$, whose sample distribution resemble the distribution of $k'$. $\kappa$ and $k'$ are semantically similar subdomains.}
        \vspace{-0.3cm}
        \label{fig:illustration}
\end{figure*}

\section{Design Principles of GeLDA}

\subsection{Why Data Augmentation in Latent Space?}
\label{subsec:latent_da_motivation}
The goal of DA for a low-resource class $\gamma$ is to reconstruct its class-conditional distribution $p(x|y=\gamma)$. However, na\"ive strategies (e.g., sample interpolation) are often suboptimal as the raw data distribution $p(x|y)$ is heavily affected by task-irrelevant nuances, such as sensor noise, visual style, speaker identity, or acquisition conditions, that vary independently of $y$. As a result, DA may fail to preserve semantic relatedness between the generated samples and the label.

To address this, GeLDA augments data in a latent embedding space $\mathcal{Z}$ extracted from an FM and task adapter. This space is more structured than $\mathcal{X}$, as it abstracts away task-irrelevant low-level variations while preserving task-relevant semantic information. Formally, 

\noindent\textbf{Proposition 1.}
Given a pair of samples $(x_i, x_j)$ and their task-relevant semantic similarity $\mathcal{E}_{\text{semantic}}(x_i, x_j)$, there exists a function $\mathcal{F}: \mathbb{R}^D \rightarrow \mathbb{R}^M$ with $M < D$, whose output $z_i \leftarrow \mathcal{F}(x_i)$ and $z_j \leftarrow \mathcal{F}(x_j)$ provide a better preservation of the original semantic similarity in the form of their Euclidean distance than the raw data do. Formally:
\begin{align}
\nonumber &\mathbb{E}_{x_i, x_j \sim p(x_{\text{data}})} \bigl| \| z_i - z_j \|_2^2 - \mathcal{E}_{\text{semantic}}(x_i, x_j) \bigr| \\
&\quad < \mathbb{E}_{x_i, x_j \sim p(x_{\text{data}})} \bigl| \| x_i - x_j \|_2^2 - \mathcal{E}_{\text{semantic}}(x_i, x_j) \bigr|.
\end{align}
where $p(x_{\text{data}})$ denotes the data distribution.

Figs.~\ref{fig:input_dist} and \ref{fig:feature_dist} contrast the input and feature spaces in terms of their relationship with the complexity of the decision boundary. While this proposition is not proven\footnote{Our empirical analysis measures the average intra-class Euclidean distance between  ImageNet-LT test samples; CLIP embeddings yield 37.8\% smaller distances than raw images.}, it is strongly supported by empirical studies in the literature. Prior work shows that supervised feature transformations progressively increase class separability, abstraction, and task-specificity in deeper layers~\cite{pmlr-v202-rangamani23a, Prevalence_of_neural_collapse}. Similarly, latent diffusion models operate in a structured latent space $\mathcal{Z}$ rather than $\mathcal{X}$ to generate high-level semantics first, improving stability and semantic fidelity \cite{ldm, audio_ldm}.

\subsection{Class-Conditioned GeLDA}
Inspired by this, we train the generative model directly in the latent space $\mathcal{Z}^{(l)}$, reducing the burden of modeling low-level variability and making it easier to exploit semantic relatedness between the low-resource class $\gamma$ and its relevant high-resource classes. Our key assumption is that the semantic similarity between labels is learned in an embedding space $u(c)$. Hence, $u(\gamma)$ can be seen as an \emph{augmented} version of the label information $\gamma$, overcoming its underrepresentedness via its semantic relationship with other classes. The augmented reference conditions are generated as follows:
\begin{equation}
z = \mathcal{G}(\epsilon; u(\gamma)), \quad \epsilon \sim \mathcal{N}(0, \mathbf{I})
\label{eq:gen-process}
\end{equation}
where $\mathcal{G}$ denotes a conditional generator.

For example, $u(c)$ can be a multimodal FM (e.g., CLIP~\cite{clip}) that learns the semantic relationship between images and text in a shared embedding space, with text labels also contextualized. Hence, even if $\gamma$ is unseen during image recognition training, $u(\gamma)$ can still condition generation via its semantic similarity to other classes.

\subsection{Subdomain and class-conditioned GeLDA}
\label{sec:principles-conditioning}
In order to consider subdomain specificity, our key departure from class-conditional generation is to additionally condition on a \emph{subdomain reference} $\kappa$ alongside the low-resource label condition $\gamma$. This extra conditioning provides structural cues about the intended latent distribution, enabling more controllable data augmentation for $\kappa$. To this end, we redefine the label augmentation process in Eq. \eqref{eq:gen-process} by putting $\kappa$ as additional information: $u(\gamma, \kappa)$. For a language-specific subdomain of the SER problem, for example, $u(\gamma, \kappa)$ computes an embedding vector that informs the generation process of the target emotion $\gamma$ and the language subdomain $\kappa$ altogether. In doing so, we postulate that $u(\gamma, \kappa)$ can be compared with $u(\gamma, k')$, an adjacent subdomain's conditioning vector due to the cross-language similarity between $\kappa$ and $k'$. Figs.~\ref{fig:low-resource-samples} and \ref{fig:augmented} illustrate this process: In \ref{fig:low-resource-samples}, with only two samples per class, i.e., $|\mathcal{D}^{(\kappa)}_{c=1}|=|\mathcal{D}^{(\kappa)}_{c=2}|=2$, the empirical distribution $\hat{p}^{(\kappa)}(x|y)$ must be unstable compared to the ground-truth ${p}^{(\kappa)}(x|y)$ (the contours). GeLDA conditions on both $\gamma$ and $\kappa$ to synthesize samples that follow the adjacent subdomain-specific distributions $p^{(k')}(x|y)$. Although there is still some discrepancy, the learned decision boundary in \ref{fig:augmented} is comparable to the ground-truth (GT) in \ref{fig:low-resource-samples}.

\subsection{On the choice of a proper feature layer}
If the task adapter consists of multiple layers, GeLDA can operate in different embedding spaces $\mathcal{Z}^{(l)}$, which involves an inherent trade-off. If $l$ is too close to the final layer $L$, the layers before $l$ (i.e., the first linear layer in Fig.~\ref{fig:model}) cannot be updated by the generated data, reducing the model’s flexibility and the benefit of DA during fine-tuning. Moreover, as $l$ increases, class representations in $\mathcal{Z}^{(l)}$ tend to become increasingly isotropic and convex, and may collapse into a simplex structure~\cite{pmlr-v202-rangamani23a, Prevalence_of_neural_collapse}, thereby diminishing the feature space’s diversity and the effectiveness of GeLDA. Conversely, applying GeLDA at very early layers or to feature spaces learned by a weak FM yields noisy, unstable representations, making it difficult for the generative model to capture semantically meaningful variations. Thus, selecting an appropriate layer $l$ for the given task is essential for GeLDA. We investigate this trade-off in Sec. \ref{sec:exp_ser}.
\section{Training Methods for GeLDA}
\label{sec:training_methods}
Fig.~\ref{fig:model} illustrates the overall pipeline of GeLDA. In this section, we consider the class-imbalance setting characterized by a low-resource class $\mathcal{D}_\gamma$, while the same procedure can be applied to the subdomain imbalance case, $\mathcal{D}_\gamma^{(\kappa)}$, too.

\paragraph{Stage 1: Generic recognition model training:}
Fig. \ref{fig:model_pretrain} depicts Stage 1. Throughout all stages, the FM is kept \emph{fully frozen}, so that GeLDA can work with any black-box FMs. We instead train a lightweight task adapter $\mathcal{H}$ on top of the FM representations. For speech FMs, we apply temporal average pooling to aggregate frame-level representations into an utterance-level feature. For images, CLIP's class token is extracted as features. In this stage, the model is trained on the full, imbalanced dataset $\mathcal{D}$, which can lead to suboptimal performance on the low-resource class $\gamma$.

\paragraph{Stage 2: Training a diffusion model for GeLDA:}
To augment training data for the low-resource class $\gamma$, we train a diffusion model to synthesize feature vectors in $\gamma$. Unlike latent diffusion models~\cite{ldm} learned to work with an autoencoder, our diffusion model generates in a task-specific feature space $\mathcal{Z}^{(l)}$.
To incorporate conditioning as in Eq.~\ref{eq:gen-process}, we adopt CFG~\cite{cfg}.
After training, we generate augmented latent vectors $\bar{z}^{(l)}$ from the diffusion model and construct the augmented set $\bar{\mathcal{D}}_{\gamma}$. Fig. \ref{fig:model_diffusion} describes the process. 

\paragraph{Stage 3: Fine-tuning with augmented latent vectors:}
As shown in Fig. \ref{fig:model_finetune}, we fine-tune the recognition model on the original low-resource dataset $\mathcal{D}_{\gamma}$ by passing it through the FM and adapter layers. As for the augmented feature set $\bar{\mathcal{D}}_{\gamma}$, they are fed directly to $\mathcal{H}^{(l+1)}$, as they are already in the $\mathcal{Z}^{(l)}$ space.
In doing so, we freeze the FM and all layers up to the target latent space $\leq l$, and fine-tune layers beyond it (i.e., $> l$).
After this stage, the resulting model specializes from a generic recognition model into a class-specific recognition model for the low-resource class $\gamma$.
\section{Experiments}
We evaluate the effectiveness and generalization of GeLDA on two challenging tasks from different domains: zero-shot SER and imbalanced image classification. 

\subsection{Zero-shot Speech Emotion Recognition}
\label{sec:exp_ser}
\subsubsection{Task definition and dataset}
In multilingual SER, the output space $\mathcal{Y}$ represents emotion classes and is shared across subdomains $k$, where each subdomain corresponds to a language. We define a low-resource language such that the $\kappa$-th subset $\mathcal{D}^{(\kappa)}$ contains no labeled samples except for a few \textsc{neutral} utterances, i.e., $\mathcal{Y}^{(\kappa)}=\{\textsc{neutral}\}$, which yields a zero-shot setting.

We collect a multilingual speech emotion dataset following EmoBox~\cite{emobox}; its language and emotion distribution is summarized in Table \ref{tab:ser_dataset}. The dataset exhibits a heavy but common imbalance issue across emotions and languages (e.g., $>50$ hours of English vs. $0.21$ hours of German, missing \textsc{surprise} for German and Spanish). To simulate a zero-shot recognition environment, we select a target low-resource language $\kappa$ and we remove all its samples except those in the neutral class, repurposing $\mathcal{D}^{(\kappa)}$ into $\mathcal{D}^{(\kappa)}_{\textsc{neu}}=\{(x_i, y_i)\mid y_i=\textsc{neutral}\}$. We use 3 folds and report average results. The total number of speech samples is 84$,$283, and the average duration is 3.52 seconds.

\subsubsection{Experiment setup}
We add task adapter layers to three pretrained FMs, \textsc{WavLM-large}~\cite{wavlm}, \textsc{emotion2vec-base}~\cite{emotion2vec}, and \textsc{Whisper-large}~\cite{whisper}, to build SER models. We use task adapters with $L=3$ linear layers with ReLU activations, and choose one of the task adapter layer output spaces to train the latent diffusion model. For the generic model pretraining stage, we merge two partitions: all language subsets except for the target, $\mathcal{D}^{\setminus \kappa}$, and the neutral utterances from the target, $\mathcal{D}^{(\kappa)}_{\textsc{neu}}$. For the second stage, we condition the diffusion model on emotion class $c$ via a learnable lookup table. To condition the model on subdomain $\kappa$, we use a randomly sampled feature embedding $z^{(l)}\in{\mathcal{D}}^{(\kappa)}$ when operating in $\mathcal{Z}^{(l)}$ space. Ablation study on this choice will be described in Table~\ref{tab:ser-ablation_cond}. The diffusion model then generates new feature vectors $\bar{z}^{(l)}$ for each emotion $c$ and the target language $\kappa$, forming a synthetic dataset $\bar{\mathcal{D}}^{(\kappa)}$. Finally, in the fine-tuning stage, the task adapter layers later than $l$ are updated using neutral GT examples $\mathcal{D}^{(\kappa)}_{\textsc{neu}}$ as well as the synthesized examples $\bar{\mathcal{D}}^{(\kappa)}$ in $\mathcal{Z}^{(l)}$. Additional implementation details are in Appendix~\ref{apdx:implementation_details}. 

Due to the strong class imbalance in emotional speech datasets, we evaluate SER models using unweighted average (UA) recall and weighted average (WA) recall~\cite{interspeech_emotion_challenge}. UA computes recall per emotion and then averages across classes, preventing the majority classes from dominating the score. We also report Macro-F1, which averages the per-class F1 scores.

\subsubsection{Comparison models}
For comparison, we construct two baseline models using the original data (i.e., without augmentation): (1) a \textsc{pretrained} SER model trained on all other languages than $\kappa$, $\mathcal{D}^{\setminus \kappa}$, serving as a lower bound, and (2) a model trained on the full multilingual dataset $\mathcal{D}$, serving as an \textsc{oracle} upper bound. We omit the results from a language-specific model trained or fine-tuned on the GT $\mathcal{D}^{(\kappa)}_{\textsc{neu}}$, as it is suboptimal due to the lack of labeled data: the multilingual version is better as it leverages language-agnostic features. 

As input-space DA baselines, we use SOTA speech translation systems, \textsc{SeamlessM4T-Large} and \textsc{Seamless Expressive}~\cite{seamless}, to synthesize speech in $\mathcal{X}$, where \textsc{Seamless Expressive} is a variant of \textsc{Seamless M4T} focusing on speech expressiveness. These systems translate a prompt utterance in the source language into the target language while imitating the source's speaking style and speaker identity. For any given target language $\kappa$, we use randomly chosen English utterances with varying emotion categories as sources.

These input-space generative models require $>20$K hours of training data and 450M parameters, making them difficult to generalize to low-resource target languages. \textsc{Seamless Expressive} supports only French, German, Italian, and Spanish in our dataset, so we report the average results over these four target languages. In contrast, GeLDA requires only 83 hours of data and 21M parameters for its diffusion-based generation, showing much greater efficiency and generalization capability, beyond the four targets this section covers (Appendix \ref{apdx:ser-per-lang}). Additional efficiency and performance comparisons are provided in Figs.~\ref{fig:diffusion_size} and Appendix~\ref{apdx:efficiency_comparison}.

\begin{table*}[ht]
\caption{Average test results for zero-shot SER experiments on four target languages: French, German, Spanish, Italian. Aug. Space indicates the different locations where DAs are applied. For the augmentation, we generate 200 samples per emotion. ``UA w/o Neutral'' indicates the UA over all emotions except \textsc{neutral}. $\mathcal{D}^{\setminus \kappa}$, $\mathcal{D}^{(\kappa)}_{\textsc{neu}}$, and $\bar{\mathcal{D}}^{(\kappa)}$, indicate the original dataset \textit{without} target language $\kappa$, the target language $\kappa$'s subset with only neutral emotion samples, and the augmented dataset for $\kappa$, respectively.}
    \label{tab:result-main}
    \centering
    \footnotesize
    \resizebox{\textwidth}{!}{
    \begin{tabular}{llcccrrrrrrr|cccc}
    \toprule
    \multirow{2}{*}{\textbf{FM}} & 
        \multirow{2}{*}{\textbf{Model}} & \textbf{Pretrain} & \textbf{Fine-tune} & \multirow{2}{*}{\textbf{Aug. Space}} & \multicolumn{7}{c}{\textbf{Per emotion recall (\%)}} & 
        \textbf{UA} & \textbf{WA} & \textbf{Macro-F1} & \textbf{UA w/o}\\ 
        \cline{6-12}
        ~ & ~ & \textbf{Dataset} & \textbf{Dataset} & ~ &\textbf{Angry} & \textbf{Disgust} & \textbf{Fear} & \textbf{Happy} & \textbf{Neutral} & \textbf{Sad} & \multicolumn{1}{c}{\textbf{Surprise}} & \textbf{(\%)} & \textbf{(\%)} & \textbf{(\%)} & \textbf{ Neutral (\%)}\\
        \midrule

\multirow{7}{*}{WavLM-large} & 
Pretrained & $\mathcal{D}^{\setminus \kappa}$ & -- & -- & 67.5 & 46.0 & 34.0 & 54.5 & 71.8 & 50.0 & 31.5 & 52.4 & 53.2 & 52.5 & 48.9\\ 
~ & Oracle & $\mathcal{D}$ & -- & -- & 79.2 & 67.8 & 63.5 & 68.0 & 76.8 & 72.5 & 61.0 & 70.8 & 71.0 & 71.1 & 69.6\\ 
\cline{2-16}
~ & Seamless M4T & \multirow{2}{*}{$\mathcal{D}^{\setminus \kappa} \bigcup\mathcal{D}^{(\kappa)}_{\textsc{neu}}$} & \multirow{2}{*}{$\mathcal{D}^{(\kappa)}_{\textsc{neu}} \bigcup\bar{\mathcal{D}}^{(\kappa)}$} & \multirow{2}{*}{$\mathcal{X}$} & 32.5 & 15.5 & 10.8 & 26.0 & 96.0 & 33.0 & 17.5 & 34.7 & 34.9 & 30.7 & 23.5\\ 
~ & Seamless Expressive & ~ & ~ & ~ & 45.8 & 20.8 & 22.0 & 34.8 & 91.0 & 46.0 & 33.0 & 42.8 & 42.9 & 40.6 & 34.0\\ 
\cline{2-16}
~ & ~ & \multirow{3}{*}{$\mathcal{D}^{\setminus \kappa} \bigcup\mathcal{D}^{(\kappa)}_{\textsc{neu}}$} & \multirow{3}{*}{$\mathcal{D}^{(\kappa)}_{\textsc{neu}} \bigcup\bar{\mathcal{D}}^{(\kappa)}$} &  $\mathcal{Z}^{(0)}$ & 68.5 & 47.0 & 42.7 & 43.0 & 64.7 & 51.7 & 47.5 & 52.6 & 53.3 & 52.1 & 50.6\\ 
~ & GeLDA (Ours) & ~ & ~ & $\mathcal{Z}^{(1)}$ & 65.0 & 43.7 & 47.7 & 44.7 & 53.2 & 47.0 & 42.5 & 49.6 & 51.0 & 48.8 & 49.2\\ 
~ & ~ & ~ & ~ & $\mathcal{Z}^{(2)}$ & 66.7 & 46.5 & 34.2 & 52.5 & 59.5 & 55.0 & 45.0 & 51.8 & 52.1 & 50.6 & 50.7\\ 
\hline

\multirow{7}{*}{emotion2vec-base} & Pretrained & $\mathcal{D}^{\setminus \kappa}$ & -- & -- & 58.2 & 40.7 & 27.5 & 33.2 & 52.7 & 53.7 & 23.5 & 42.8 & 43.0 & 42.7 & 41.0 \\
~ & Oracle & $\mathcal{D}$ & -- & -- & 66.2 & 57.2 & 50.0 & 49.2 & 69.5 & 65.0 & 49.5 & 58.8 & 59.0 & 58.9 & 56.9 \\
\cline{2-16}
~ & Seamless M4T & \multirow{2}{*}{$\mathcal{D}^{\setminus \kappa} \bigcup\mathcal{D}^{(\kappa)}_{\textsc{neu}}$} & \multirow{2}{*}{$\mathcal{D}^{(\kappa)}_{\textsc{neu}} \bigcup\bar{\mathcal{D}}^{(\kappa)}$} & \multirow{2}{*}{$\mathcal{X}$} & 18.2 & 12.5 & 9.2 & 6.7 & 94.0 & 20.2 & 6.5 & 25.5 & 25.3 & 19.5 & 13.1\\
~ & Seamless Expressive & ~ & ~ & ~ & 32.7 & 16.7 & 24.5 & 17.5 & 92.2 & 41.0 & 13.5 & 36.0 & 35.7 & 32.8& 25.7 \\
\cline{2-16}
~ & ~ & \multirow{3}{*}{$\mathcal{D}^{\setminus \kappa} \bigcup\mathcal{D}^{(\kappa)}_{\textsc{neu}}$} & \multirow{3}{*}{$\mathcal{D}^{(\kappa)}_{\textsc{neu}} \bigcup\bar{\mathcal{D}}^{(\kappa)}$} & $\mathcal{Z}^{(0)}$ & 44.2 & 47.2 & 37.2 & 31.0 & 78.2 & 44.2 & 29.0 & 46.0 & 45.0 & 44.4 & 40.3\\
~ & GeLDA (Ours) & ~ & ~ & $\mathcal{Z}^{(1)}$ & 47.2 & 38.0 & 33.2 & 26.0 & 54.5 & 43.2 & 34.5 & 39.9 & 39.7 & 38.4 & 37.4\\
~ & ~ & ~ & ~ & $\mathcal{Z}^{(2)}$ & 58.5 & 44.5 & 32.7 & 28.7 & 55.7 & 45.0 & 28.5 & 43.1 & 42.9 & 41.9& 40.9 \\ 
\hline
\multirow{7}{*}{Whisper-large} & Pretrained & $\mathcal{D}^{\setminus \kappa}$ & -- & -- & 68.0 & 46.2 & 41.7 & 56.0 & 76.0 & 64.0 & 43.5 & 57.4 & 58.5 & 57.2 & 54.1 \\
~ & Oracle & $\mathcal{D}$ & -- & -- & 85.0 & 71.2 & 70.0 & 75.7 & 88.5 & 74.5 & 70.5 & 77.0 & 77.4 & 77.2& 75.0\\
\cline{2-16}
~ & Seamless M4T & \multirow{2}{*}{$\mathcal{D}^{\setminus \kappa} \bigcup\mathcal{D}^{(\kappa)}_{\textsc{neu}}$} & \multirow{2}{*}{$\mathcal{D}^{(\kappa)}_{\textsc{neu}} \bigcup\bar{\mathcal{D}}^{(\kappa)}$} & \multirow{2}{*}{$\mathcal{X}$} & 39.2 & 22.2 & 13.5 & 38.0 & 97.2 & 36.5 & 25.5 & 40.4 & 41.2 & 37.9 & 30.0\\
~ & Seamless Expressive & ~ & ~ & ~ & 49.7 & 29.7 & 32.2 & 46.2 & 94.2 & 41.7 & 49.0 & 49.1 & 49.3 & 48.7& 40.8\\
\cline{2-16}
~ & ~ & \multirow{3}{*}{$\mathcal{D}^{\setminus \kappa} \bigcup\mathcal{D}^{(\kappa)}_{\textsc{neu}}$} & \multirow{3}{*}{$\mathcal{D}^{(\kappa)}_{\textsc{neu}} \bigcup\bar{\mathcal{D}}^{(\kappa)}$} & $\mathcal{Z}^{(0)}$ & 75.5 & 51.2 & 59.7 & 64.0 & 77.5 & 54.7 & 57.5 & 63.6 & 64.5 & 63.3& 61.1 \\
~ & GeLDA (Ours) & ~ & ~ & $\mathcal{Z}^{(1)}$ & 69.5 & 53.5 & 44.0 & 62.0 & 62.2 & 60.0 & 54.5 & 58.2 & 59.1 & 57.6 & 57.7\\
~ & ~ & ~ & ~ & $\mathcal{Z}^{(2)}$ & 70.5 & 56.0 & 47.5 & 59.5 & 73.2 & 55.5 & 51.5 & 59.9 & 60.7 & 59.4 & 57.5\\ 
\bottomrule
    \end{tabular}
}
\end{table*}

\subsubsection{Results}
\paragraph{Unaugmented baselines:} Table~\ref{tab:result-main} shows results for three different backbone FMs, averaged over all choices of target languages. Pretraining on the non-target multilingual set shows a certain level of effectiveness (UA 52.4, 42.8, and 57.4\% for \textsc{WavLM-large}, \textsc{emotion2vec-base}, and \textsc{Whisper-large}, respectively). This is because the model learns the shared emotion expressions across languages.

\paragraph{Input-space DA baselines:} \textsc{Seamless Expressive} outperforms \textsc{Seamless M4T} for all three FM backbones, but neither improves performance over the baselines, and both remain biased toward predicting \textsc{neutral}. We attribute this to the difficulty of generating high-quality, emotion-preserving speech in the input space.

\paragraph{Our proposed GeLDA method:} We apply GeLDA at different adapter layers for comparison and achieve the best average performance at $\mathcal{Z}^{(0)}$ for all backbone FMs. GeLDA outperforms the baseline models, \textsc{WavLM-large}, \textsc{emotion2vec-base}, and \textsc{Whisper-large}, by 0.2\%,  3.2\%, and 6.2\% in UA, respectively. Moreover, GeLDA's improvement is more prominent in emotion categories where the baseline models suffer (e.g., 18.0\% point improvement in \textsc{fear} and 14.0\% in \textsc{surprise} compared to the best baseline backbone, \textsc{Whisper-large}). These results demonstrate that GeLDA effectively addresses the data imbalance issue by improving the long-tail classes' performance more. We attribute the superiority of $\mathcal{Z}^{(0)}$ to its rich, well-structured representations inherited from the FMs. Augmenting at $\mathcal{Z}^{(0)}$ with a sufficiently powerful diffusion model generates diverse, high-quality samples and enables the fine-tuning of more downstream layers. Conversely, $\mathcal{Z}^{(1)}$ and $\mathcal{Z}^{(2)}$ discard task-irrelevant information, which reduces sample diversity. Appendix~\ref{apdx:additional_zero_shot_SER} provides additional ablation tests. As shown in Fig.~\ref{fig:aug_num}, although these feature spaces can simplify generation, reduced diversity leads to performance degradation as the number of augmented samples increases. We also observe that the input-space DA methods show a sharp decrease in performance as they augment more data. Per-language breakdowns of the results are also provided in Appendix~\ref{apdx:ser-per-lang}.

\begin{figure}[ht]
  \centering
  \includegraphics[width=\linewidth]{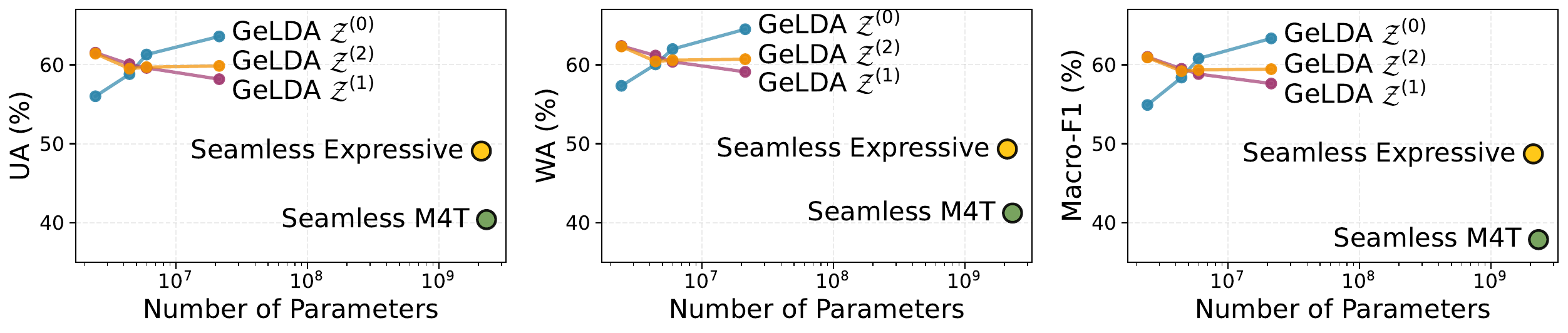}
  \caption{Graph of GeLDA performance across various diffusion model sizes. Results use \textsc{Whisper-large} as the backbone FM with 200 augmented samples per emotion.}
  \label{fig:diffusion_size}
\end{figure}

\paragraph{Ablation study on different diffusion model sizes:}
Fig.~\ref{fig:diffusion_size} analyzes the effect of diffusion model size. Small diffusion models suffice for DA in $\mathcal{Z}^{(1)}$ and $\mathcal{Z}^{(2)}$, and can outperform $\mathcal{Z}^{(0)}$ due to their lower-dimensional, more abstract feature spaces that require less generative power. As model size increases, however, they become more prone to overfitting. In contrast, the lowest-layer feature $\mathcal{Z}^{(0)}$ tends to contain richer information, which necessitates a more capable diffusion model to improve the representativeness of the synthesized features. Finally, we emphasize that, compared to input-space DA, which requires larger models to capture surface-level variability, our approach achieves significantly better performance while requiring a model size more than an order of magnitude smaller.

\begin{table}[t]
\centering
\caption{Ablation study on subdomain conditioning for the diffusion model. The backbone FM is \textsc{Whisper-large}.}
\label{tab:ser-ablation_cond}
\small
\setlength{\tabcolsep}{5pt}
\renewcommand{\arraystretch}{1.1}
\resizebox{\linewidth}{!}{
\begin{tabular}{lcrrrr}
\toprule
\textbf{Method} & \textbf{Aug. Space} & \textbf{UA (\%)} & \textbf{WA (\%)} & 
        \textbf{Macro-F1 (\%)} & 
        \textbf{UA w/o Neutral (\%)}\\
\midrule
\multirow{3}{*}{\makecell{No\\language\\condition}}& $\mathcal{Z}^{(0)}$ &57.8 & 58.7 & 56.5 & 50.7 \\ 
 & $\mathcal{Z}^{(1)}$ &57.9 & 58.7 & 56.9 & 51.3 \\ 
& $\mathcal{Z}^{(2)}$ &57.6 & 58.8 & 56.6 & 50.8 \\ 
\hline
\multirow{3}{*}{\makecell{Fixed\\language\\embedding}} & $\mathcal{Z}^{(0)}$ &63.1 & 64.0 & 63.1 & 60.6 \\ 
~ & $\mathcal{Z}^{(1)}$ &58.9 & 59.6 & 58.4 & 59.0 \\ 
~ & $\mathcal{Z}^{(2)}$ &59.2 & 60.1 & 59.3 & 58.0 \\
\bottomrule
\end{tabular}
}
\vspace{-0.2cm}
\end{table}

\paragraph{Ablation study on different subdomain conditioning:}
To assess the role of subdomain (language) conditioning, we compare several conditioning strategies in Table~\ref{tab:ser-ablation_cond}. Removing conditioning causes a large performance drop, showing that subdomain conditioning is essential for effective DA in zero-shot SER. We also test a fixed language embedding baseline by extracting the learned language embedding token from \textsc{Whisper} and using the same embedding for all utterances in a given language. This performs slightly worse than our method but is overall comparable. We hypothesize that our random embedding selection provides additional implicit subdomain cues beyond language, such as speaker characteristics, thereby increasing diversity and improving robustness.

\paragraph{Additional ablations:}
Since the GT $\mathcal{D}^{(\kappa)}$ is already a few-shot dataset for the subdomain (language) $\kappa$, additional results for this few-shot setting are provided in Appendix~\ref{apdx:ser-few-shot}.

\subsection{Imbalanced Image Classification}\label{sec:exp_cv}

\subsubsection{Task Definition and Dataset}
We further evaluate GeLDA on imbalanced image classification using ImageNet-LT and Places-LT datasets~\cite{imagenetlt}. ImageNet-LT is derived from ImageNet-2012~\cite{imagenet_orig} and contains 1,000 categories and 186.3K images. Places-LT is a large-scale dataset for place recognition and scene understanding, with 365 categories and 1.7M images. Following prior work, we report accuracy on \textsc{many} ($>100$ samples per class), \textsc{medium} ($20$--$100$), and \textsc{small} ($<20$) classes to reflect long-tailed learning. Our goal is to improve accuracy on \textsc{small} classes while preserving performance on \textsc{many} and \textsc{medium} classes. 

\subsubsection{Experiment setup}
As the baseline, we adopt LIFT~\cite{lift}, which uses \textsc{CLIP}~\cite{clip} \textsc{ViT-B/16} as the backbone and achieves SOTA performance. We implement GeLDA on top of LIFT and set $L{=}1$ for the task adapter to match its architecture. In GeLDA, we condition the diffusion model on the \textsc{CLIP} text embedding obtained from the prompt ``This is a picture of [label]''. For each label in the \textsc{small} classes, we generate 100 additional augmented samples and fine-tune LIFT using them along with the existing labeled samples. Additional details are in Appendix~\ref{apdx:implementation_details}.

Among the compared methods, LiVT~\cite{learning_imbalanced_data_with_vision_transformers} and DeiT-LT~\cite{DeiT-LT} are trained from scratch, while BALLAD~\cite{ballad}, VL-LTR~\cite{vl-ltr}, and LIFT~\cite{lift} fine-tune \textsc{CLIP} on the target dataset. LDMLR~\cite{ldmlr} also performs generative latent DA, but does not leverage FMs or semantic relationships across subdomains. In addition, we employ Stable Diffusion 3 (\textsc{SD3})~\cite{sd3}, a SOTA image generation model, for input-space DA. \textsc{SD3} uses the same text prompt as GeLDA, and \textsc{SD3-Qwen} uses Qwen~\cite{qwen2.5} to generate diverse prompts per label (image examples in Appendix~\ref{apdx:image_generation}). We apply the test-time ensembling (TTE)~\cite{lift} technique to each method to improve robustness at test time.

\begin{table}[t]
\centering
\caption{Imbalanced image classification performance (\%) on ImageNet-LT and Places-LT. 
}
\label{tab:imagenetlt}
\small
\setlength{\tabcolsep}{3pt}
\renewcommand{\arraystretch}{1.1}

\resizebox{\linewidth}{!}{
\begin{tabular}{lcc|ccc}
\toprule
\textbf{Method} & \textbf{Aug. Space} & \textbf{Overall} & \textsc{many} & \textsc{medium} & \textsc{small} \\
\hline
\rowcolor{gray!20}\multicolumn{6}{c}{Dataset: ImageNet-LT} \\
LiVT~\cite{learning_imbalanced_data_with_vision_transformers}        & -- & 60.9 & 73.6 & 56.4 & 41.0 \\
DeiT-LT~\cite{DeiT-LT}     & -- & 59.1 & 66.6 & 58.3 & 40.0 \\
BALLAD~\cite{ballad}      & -- & 75.7 & 79.1 & 74.5 & 69.8 \\
VL-LTR~\cite{vl-ltr}      & -- & 77.2 & 84.5 & 74.6 & 59.3 \\
LCReg$^*$~\cite{LCReg} & $\mathcal{Z}$ & 55.3 & 66.1 & 52.8 & 36.2 \\
LDMLR$^*$~\cite{ldmlr} & $\mathcal{Z}$ & 44.8 & 57.0 & 41.2 & 23.4 \\ 
\hline
LIFT~\cite{lift}        &  \multirow{2}{*}{--} & 77.0 & 80.2 & 76.1 & 71.5 \\
+ TTE       & ~ & 78.3 & 81.2 & 77.4 & 73.4 \\
\hline
\multicolumn{6}{l}{\textit{Augmentation on top of LIFT}} \\
\hline
SD3                  & \multirow{2}{*}{$\mathcal{X}$} & 76.8 & 80.5 & 76.5 & 67.6 \\
+ TTE            & ~ & 78.0 & 81.4 & 77.7 & 69.2 \\
SD3 Qwen             & \multirow{2}{*}{$\mathcal{X}$} & 76.8 & 80.4 & 76.4 & 68.0 \\
+ TTE       & ~ & 77.9 & 81.3 & 77.7 & 69.2 \\
\hline
GeLDA (Ours)         & \multirow{2}{*}{$\mathcal{Z}^{(0)}$} & 77.0 & 79.7 & 75.7 & 73.9 \\
+ TTE   & ~ & 78.2 & 80.8 & 77.1 & \textbf{74.7} \\
\hline
\rowcolor{gray!20}\multicolumn{6}{c}{Dataset: Places-LT} \\
LiVT~\cite{learning_imbalanced_data_with_vision_transformers}        & -- & 40.8 & 48.1 & 40.6 & 27.5 \\
BALLAD~\cite{ballad}      & -- & 49.5 & 49.3 & 50.2 & 48.4 \\
VL-LTR~\cite{vl-ltr}      & -- & 50.1 & 54.2 & 48.5 & 42.0 \\
\hline
LIFT~\cite{lift}        &  \multirow{2}{*}{--} & 51.5 & 51.3 & 52.2 & 50.4 \\
+ TTE  & ~ & 52.2 & 51.7 & 53.1 & 50.9 \\
\hline
\multicolumn{6}{l}{\textit{Augmentation on top of LIFT}} \\
\hline
SD3                 & \multirow{2}{*}{$\mathcal{X}$} & 50.9 & 51.8 & 52.1 & 46.3 \\
+ TTE           & ~ & 51.6 & 52.0 & 53.0 & 47.4 \\
SD3 Qwen            & \multirow{2}{*}{$\mathcal{X}$} & 50.8 & 51.5 & 52.5 & 45.6 \\
+ TTE      & ~ & 51.5 & 51.7 & 53.3 & 47.0 \\
\hline
GeLDA (Ours)        & \multirow{2}{*}{$\mathcal{Z}^{(0)}$}& 51.5 & 50.6 & 51.8 & \textbf{52.5} \\
+ TTE  & ~ & 52.1 & 51.4 & 52.6 & 52.4 \\
\bottomrule
\multicolumn{5}{l}{
{\scriptsize
*: Train ResNet from scratch as the backbone model.}}
\end{tabular}
}
\vspace{-0.4cm}
\end{table}

\begin{table}[t]
\centering
\caption{Performance (\%) on few- and zero-shot image classification results on ImageNet-LT and Places-LT}
\label{tab:imagenetlt-zero-shot}
\begingroup
\setlength{\tabcolsep}{3pt}
\resizebox{1.0\linewidth}{!}{
\begin{tabular}{lcc|ccccccc}
\toprule
\textbf{Method} & \textbf{Aug. Space} & \textbf{Overall} & \textbf{Many} & \textbf{Med} & \textbf{Small} & \textbf{5-shot} & \textbf{3-shot} & \textbf{1-shot} & \textbf{0-shot} \\
\hline
\rowcolor{gray!20}\multicolumn{10}{c}{Dataset: ImageNet-LT} \\
LIFT & \multirow{2}{*}{--} & 76.5 & 80.3 & 76.2 & 66.9 & 54.7 & 44.3 & 26.2 & 8.3 \\
+ TTE       & ~ &77.8 & 81.3 & 77.6 & 68.7 & 56.3 & 47.4 & 26.7 & 10.2 \\
\hline
\multicolumn{9}{l}{\textit{Augmentation on top of LIFT}} \\
\hline
SD3 & \multirow{2}{*}{$\mathcal{X}$} & 76.6 & 80.5 & 76.5 & 66.1 & 57.2 & 55.2 & 54.8 & 41.4 \\
+ TTE       & ~ & 77.9 & \textbf{81.5} & \textbf{77.9} & 67.4 & 59.0 & 58.8 & 56.2 & 42.6 \\
SD3 Qwen & \multirow{2}{*}{$\mathcal{X}$} & 76.6 & 80.3 & 76.5 & 66.3 & 54.3 & 56.2 & 51.4 & 42.5 \\
+TTE & ~ & 77.8 & 81.4 & 77.9 & 67.5 & 56.6 & 59.2 & 52.2 & 43.8 \\
\hline
GeLDA (Ours) &  \multirow{2}{*}{$\mathcal{Z}^{(0)}$} & 76.9 & 79.8 & 75.6 & 72.7 & 62.9 & 67.4 & 65.0 & 56.2 \\
+ TTE       & ~ & \textbf{78.1} & 80.9 & 77.1 & \textbf{73.9} & \textbf{65.2} & \textbf{69.7} & \textbf{65.9} & \textbf{57.5} \\
\hline
\rowcolor{gray!20}\multicolumn{10}{c}{Dataset: Places-LT} \\
LIFT & \multirow{2}{*}{--} & 50.3 & 51.4 & 52.5 & 43.2 & 46.0 & 25.4 & 11.0 & 3.2 \\
+TTE & ~ &51.0 & 52.0 & \textbf{53.4} & 43.7 & 46.7 & 26.3 & 11.5 & 3.4 \\
\hline
\multicolumn{9}{l}{\textit{Augmentation on top of LIFT}} \\
\hline
SD3 & \multirow{2}{*}{$\mathcal{X}$} & 50.4 & 51.6 & 52.4 & 43.7 & 44.8 & 36.0 & 36.4 & 27.6 \\
+TTE               & ~ & 51.2 & \textbf{52.0} & 53.3 
& 44.8 & 46.8 & 37.5 & 37.4 & 28.8 \\
SD3 Qwen & \multirow{2}{*}{$\mathcal{X}$} & 50.3 & 51.3 & 52.7 & 42.8 & 42.3 & 34.8 & 34.2 & 26.3 \\
+TTE & ~ & 51.1 & 51.7 & 53.6 & 44.4 & 44.4 & 36.4 & 35.8 & 27.9 \\
\hline
GeLDA (Ours) &  \multirow{2}{*}{$\mathcal{Z}^{(0)}$} & 50.9 & 50.3 & 51.6 & \textbf{50.6} & \textbf{52.8} & 44.9 & \textbf{48.1} & \textbf{35.5} \\
+TTE         & ~ &  \textbf{51.7} & 51.3 & 52.6 & 50.2 & 52.6 & \textbf{45.1} & 46.7 & 33.4 \\
\bottomrule
\end{tabular}
}
\vspace{-0.3cm}
\endgroup
\end{table}

\subsubsection{Results}
\paragraph{Classification of imbalanced image datasets (original benchmark setup):} 
Tables~\ref{tab:imagenetlt} present the results of different augmentation methods applied on top of the LIFT baseline model for ImageNet-LT and Places-LT. Using \textsc{SD3} as an augmentation method degrades the performance of the \textsc{small} class, suggesting its generated samples are not helpful and can further bias the classifier toward \textsc{many} and \textsc{medium} classes. In contrast, GeLDA achieves new SOTA performance on the \textsc{small} classes for both datasets (74.7 and 52.5\%, respectively), improving tail accuracy while maintaining nearly the same performance for the \textsc{many} and \textsc{medium} classes.

\paragraph{Zero-shot image classification:}
Since each \textsc{small} class has at least five samples, we further evaluate GeLDA under more extreme long-tail conditions by randomly selecting 20 \textsc{small} classes and simulating 5, 3, 1, and 0-shot training samples per class, i.e., by discarding some samples. We repeat each experiment five times and report the averaged results in Tables~\ref{tab:imagenetlt-zero-shot}. GeLDA significantly improves performance on fewer and zero-shot classes with more pronounced improvements as the number of original samples decreases. In the zero-shot case, for example, training only with GeLDA-synthesized samples improves accuracy by 47.3\% and 32.1\% points on ImageNet-LT and Places-LT, respectively. Additionally, GeLDA achieves the best overall performance across all settings, indicating its effectiveness under severe imbalance and when adding new categories to existing classifiers. We postulate that the label information, fed to the CLIP text encoder, plays a key role as it contains the semantics about the label through the FM's pretraining, which LIFT might have forgotten during its fine-tuning. 
\section{Conclusion and Future Work}
We introduced GeLDA, a semantics-aware generative latent data augmentation (DA) framework that addresses data scarcity using conditional diffusion models. By operating in a low-dimensional latent space, GeLDA was substantially more efficient than input-space DA. We introduced augmented label conditioning to provide semantic cues that help low-resource classes and subdomains connect to high-resource data. We also analyzed how different choices of latent space affect GeLDA. We demonstrated its effectiveness and SOTA performance on two practical yet challenging tasks: zero-shot speech emotion recognition for low-resource languages and long-tailed image classification.

While GeLDA is promising, several directions remain open. We applied it only to a lightweight task adapter. Although this ensures foundation-model-agnostic applicability, the latent spaces of the foundation model's intermediate layers remain underexplored. In addition, we focused on recognition tasks, leaving the opportunity open to extending GeLDA to sequence learning tasks, such as TTS and ASR. 
\section{Broader Impact}
The motivation of this work is to broaden the practical accessibility of recent advances in deep learning. Although large foundation models have demonstrated strong performance across a wide range of tasks, they are often difficult to apply directly in data-scarce real-world settings, including low-resource language technologies, healthcare applications, and recognition of underrepresented (minority) classes. We focus on two representative challenges, zero-shot speech emotion recognition and highly imbalanced image classification, and demonstrate that the proposed GeLDA framework can be applied effectively and efficiently across various domains. By enabling more reliable learning under limited supervision and skewed data distributions, this approach has the potential to improve performance for languages, classes, and objects that are often underrepresented in real-world datasets and, consequently, in the models trained on those data. Eventually, we expect that the GeLDA method can improve the privacy-preservation and social bias in AI models by overcoming the inherent lack of labeled data issue.


\bibliography{ref_clean_short_authors}
\bibliographystyle{sty/icml2026}

\clearpage
\newpage
\appendix
\onecolumn

\section{Multilingual Speech Emotion Dataset}
\begin{table}[ht]
    \centering
    \caption{Summary of multilingual speech emotion dataset in hours.}
    {
    \begingroup
    
\setlength{\tabcolsep}{3pt}
    \begin{tabular}{l|rrrrrrr|r}
        \toprule
        \multirow{2}{*}{\textbf{Language}} & \multicolumn{7}{c|}{\textbf{Emotion}} & 
        \multirow{2}{*}{\textbf{Total}}\\
        \cline{2-8}
        ~ &\textbf{Angry} & \textbf{Disgust} & \textbf{Fear} & \textbf{Happy} & \textbf{Neutral} & \textbf{Sad} & \multicolumn{1}{c|}{\textbf{Surprise}} & ~ \\
        \midrule
Amharic &0.21 & 0.00 & 0.29 & 0.24 & 0.27 & 0.25 & 0.00 & 1.27 \\
Bangla &0.72 & 0.70 & 0.73 & 0.72 & 0.71 & 0.71 & 0.72 & 5.01 \\
English &7.89 & 4.67 & 4.41 & 9.12 & 9.45 & 8.67 & 6.27 & 50.48 \\
French &0.21 & 0.16 & 0.13 & 0.21 & 0.15 & 0.28 & 0.16 & 1.30 \\
German &0.06 & 0.03 & 0.02 & 0.03 & 0.04 & 0.04 & 0.00 & 0.22 \\
Greek &0.08 & 0.08 & 0.07 & 0.08 & 0.00 & 0.10 & 0.00 & 0.41 \\
Italian &0.54 & 0.64 & 0.63 & 0.51 & 0.57 & 0.65 & 0.59 & 4.13 \\
Mandarin &3.06 & 0.03 & 0.07 & 2.84 & 3.42 & 3.81 & 2.51 & 15.74 \\
Persian &0.62 & 0.00 & 0.02 & 0.13 & 0.84 & 0.30 & 0.06 & 1.96 \\
Polish &0.03 & 0.00 & 0.02 & 0.00 & 0.02 & 0.00 & 0.00 & 0.07 \\
Russian &0.23 & 0.22 & 0.24 & 0.24 & 0.18 & 0.18 & 0.00 & 1.29 \\
Spanish &0.02 & 0.02 & 0.01 & 0.02 & 0.01 & 0.02 & 0.00 & 0.11 \\
Turkish &0.07 & 0.00 & 0.00 & 0.06 & 0.00 & 0.08 & 0.00 & 0.21 \\
Urdu &0.04 & 0.00 & 0.00 & 0.04 & 0.04 & 0.04 & 0.00 & 0.17 \\
\midrule
Total & 13.78 & 6.56 & 6.66 & 14.24 & 15.69 & 15.15 & 10.30 & 82.37 \\
        \bottomrule
    \end{tabular}
    \endgroup
    }
    \label{tab:ser_dataset}
\end{table}

We collect a multilingual speech emotion dataset following EmoBox~\cite{emobox}, and its language and emotion distribution is summarized in Table \ref{tab:ser_dataset}, which exhibits a heavy but common imbalance issue across emotion labels and languages, e.g., \textsc{disgust}, \textsc{fear}, and \textsc{surprise} are entirely missing in Turkish and Urdu datasets, more than 50 hours of English data vs. 0.07 hours of Polish data, etc. Such extreme sparsity and missing labels make multilingual SER particularly difficult.

\section{Implementation Details}
\label{apdx:implementation_details}
In this section, we described the detailed model architecture and training details of the GeLDA framework.

\subsection{Model architecture details}

\begin{table}[!ht]
    \footnotesize
    \centering
    \caption{
    Detailed model architecture for the multilingual SER model.
    }
    \begin{tabular}{|l|c|c|c|c|}
    \hline
        ~ & \textbf{Target}& \multicolumn{3}{c|}{\textbf{Backbone foundation model}} \\\cline{3-5}
        ~ & \textbf{latent space} & \textbf{WavLM-large} & \textbf{emotion2vec-base} & \textbf{Whisper-large} \\  \hline
        \rowcolor{gray!20}\multicolumn{5}{|c|}{Task adapter} \\ \hline
        1st linear layer output dim & -- & \multicolumn{3}{c|}{512} \\ \hline
        2nd linear layer output dim & -- & \multicolumn{3}{c|}{256}\\ \hline
        Last linear layer output dim & -- & \multicolumn{3}{c|}{7}\\ \hline
        \rowcolor{gray!20}\multicolumn{5}{|c|}{Diffusion model} \\ \hline
        \multirow{3}{*}{2D reshape size}& $\mathcal{Z}^{(0)}$ & (32, 32) & (28, 28) & (36, 36) \\ \cline{2-5}
        ~ & $\mathcal{Z}^{(1)}$ & \multicolumn{3}{c|}{(24, 24)}\\ \cline{2-5}
        ~ & $\mathcal{Z}^{(2)}$ & \multicolumn{3}{c|}{(16, 16)}\\ \hline
        Timestep embedding dim & -- & \multicolumn{3}{c|}{256}  \\ \hline 
        Emotion embedding dim & -- & \multicolumn{3}{c|}{256}  \\ \hline 
        Final conditioning vector dimension & -- &\multicolumn{3}{c|}{256} \\ \hline
    \end{tabular}
    \label{tab:model-hyp-ser}
\end{table}

\begin{table}[!ht]
    \footnotesize
    \centering
    \caption{Detailed model architecture for imbalanced image classification.}
    \begin{tabular}{|l|c|c|}
    \hline
        ~ & \textbf{Target}& \multicolumn{1}{c|}{\textbf{Backbone foundation model}} \\ \cline{3-3}
        ~ & \textbf{latent space}&  \textbf{ViT-B/16}\\  \hline
        \rowcolor{gray!20}\multicolumn{3}{|c|}{Task adapter} \\ \hline
        Single linear layer output dim & -- & 1000 for ImageNet-LT and 365 for Places-LT \\ \hline
        \rowcolor{gray!20}\multicolumn{3}{|c|}{Diffusion model} \\ \hline
        2D reshape size & $\mathcal{Z}^{(0)}$ & (28, 28) \\ \hline
        Timestep embedding dim & -- &  256 \\ \hline
        Text embedding dim & -- &  256 \\ \hline
        Final conditioning vector dimension & -- & 256\\ \hline
    \end{tabular}
    \label{tab:model-hyp-image}
\end{table}

\paragraph{Task Adapter Architecture:}
The task adapters for SER are composed of three linear layers applied to the temporally pooled representations from the FMs. Temporal pooling is performed via average pooling. The output dimensions of the first two layers are listed in Table~\ref{tab:model-hyp-ser}. ReLU activations are applied after the first and second layers, and a softmax activation is used in the final layer. For imbalanced image classification, we apply GeLDA to LIFT~\cite{lift}, which employs AdaptFormer~\cite{chen2022adaptformer} and a single linear classifier as its adapter. The dimension of the linear layer is the same as the number of classification labels of each task.

\paragraph{Diffusion Model Architecture:}
We employ a Transformer-based diffusion model adapted from the \textsc{image\_transformer\_v2} architecture in the $k$-diffusion framework \cite{k-diffusion}\footnote{\url{https://github.com/crowsonkb/k-diffusion}}. To bridge the gap between the 1D latent vectors generated by GeLDA and the 2D requirement of the original architecture, we first reshape the 1D vectors into 2D matrices. Detailed dimensions for this reshaping are provided in Tables~\ref{tab:model-hyp-ser} and~\ref{tab:model-hyp-image}. If the size of the 2D matrix is larger than that of the 1D vector, we pad the vector to match the matrix size. In our pilot study, we observed that using a 2D matrix as input performs better than using a 1D vector directly. The resulting matrices are then partitioned into non-overlapping $2 \times 2$ patches and embedded as a sequence of tokens. The backbone of the diffusion models features a two-stage hierarchical Transformer design: the first stage comprises two global self-attention blocks with a width of 256, while the second stage consists of four blocks with a width of 512. Each block integrates multi-head self-attention and feed-forward layers with residual connections, utilizing a dropout rate of 0.05. Please refer to the original code for further details.

To condition the diffusion model during generation, we adopt a CFG approach~\cite{cfg}. We apply conditioning dropout with probability 0.1 during training. For SER, the conditioning signal is constructed by combining an emotion embedding (from a learnable lookup table), a language embedding, and a diffusion's timestep embedding. We find that using a randomly sampled latent vector for language $k$, $z^{(l)} \sim \mathcal{D}^{(k)}$, is more effective than using a fixed language embedding (Table~\ref{tab:ser-ablation_cond}). Finally, the emotion, language, and timestep embeddings are concatenated and passed through a linear projection layer to match the final conditioning dimension of 256. For the image classification task, we use text embeddings as conditioning signals by feeding the prompt template ``This is a photo of a [label]'' into a pretrained text encoder of \textsc{CLIP}. Text embedding is also concatenated with the timestep embedding and projected to 256 dimensions through a linear layer.

During inference for both tasks, we employ the DPM++~SDE sampler~\cite{dpm_solver_pp} with a CFG scale of 1.2 and 20 sampling steps, chosen empirically. We observe that larger CFG scales or more sampling steps reduce the diversity of the generated latent vectors, which harms the effectiveness of DA.

\subsection{Training details}
\paragraph{SER model:}
To train the SER model, we follow the training settings of EmoBox~\cite{emobox}. The backbone foundation models are kept frozen throughout training, and only the task adapter parameters are optimized. We use the Adam optimizer with $\beta_1=0.9$ and $\beta_2=0.999$, a learning rate of $1\times10^{-3}$, and a batch size of 32. All models are trained for 100 epochs using a learning rate warm-up over the first 10 epochs. For language-specific fine-tuning, the task adapter is further trained for an additional 100 epochs under the same optimization settings.

\paragraph{Image Classification Model:} All training settings follow the original LIFT framework~\cite{lift}. We adopt the logit-adjusted (LA) loss~\cite{la_loss} and optimize the model using SGD with a batch size of 128, momentum of 0.9, and weight decay of $5 \times 10^{-4}$. In Stage 1 (generic model training), we train the LIFT model following the original setup for 10 epochs. In Stage 3 (fine-tuning), we further train the model for four additional epochs using GeLDA-generated and GT samples from the \textsc{small} classes. During fine-tuning, we freeze all LIFT parameters except the final linear layer. We find this freezing technique avoids performance degradation on the \textsc{many} and \textsc{medium} classes, even though fine-tuning uses only \textsc{small}-class samples.

\paragraph{Diffusion Model:}
We follow the training setup of the $k$-diffusion framework. Noise levels are sampled from a cosine-interpolated distribution, and loss weighting is based on a soft-min-SNR scheme. Optimization is performed using AdamW with a learning rate of $5 \times 10^{-4}$ and weight decay of $1 \times 10^{-4}$. An exponential moving average (EMA) of parameters is maintained using an inverse decay schedule with a maximum value of 0.9999. We trained the diffusion model for 400K iterations with a batch size of 64.

\section{Efficiency comparison between GeLDA and input-space DA}
\label{apdx:efficiency_comparison}
\begin{table}[ht]
    \centering
        \footnotesize
        \caption{Efficiency comparison between input-space DA with generative models and GeLDA. NP refers to the number of parameters of the generative models.} 
        {\footnotesize
        \begin{tabular}{cccccc}
            \toprule 
            \textbf{Model} & \multicolumn{1}{c}{\textbf{Training time}} & \multicolumn{1}{c}{\textbf{Training data}} & \multicolumn{1}{c}{\textbf{Inference time}} & \multicolumn{1}{c}{\textbf{NP}} & \multicolumn{1}{c}{\textbf{Data size}}\\
            \midrule
            Seamless M4T v2 & \multicolumn{1}{c}{-} & 4.5M hours & 480 ms/sample& 2.3 B & \multirow{2}{*}{139.5 KB/sample}\\
            Seamless Expressive &  \multicolumn{1}{c}{-}& 29.8K hours & 420 ms/sample& 2.1 B & \\
            \hline
            Stable Diffusion 3 & \multicolumn{1}{c}{-} & \multicolumn{1}{c}{-} & 3,990 ms/sample & 2.2 B & 1560.7 KB/sample  \\
            \hline
            GeLDA at $\mathcal{Z}^{(0)}$ & 4.8 hours & $\approx$82.4 hours (for SER)& 160 ms/sample& 21.3 M & 6.7 KB/sample\\
            \bottomrule
        \end{tabular}
        }
        \label{tab:compare2}
    \end{table}

The efficiency comparison between the input-space DA and GeLDA is presented in Table~\ref{tab:compare2}. For the GeLDA, we employ the one from SER task, which uses the \textsc{Whisper-large} as the backbone and $\mathcal{Z}^{(0)}$ as the target embedding space. Note that the average durations of speech utterances of the generated speech of \textsc{Seamless M4T} and \textsc{Seamless Expressive} models are approximately three seconds. The training and inference time are calculated with an NVIDIA L40S GPU with 46 GB of VRAM.

GeLDA is vastly more efficient than input-space DA, requiring far fewer parameters and substantially less training data while enabling significantly faster inference. This efficiency makes GeLDA particularly well-suited for extremely low-resource settings.

\section{Additional Results for Zero-Shot SER Experiments}\label{apdx:additional_zero_shot_SER}
\subsection{Ablation Study on the Number of Augmented Samples}
\label{apdx:num_aug_samples}

\begin{figure}[ht]
    \centering
    \includegraphics[width=0.78\textwidth]{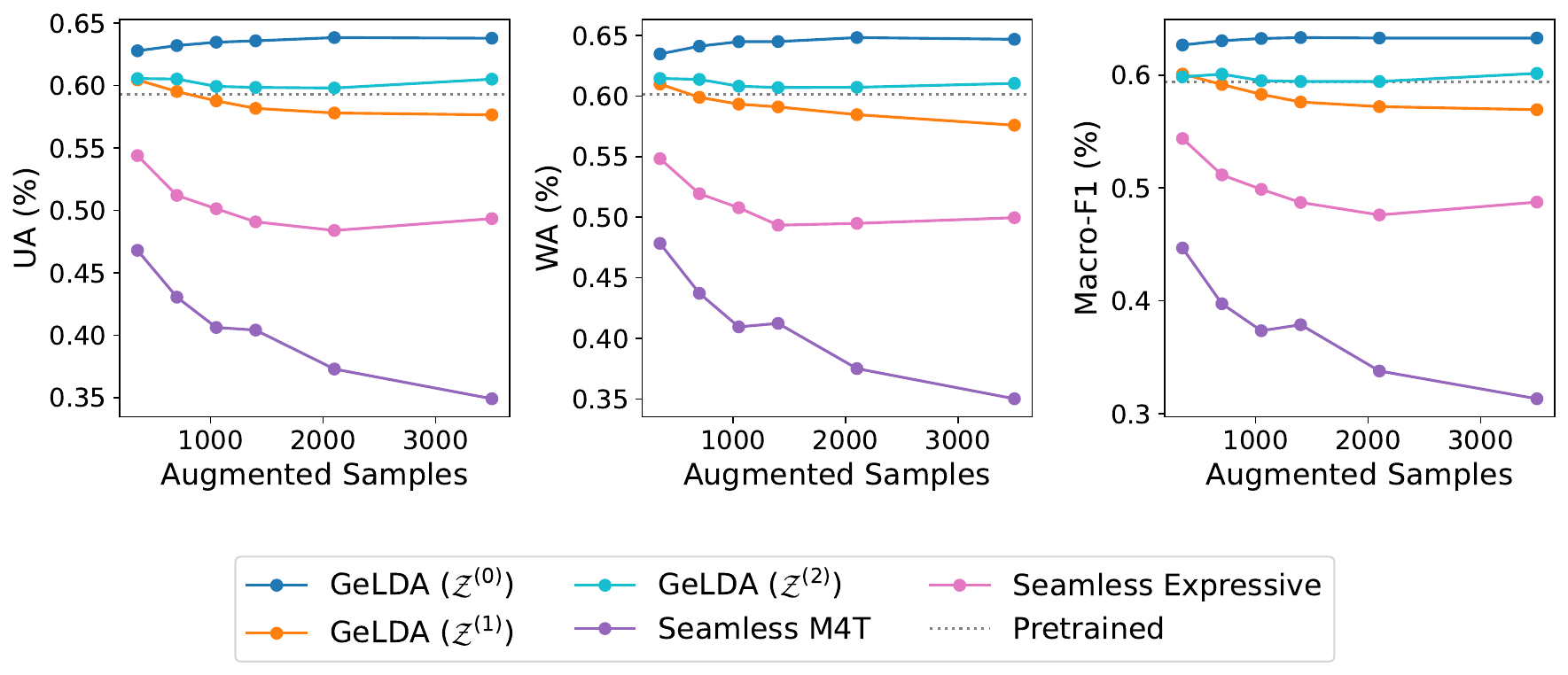}
    \caption{GeLDA performance with varying numbers of augmented samples using \textsc{Whisper-large} as the backbone FM.}
    \label{fig:aug_num}
\end{figure}

We vary the number of augmented samples and examine performance trends for \textsc{Whisper-large} backbone (Fig.~\ref{fig:aug_num}). Augmentation in $\mathcal{Z}^{(0)}$ improves performance as the number of augmented samples increases, suggesting that the generated samples are well aligned with the GT conditional distribution $p^{(\kappa)}(x|y)$ and thus improve overall performance. In contrast, for $\mathcal{Z}^{(1)}$ and $\mathcal{Z}^{(2)}$, performance typically drops as the number of augmented samples increases. This indicates that the generated samples in these latent spaces are not well aligned with $p^{(\kappa)}(x|y)$, causing the model to overfit to the augmented-sample distribution. Moreover, the input-space DA methods (\textsc{Seamless M4T} and \textsc{Seamless Expressive}) degrade even more severely, highlighting the difficulty of applying input-space DA.

\clearpage
\subsection{Visualization of the Augmented Latent Vectors}

\begin{figure}[ht]
    \centering
    \begin{subfigure}{0.30\textwidth}
        \centering
        \includegraphics[width=\linewidth]{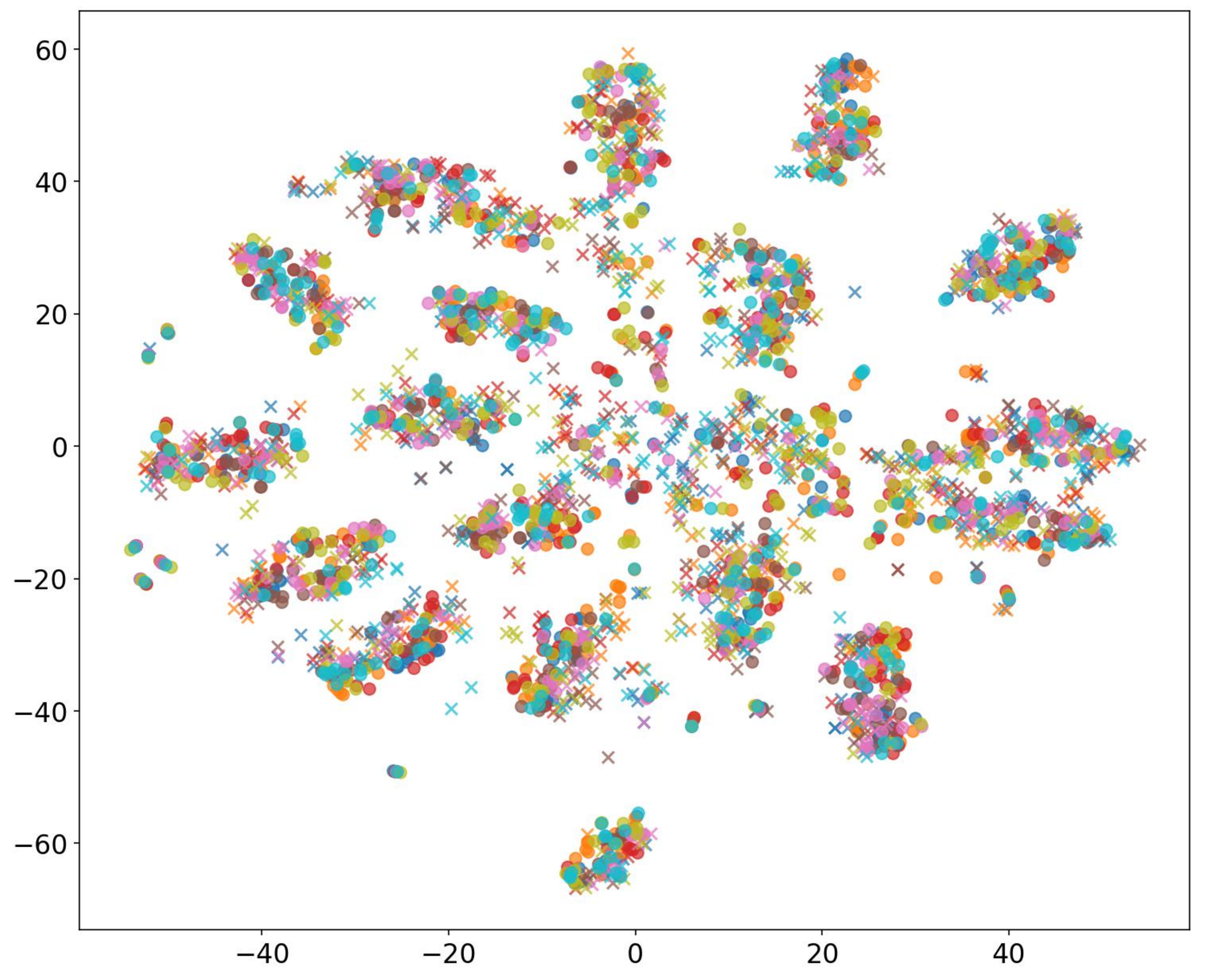}
        \caption{$\mathcal{Z}^{(0)}$}
        \label{fig:latent_vector_visualization_z0}
    \end{subfigure}
    \begin{subfigure}{0.30\textwidth}
        \centering
        \includegraphics[width=\linewidth]{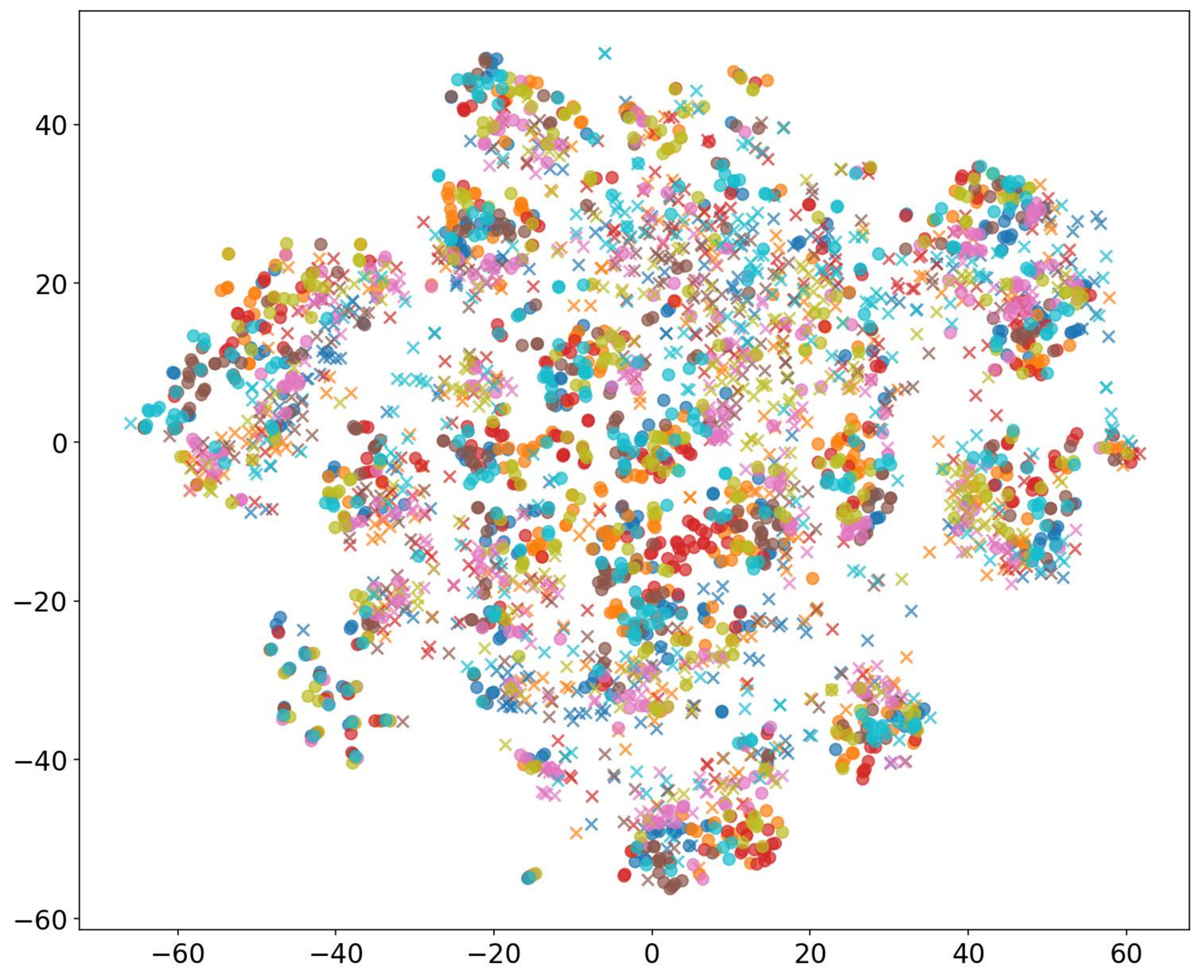}
        \caption{$\mathcal{Z}^{(1)}$}
        \label{fig:latent_vector_visualization_z1}
    \end{subfigure}
    \begin{subfigure}{0.377\textwidth}
        \centering
        \includegraphics[width=\linewidth]{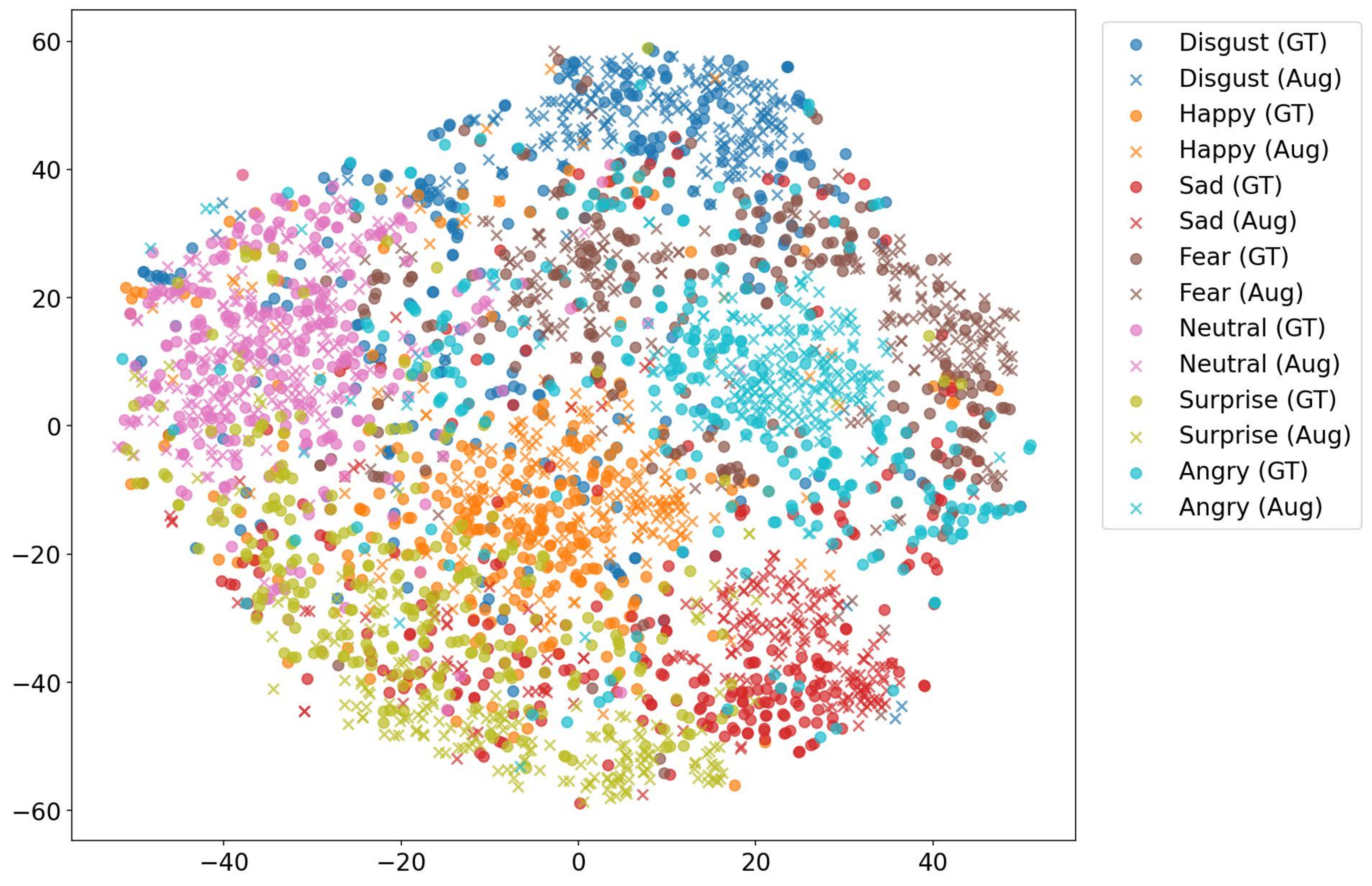}
        \caption{$\mathcal{Z}^{(2)}$}
        \label{fig:latent_vector_visualization_z2}
    \end{subfigure}

    \caption{Examples of t-SNE plots of generated embeddings from GeLDA and GT embeddings from the test set for the SER task. From left to right, the plots correspond to the embedding spaces $\mathcal{Z}^{(0)}$, $\mathcal{Z}^{(1)}$, and $\mathcal{Z}^{(2)}$. All results are shown for the Italian case, and the foundation model is \textsc{Whisper-large}.}
    \label{fig:latent_vector_visualization}
\end{figure}

Fig.~\ref{fig:latent_vector_visualization} illustrates the embedding-space distributions of the augmented latent vectors and the GT test embeddings for the Italian SER case. Since we simulate a zero-shot SER setting, the generative model used to produce augmented embeddings has not been exposed to any emotion classes other than \textsc{neutral} during training. Despite this constraint, the augmented embeddings are well aligned with the GT test embeddings across all three embedding spaces. This close overlap indicates that the proposed GeLDA framework is capable of generating high-quality latent representations even under zero-shot conditions. Among them, the augmented embeddings in $\mathcal{Z}^{(0)}$ are observed to be the most overlapped with the GT embeddings, which aligns with our results in Table~\ref{tab:result-main}.

Furthermore, a clear evolution of representation structure can be observed from Fig. \ref{fig:latent_vector_visualization_z0} to \ref{fig:latent_vector_visualization_z2}. In $\mathcal{Z}^{(0)}$, the embeddings form a small number of clusters, which likely encode non-emotional factors such as speaker identity or recording environment. As the embeddings propagate toward deeper layers, these task-irrelevant factors are progressively suppressed. In $\mathcal{Z}^{(2)}$, the embeddings exhibit more distinct emotion-oriented separation, suggesting that the classifier increasingly discards irrelevant information and focuses on emotion-discriminative features.

\subsection{Per Language Breakdown of Zero-shot SER Experiments}
\label{apdx:ser-per-lang}

\begin{table*}[ht]
\caption{Per-language breakdown of zero-shot SER experiments using \textsc{Whisper-large} as the backbone FM. Note that we also report results for Amharic, Bangla, Persian, Polish, Russian, and Urdu, which are not included in the average results in Table~\ref{tab:result-main}. Greek and Turkish are not selected as target languages for zero-shot SER because their datasets do not include a \textsc{neutral} emotion category.}
    \label{tab:result-ser-zero-shot-per-lang}
    \centering
    \footnotesize
    \resizebox{\textwidth}{!}{
    \begin{tabular}{lcccccccccc|cccc}

    \toprule
        \multirow{2}{*}{\textbf{Model}} & \textbf{Pretrain} & \textbf{Fine-tune} & \multirow{2}{*}{\textbf{Aug. Space}} & \multicolumn{7}{c}{\textbf{Per emotion recall (\%)}} & 
        \multirow{2}{*}{\textbf{UA (\%)}} & \multirow{2}{*}{\textbf{WA (\%)}} & 
        \multirow{2}{*}{\textbf{Macro-F1 (\%)}} & 
        \multirow{2}{*}{\textbf{UA w/o Neutral(\%)}}\\ 
        \cline{5-11}
        ~ & \textbf{Dataset} & \textbf{Dataset} & ~ &\textbf{Angry} & \textbf{Disgust} & \textbf{Fear} & \textbf{Happy} & \textbf{Neutral} & \textbf{Sad} & \multicolumn{1}{c}{\textbf{Surprise}} & ~ & ~ & ~ & ~\\
        \midrule
  \hline
\rowcolor{gray!20}\multicolumn{15}{c}{Language: Amharic} \\
Pretrained & $\mathcal{D}^{\setminus \kappa}$ & -- & -- & 79.61 & -- & 58.27 & 41.60 & 72.56 & 24.50 & -- & 55.31 & 55.79 & 57.81 & 51.00 \\
Oracle & $\mathcal{D}$ & -- & -- & 98.00 & -- & 92.67 & 97.33 & 95.33 & 93.33 & -- & 95.46 & 95.45 & 95.44 & 95.33 \\
\hline
~ & \multirow{3}{*}{$\mathcal{D}^{\setminus \kappa} \cup \mathcal{D}^{(\kappa)}_{\texttt{neu}}$} & \multirow{3}{*}{$\mathcal{D}^{(\kappa)}_{\texttt{neu}} \cup \bar{\mathcal{D}}^{(\kappa)}$} & $\mathcal{Z}^{(0)}$ & 83.20 & -- & 75.85 & 48.76 & 87.18 & 25.93 & -- & 64.18 & 64.88 & 62.77 & 58.43 \\
GeLDA (Ours) & ~ & ~ & $\mathcal{Z}^{(1)}$ & 83.75 & -- & 74.02 & 47.38 & 71.28 & 19.94 & -- & 59.27 & 59.85 & 57.77 & 56.27 \\
~ & ~ & ~ & $\mathcal{Z}^{(2)}$ & 86.50 & -- & 61.15 & 57.58 & 76.92 & 23.65 & -- & 61.16 & 61.64 & 60.34 & 57.22 \\
\hline
\rowcolor{gray!20}\multicolumn{15}{c}{Language: Bangla} \\
Pretrained & $\mathcal{D}^{\setminus \kappa}$ & -- & -- & 86.83 & 3.67 & 41.67 & 29.17 & 48.00 & 86.00 & 41.67 & 48.14 & 48.14 & 44.39 & 48.17 \\
Oracle & $\mathcal{D}$ & -- & -- & 76.00 & 60.33 & 73.00 & 77.67 & 90.00 & 78.00 & 75.67 & 75.81 & 75.81 & 75.79 & 73.44 \\
\hline
~ & \multirow{3}{*}{$\mathcal{D}^{\setminus \kappa} \cup \mathcal{D}^{(\kappa)}_{\texttt{neu}}$} & \multirow{3}{*}{$\mathcal{D}^{(\kappa)}_{\texttt{neu}} \cup \bar{\mathcal{D}}^{(\kappa)}$} & $\mathcal{Z}^{(0)}$ & 89.00 & 1.50 & 54.00 & 78.50 & 86.83 & 50.83 & 70.83 & 61.64 & 61.64 & 57.03 & 57.44 \\
GeLDA (Ours) & ~ & ~ & $\mathcal{Z}^{(1)}$ & 92.50 & 2.17 & 53.83 & 87.50 & 61.50 & 51.33 & 42.00 & 55.83 & 55.83 & 51.51 & 54.89 \\
~ & ~ & ~ & $\mathcal{Z}^{(2)}$ & 80.00 & 5.17 & 44.33 & 78.33 & 66.17 & 51.67 & 45.00 & 52.95 & 52.95 & 49.64 & 50.75 \\
\hline
\rowcolor{gray!20}\multicolumn{15}{c}{Language: French} \\
Pretrained & $\mathcal{D}^{\setminus \kappa}$ & -- & -- & 67.87 & 59.21 & 59.60 & 63.14 & 67.26 & 66.67 & 64.47 & 64.03 & 64.05 & 63.85 & 63.49 \\
Oracle & $\mathcal{D}$ & -- & -- & 81.67 & 67.33 & 73.67 & 75.33 & 81.67 & 70.33 & 70.00 & 74.29 & 74.00 & 73.80 & 73.06 \\
\hline
Seamless M4T & \multirow{2}{*}{$\mathcal{D}^{\setminus \kappa} \cup \mathcal{D}^{(\kappa)}_{\texttt{neu}}$} & \multirow{2}{*}{$\mathcal{D}^{(\kappa)}_{\texttt{neu}} \cup \bar{\mathcal{D}}^{(\kappa)}$} & $\mathcal{X}$ & 38.55 & 29.39 & 16.67 & 41.96 & 92.26 & 45.15 & 45.05 & 44.15 & 42.78 & 43.05 & 36.13 \\
Seamless Expressive & ~ & ~ & $\mathcal{X}$ & 51.81 & 37.28 & 42.42 & 48.63 & 84.52 & 50.21 & 61.90 & 53.82 & 52.99 & 52.87 & 48.71 \\
\hline
~ & \multirow{3}{*}{$\mathcal{D}^{\setminus \kappa} \cup \mathcal{D}^{(\kappa)}_{\texttt{neu}}$} & \multirow{3}{*}{$\mathcal{D}^{(\kappa)}_{\texttt{neu}} \cup \bar{\mathcal{D}}^{(\kappa)}$} & $\mathcal{Z}^{(0)}$ & 68.27 & 56.58 & 60.10 & 57.25 & 77.98 & 37.13 & 64.10 & 60.20 & 59.58 & 59.24 & 57.24 \\
GeLDA (Ours) & ~ & ~ & $\mathcal{Z}^{(1)}$ & 68.67 & 61.84 & 41.41 & 61.18 & 78.57 & 56.96 & 67.03 & 62.24 & 62.19 & 62.12 & 59.52 \\
~ & ~ & ~ & $\mathcal{Z}^{(2)}$ & 73.49 & 52.19 & 53.03 & 60.00 & 72.02 & 58.23 & 68.13 & 62.44 & 62.50 & 62.45 & 60.85 \\
\hline
\rowcolor{gray!20}\multicolumn{15}{c}{Language: German} \\
Pretrained & $\mathcal{D}^{\setminus \kappa}$ & -- & -- & 97.92 & 44.32 & 74.04 & 70.09 & 89.77 & 89.74 & -- & 77.65 & 81.23 & 77.78 & 75.22 \\
Oracle & $\mathcal{D}$ & -- & -- & 99.00 & 73.00 & 86.00 & 93.33 & 97.67 & 97.33 & -- & 91.10 & 92.73 & 91.79 & 89.73 \\
\hline
Seamless M4T & \multirow{2}{*}{$\mathcal{D}^{\setminus \kappa} \cup \mathcal{D}^{(\kappa)}_{\texttt{neu}}$} & \multirow{2}{*}{$\mathcal{D}^{(\kappa)}_{\texttt{neu}} \cup \bar{\mathcal{D}}^{(\kappa)}$} & $\mathcal{X}$ & 77.13 & 23.48 & 22.19 & 70.94 & 100.00 & 66.90 & -- & 60.11 & 65.06 & 58.34 & 52.13 \\
Seamless Expressive & ~ & ~ & $\mathcal{X}$ & 77.42 & 34.09 & 47.88 & 56.75 & 95.69 & 79.03 & -- & 65.14 & 67.31 & 67.56 & 59.03 \\
\hline
~ & \multirow{3}{*}{$\mathcal{D}^{\setminus \kappa} \cup \mathcal{D}^{(\kappa)}_{\texttt{neu}}$} & \multirow{3}{*}{$\mathcal{D}^{(\kappa)}_{\texttt{neu}} \cup \bar{\mathcal{D}}^{(\kappa)}$} & $\mathcal{Z}^{(0)}$ & 100.00 & 39.02 & 90.76 & 91.11 & 87.31 & 93.27 & -- & 83.58 & 87.97 & 83.52 & 82.83 \\
GeLDA (Ours) & ~ & ~ & $\mathcal{Z}^{(1)}$ & 84.43 & 34.85 & 73.03 & 79.32 & 48.61 & 88.51 & -- & 68.12 & 71.98 & 67.01 & 72.03 \\
~ & ~ & ~ & $\mathcal{Z}^{(2)}$ & 98.67 & 53.79 & 70.11 & 76.41 & 60.23 & 83.56 & -- & 73.80 & 77.34 & 73.31 & 76.51 \\
\hline
\rowcolor{gray!20}\multicolumn{15}{c}{Language: Italian} \\
Pretrained & $\mathcal{D}^{\setminus \kappa}$ & -- & -- & 40.64 & 58.21 & 31.79 & 75.64 & 78.59 & 64.87 & 18.33 & 52.58 & 52.58 & 50.56 & 48.25 \\
Oracle & $\mathcal{D}$ & -- & -- & 77.67 & 73.00 & 71.67 & 75.67 & 87.33 & 75.00 & 70.67 & 75.88 & 75.88 & 75.84 & 73.94 \\
\hline
Seamless M4T & \multirow{2}{*}{$\mathcal{D}^{\setminus \kappa} \cup \mathcal{D}^{(\kappa)}_{\texttt{neu}}$} & \multirow{2}{*}{$\mathcal{D}^{(\kappa)}_{\texttt{neu}} \cup \bar{\mathcal{D}}^{(\kappa)}$} & $\mathcal{X}$ & 18.97 & 10.38 & 6.03 & 28.85 & 98.33 & 7.56 & 6.15 & 25.18 & 25.18 & 21.04 & 12.99 \\
Seamless Expressive & ~ & ~ & $\mathcal{X}$ & 29.87 & 25.00 & 26.28 & 52.05 & 97.69 & 18.97 & 35.51 & 40.77 & 40.77 & 39.67 & 31.28 \\
\hline
~ & \multirow{3}{*}{$\mathcal{D}^{\setminus \kappa} \cup \mathcal{D}^{(\kappa)}_{\texttt{neu}}$} & \multirow{3}{*}{$\mathcal{D}^{(\kappa)}_{\texttt{neu}} \cup \bar{\mathcal{D}}^{(\kappa)}$} & $\mathcal{Z}^{(0)}$ & 64.10 & 55.38 & 56.41 & 65.26 & 85.00 & 44.62 & 50.90 & 60.24 & 60.24 & 59.72 & 56.11 \\
GeLDA (Ours) & ~ & ~ & $\mathcal{Z}^{(1)}$ & 56.79 & 62.31 & 47.82 & 59.49 & 72.31 & 45.26 & 42.44 & 55.20 & 55.20 & 54.79 & 52.35 \\
~ & ~ & ~ & $\mathcal{Z}^{(2)}$ & 48.46 & 60.64 & 40.90 & 57.56 & 89.87 & 46.41 & 34.62 & 54.07 & 54.07 & 53.13 & 48.10 \\
\hline
\rowcolor{gray!20}\multicolumn{15}{c}{Language: Persian} \\
Pretrained & $\mathcal{D}^{\setminus \kappa}$ & -- & -- & 58.40 & -- & 76.19 & 78.91 & 42.06 & 77.04 & 19.05 & 58.61 & 53.81 & 45.64 & 61.92 \\
Oracle & $\mathcal{D}$ & -- & -- & 94.67 & -- & 57.00 & 88.00 & 91.67 & 80.00 & 69.67 & 80.22 & 89.17 & 79.10 & 77.87 \\
\hline
~ & \multirow{3}{*}{$\mathcal{D}^{\setminus \kappa} \cup \mathcal{D}^{(\kappa)}_{\texttt{neu}}$} & \multirow{3}{*}{$\mathcal{D}^{(\kappa)}_{\texttt{neu}} \cup \bar{\mathcal{D}}^{(\kappa)}$} & $\mathcal{Z}^{(0)}$ & 75.71 & -- & 61.90 & 77.55 & 83.20 & 48.89 & 50.34 & 66.27 & 73.19 & 59.47 & 62.88 \\
GeLDA (Ours) & ~ & ~ & $\mathcal{Z}^{(1)}$ & 72.09 & -- & 52.38 & 76.87 & 70.11 & 46.30 & 68.03 & 64.30 & 67.94 & 55.47 & 63.13 \\
~ & ~ & ~ & $\mathcal{Z}^{(2)}$ & 61.76 & -- & 66.67 & 73.47 & 87.70 & 61.11 & 47.62 & 66.39 & 70.83 & 58.04 & 62.12 \\
\hline
\rowcolor{gray!20}\multicolumn{15}{c}{Language: Polish} \\
Pretrained & $\mathcal{D}^{\setminus \kappa}$ & -- & -- & 77.78 & -- & 3.33 & -- & 13.33 & -- & -- & 31.48 & 31.48 & 29.80 & 40.56 \\
Oracle & $\mathcal{D}$ & -- & -- & 89.00 & -- & 35.33 & -- & 51.33 & -- & -- & 58.52 & 58.52 & 52.07 & 62.17 \\
\hline
~ & \multirow{3}{*}{$\mathcal{D}^{\setminus \kappa} \cup \mathcal{D}^{(\kappa)}_{\texttt{neu}}$} & \multirow{3}{*}{$\mathcal{D}^{(\kappa)}_{\texttt{neu}} \cup \bar{\mathcal{D}}^{(\kappa)}$} & $\mathcal{Z}^{(0)}$ & 74.44 & -- & 20.00 & -- & 57.78 & -- & -- & 50.74 & 50.74 & 47.27 & 47.22 \\
GeLDA (Ours) & ~ & ~ & $\mathcal{Z}^{(1)}$ & 80.00 & -- & 21.11 & -- & 47.78 & -- & -- & 49.63 & 49.63 & 45.72 & 50.56 \\
~ & ~ & ~ & $\mathcal{Z}^{(2)}$ & 85.56 & -- & 18.89 & -- & 57.78 & -- & -- & 54.07 & 54.07 & 50.10 & 52.22 \\
\hline
\rowcolor{gray!20}\multicolumn{15}{c}{Language: Russian} \\
Pretrained & $\mathcal{D}^{\setminus \kappa}$ & -- & -- & 46.97 & 27.03 & 5.19 & 31.82 & 85.96 & 27.08 & -- & 37.34 & 36.81 & 34.76 & 27.62 \\
Oracle & $\mathcal{D}$ & -- & -- & 78.00 & 47.00 & 50.33 & 58.67 & 62.67 & 40.67 & -- & 56.08 & 56.95 & 56.17 & 54.93 \\
\hline
~ & \multirow{3}{*}{$\mathcal{D}^{\setminus \kappa} \cup \mathcal{D}^{(\kappa)}_{\texttt{neu}}$} & \multirow{3}{*}{$\mathcal{D}^{(\kappa)}_{\texttt{neu}} \cup \bar{\mathcal{D}}^{(\kappa)}$} & $\mathcal{Z}^{(0)}$ & 68.94 & 20.72 & 11.85 & 50.76 & 80.70 & 18.75 & -- & 41.95 & 42.64 & 38.18 & 34.20 \\
GeLDA (Ours) & ~ & ~ & $\mathcal{Z}^{(1)}$ & 81.82 & 19.82 & 20.00 & 51.52 & 51.75 & 27.08 & -- & 42.00 & 43.06 & 39.75 & 40.05 \\
~ & ~ & ~ & $\mathcal{Z}^{(2)}$ & 62.12 & 18.02 & 5.19 & 34.85 & 76.32 & 23.96 & -- & 36.74 & 36.81 & 33.03 & 28.83 \\
\hline
\rowcolor{gray!20}\multicolumn{15}{c}{Language: Spanish} \\
Pretrained & $\mathcal{D}^{\setminus \kappa}$ & -- & -- & 59.05 & 53.70 & 25.00 & 28.70 & 52.38 & 37.96 & -- & 42.80 & 42.68 & 45.29 & 40.88 \\
Oracle & $\mathcal{D}$ & -- & -- & 81.00 & 72.33 & 48.33 & 58.33 & 87.00 & 55.67 & -- & 66.98 & 66.82 & 67.29 & 63.13 \\
\hline
Seamless M4T & \multirow{2}{*}{$\mathcal{D}^{\setminus \kappa} \cup \mathcal{D}^{(\kappa)}_{\texttt{neu}}$} & \multirow{2}{*}{$\mathcal{D}^{(\kappa)}_{\texttt{neu}} \cup \bar{\mathcal{D}}^{(\kappa)}$} & $\mathcal{X}$ & 21.90 & 26.85 & 9.26 & 10.19 & 99.05 & 25.93 & -- & 32.20 & 31.93 & 29.09 & 18.83 \\
Seamless Expressive & ~ & ~ & $\mathcal{X}$ & 40.00 & 23.15 & 12.96 & 26.85 & 98.10 & 18.52 & -- & 36.59 & 36.29 & 34.80 & 24.30 \\
\hline
~ & \multirow{3}{*}{$\mathcal{D}^{\setminus \kappa} \cup \mathcal{D}^{(\kappa)}_{\texttt{neu}}$} & \multirow{3}{*}{$\mathcal{D}^{(\kappa)}_{\texttt{neu}} \cup \bar{\mathcal{D}}^{(\kappa)}$} & $\mathcal{Z}^{(0)}$ & 69.52 & 53.70 & 32.41 & 42.59 & 60.00 & 43.52 & -- & 50.29 & 50.16 & 50.84 & 48.35 \\
GeLDA (Ours) & ~ & ~ & $\mathcal{Z}^{(1)}$ & 67.62 & 54.63 & 13.89 & 49.07 & 48.57 & 49.07 & -- & 47.14 & 47.04 & 46.57 & 46.86 \\
~ & ~ & ~ & $\mathcal{Z}^{(2)}$ & 61.90 & 57.41 & 25.93 & 43.52 & 71.43 & 34.26 & -- & 49.08 & 48.91 & 48.87 & 44.60 \\
\hline
\rowcolor{gray!20}\multicolumn{15}{c}{Language: Urdu} \\
Pretrained & $\mathcal{D}^{\setminus \kappa}$ & -- & -- & 90.67 & -- & -- & 60.00 & 37.33 & 33.33 & -- & 55.33 & 55.33 & 55.72 & 61.33 \\
Oracle & $\mathcal{D}$ & -- & -- & 96.00 & -- & -- & 73.33 & 68.00 & 57.33 & -- & 73.67 & 73.67 & 73.58 & 75.56 \\
\hline
~ & \multirow{3}{*}{$\mathcal{D}^{\setminus \kappa} \cup \mathcal{D}^{(\kappa)}_{\texttt{neu}}$} & \multirow{3}{*}{$\mathcal{D}^{(\kappa)}_{\texttt{neu}} \cup \bar{\mathcal{D}}^{(\kappa)}$} & $\mathcal{Z}^{(0)}$ & 72.00 & -- & -- & 76.00 & 64.00 & 26.67 & -- & 59.67 & 59.67 & 57.81 & 58.22 \\
GeLDA (Ours) & ~ & ~ & $\mathcal{Z}^{(1)}$ & 80.00 & -- & -- & 68.00 & 44.00 & 42.67 & -- & 58.67 & 58.67 & 57.61 & 63.56 \\
~ & ~ & ~ & $\mathcal{Z}^{(2)}$ & 85.33 & -- & -- & 58.67 & 58.67 & 42.67 & -- & 61.33 & 61.33 & 60.59 & 62.22 \\
\bottomrule
    \end{tabular}
}
\end{table*}

In Table~\ref{tab:result-ser-zero-shot-per-lang}, we provide per-language results for the zero-shot SER setting summarized in Table~\ref{tab:result-main}. We use \textsc{Whisper-large} as the backbone foundation model since it achieves the strongest overall performance among others. Across all languages except French, GeLDA consistently improves over the \textsc{pretrained} baseline, demonstrating the effectiveness of our method and its broad applicability across diverse languages.
For most languages, augmenting in $\mathcal{Z}^{(0)}$ yields the best results; however, in a few cases, operating in alternative latent spaces leads to larger gains than $\mathcal{Z}^{(0)}$. We hypothesize that this reflects GeLDA’s reliance on transferring information from high-resource subdomains to augment low-resource ones, consequently, its gains may depend on the degree of structural alignment between subdomains in the chosen latent space.

\newpage
\clearpage
\section{Experiments on Few-shot Speech Emotion Recognition}
\label{apdx:ser-few-shot}

\begin{table*}[ht]
\caption{Average test results for few-shot SER models on four languages (French, German, Spanish, and Italian).}
    \label{tab:result-ser-few-shot-4lang}
    \centering
    \footnotesize
    \resizebox{\textwidth}{!}{
    \begin{tabular}{llcccccccccc|cccc}
    \toprule
\multirow{2}{*}{\textbf{FM}} & 
        \multirow{2}{*}{\textbf{Model}} & \textbf{Pretrain} & \textbf{Fine-tune} & \multirow{2}{*}{\textbf{Aug. Space}} & \multicolumn{7}{c}{\textbf{Per emotion recall (\%)}} & 
        \textbf{UA} & \textbf{WA} & \textbf{Macro-F1} & \textbf{UA w/o}\\ 
        \cline{6-12}
        ~ & ~ & \textbf{Dataset} & \textbf{Dataset} & ~ &\textbf{Angry} & \textbf{Disgust} & \textbf{Fear} & \textbf{Happy} & \textbf{Neutral} & \textbf{Sad} & \multicolumn{1}{c}{\textbf{Surprise}} & \textbf{(\%)} & \textbf{(\%)} & \textbf{(\%)} & \textbf{ Neutral (\%)}\\
        \midrule
\multirow{10}{*}{WavLM-large} & Pretrained & $\mathcal{D}$ & -- & -- & 79.00 & 68.00 & 64.00 & 68.00 & 77.00 & 72.00 & 61.00 & 70.75 & 71.00 & 71.10 & 68.67 \\ 
~ & Fine-tuned & $\mathcal{D}$ & $\mathcal{D}^{(\kappa)}$ & -- &80.76 & 69.20 & 63.76 & 66.53 & 78.37 & 73.25 & 62.91 & 71.53 & 71.76 & 71.67 & 69.40 \\ 
\cline{2-16}
~ & Seamless M4T & \multirow{2}{*}{$\mathcal{D}$} & \multirow{2}{*}{$\mathcal{D}^{(\kappa)} \bigcup\bar{\mathcal{D}}^{(\kappa)}$} & \multirow{2}{*}{$\mathcal{X}$} & 77.58 & 68.09 & 59.43 & 65.43 & 79.51 & 70.48 & 61.90 & 69.70 & 69.86 & 69.54 & 67.15 \\ 
~ & Seamless Expressive & ~ & ~ & ~ & 76.41 & 67.25 & 60.11 & 64.89 & 75.79 & 70.52 & 63.25 & 68.85 & 69.21 & 68.78 & 67.07 \\ 
\cline{2-16}
~ & ~ & ~ & ~ &$\mathcal{Z}^{(0)}$ & 80.58 & 73.20 & 69.75 & 71.10 & 82.48 & 70.69 & 66.68 & 74.39 & 74.55 & 74.38 & 72.00 \\ 
~ & Latent Filling & $\mathcal{D}$ & $\mathcal{D}^{(\kappa)} \bigcup\bar{\mathcal{D}}^{(\kappa)}$ & $\mathcal{Z}^{(1)}$ & 81.77 & 67.22 & 67.06 & 67.18 & 80.24 & 70.69 & 63.48 & 72.00 & 72.13 & 72.09 & 69.57 \\ 
~ & ~ & ~ & ~ &$\mathcal{Z}^{(2)}$ & 80.30 & 68.77 & 64.51 & 69.13 & 78.31 & 73.04 & 61.54 & 71.80 & 71.98 & 71.96 & 69.55 \\ 
\cline{2-16}
~ & ~ & ~ & ~ & $\mathcal{Z}^{(0)}$ & 82.54 & 74.85 & 70.30 & 73.53 & 84.44 & 74.43 & 66.73 & 76.34 & 76.27 & 76.31 & 73.73 \\ 
~ & GeLDA (Ours) &  $\mathcal{D}$ & $\mathcal{D}^{(\kappa)} \bigcup\bar{\mathcal{D}}^{(\kappa)}$ & $\mathcal{Z}^{(1)}$ & 80.44 & 72.25 & 67.44 & 71.80 & 81.33 & 72.38 & 62.29 & 73.87 & 73.72 & 73.74 & 71.10 \\ 
~ & ~ & ~ & ~ & $\mathcal{Z}^{(2)}$ & 78.54 & 68.74 & 62.02 & 68.61 & 75.98 & 71.89 & 61.06 & 70.49 & 70.54 & 70.46 & 68.48 \\ 
\hline
\multirow{10}{*}{emotion2vec-base} & Pretrained & $\mathcal{D}$ & -- & -- & 66.00 & 57.00 & 50.00 & 49.00 & 70.00 & 65.00 & 49.00 & 58.84 & 59.03 & 58.87 & 56.00 \\
~ & Fine-tuned & $\mathcal{D}$ & $\mathcal{D}^{(\kappa)}$ & -- & 65.66 & 63.29 & 51.46 & 49.94 & 70.75 & 66.07 & 51.60 & 60.70 & 60.73 & 60.55 & 58.00 \\
\cline{2-16}
~ & Seamless M4T & \multirow{2}{*}{$\mathcal{D}$} & \multirow{2}{*}{$\mathcal{D}^{(\kappa)} \bigcup\bar{\mathcal{D}}^{(\kappa)}$} & \multirow{2}{*}{$\mathcal{X}$} & 59.09 & 61.78 & 49.70 & 50.58 & 67.71 & 58.59 & 52.19 & 57.60 & 57.72 & 57.49 & 55.32 \\
~ & Seamless Expressive & ~ & ~ & ~ & 63.57 & 60.86 & 51.98 & 47.42 & 73.14 & 68.91 & 52.30 & 60.50 & 60.41 & 60.18 & 57.51 \\
\cline{2-16}
~ & ~ & ~ & ~ &$\mathcal{Z}^{(0)}$ & 68.45 & 61.60 & 55.66 & 53.61 & 75.43 & 73.41 & 54.26 & 64.30 & 64.37 & 64.19 & 61.17 \\
~ & Latent Filling & $\mathcal{D}$ & $\mathcal{D}^{(\kappa)} \bigcup\bar{\mathcal{D}}^{(\kappa)}$ & $\mathcal{Z}^{(1)}$ & 68.16 & 55.94 & 48.85 & 53.36 & 74.17 & 71.08 & 52.88 & 61.53 & 61.87 & 61.22 & 58.38 \\
~ & ~ & ~ & ~ & $\mathcal{Z}^{(2)}$ & 66.13 & 58.53 & 49.69 & 50.78 & 71.08 & 65.56 & 50.95 & 59.81 & 60.09 & 59.68 & 56.94 \\
\cline{2-16}
~ & ~ & ~ & ~ &$\mathcal{Z}^{(0)}$ & 67.31 & 65.05 & 60.13 & 60.08 & 76.48 & 64.94 & 57.21 & 65.43 & 65.51 & 65.25 & 62.45 \\
~ & GeLDA (Ours) & $\mathcal{D}$ & $\mathcal{D}^{(\kappa)} \bigcup\bar{\mathcal{D}}^{(\kappa)}$ & $\mathcal{Z}^{(1)}$ & 64.25 & 63.17 & 54.48 & 52.97 & 71.58 & 66.76 & 52.66 & 61.82 & 61.43 & 61.54 & 59.05 \\
~ & ~ & ~ & ~ & $\mathcal{Z}^{(2)}$ & 63.38 & 56.70 & 47.11 & 51.04 & 68.52 & 64.97 & 49.16 & 58.11 & 58.25 & 57.87 & 55.39 \\
\hline
\multirow{10}{*}{Whisper-large} & Pretrained & $\mathcal{D}$ & -- & -- & 85.00 & 71.00 & 70.00 & 76.00 & 88.00 & 75.00 & 70.00 & 77.06 & 77.36 & 77.18 & 74.50 \\
~ & Fine-tuned & $\mathcal{D}$ & $\mathcal{D}^{(\kappa)}$ & -- & 86.23 & 74.75 & 69.50 & 76.80 & 90.00 & 76.58 & 70.82 & 78.56 & 78.68 & 78.51 & 75.78 \\
\cline{2-16}
~ & Seamless M4T & \multirow{2}{*}{$\mathcal{D}$} & \multirow{2}{*}{$\mathcal{D}^{(\kappa)} \bigcup\bar{\mathcal{D}}^{(\kappa)}$} & \multirow{2}{*}{$\mathcal{X}$} & 82.01 & 73.23 & 63.46 & 76.47 & 87.43 & 76.20 & 70.27 & 76.14 & 76.44 & 75.90 & 73.61 \\
~ & Seamless Expressive & ~ & ~ & ~ & 84.27 & 72.96 & 67.08 & 75.29 & 87.49 & 76.15 & 72.14 & 77.03 & 77.39 & 77.01 & 74.65 \\
\cline{2-16}
~ & ~ & ~ & ~ & $\mathcal{Z}^{(0)}$ & 85.82 & 74.57 & 68.73 & 80.11 & 90.34 & 77.44 & 72.91 & 79.23 & 79.55 & 79.22 & 76.60 \\
~ & Latent Filling & $\mathcal{D}$ & $\mathcal{D}^{(\kappa)} \bigcup\bar{\mathcal{D}}^{(\kappa)}$ & $\mathcal{Z}^{(1)}$ & 85.12 & 72.52 & 72.72 & 78.72 & 89.08 & 73.85 & 72.90 & 78.46 & 78.72 & 78.46 & 75.97 \\
~ & ~ & ~ & ~ & $\mathcal{Z}^{(2)}$ & 85.07 & 72.79 & 69.41 & 76.51 & 89.00 & 74.85 & 71.25 & 77.61 & 77.84 & 77.61 & 74.98 \\
\cline{2-16}
~ & ~ & ~ & ~ & $\mathcal{Z}^{(0)}$ & 85.74 & 75.25 & 74.38 & 80.20 & 92.19 & 75.40 & 71.22 & 80.07 & 80.26 & 79.93 & 77.03 \\
~ & GeLDA (Ours) & $\mathcal{D}$ & $\mathcal{D}^{(\kappa)} \bigcup\bar{\mathcal{D}}^{(\kappa)}$ & $\mathcal{Z}^{(1)}$ & 83.29 & 72.83 & 70.13 & 80.70 & 90.52 & 74.20 & 68.77 & 78.19 & 78.29 & 78.01 & 74.99 \\
~ & ~ & ~ & ~ & $\mathcal{Z}^{(2)}$ & 85.19 & 72.95 & 67.71 & 77.32 & 86.45 & 75.90 & 70.05 & 77.28 & 77.50 & 77.19 & 74.85 \\ 
\bottomrule
    \end{tabular}
}
\end{table*}

\begin{table*}[ht]
\caption{Average test results for few-shot SER models across all 12 low-resource languages (Amharic, Bangla, French, German, Greek, Italian, Persian, Polish, Russian, Spanish, Turkish, and Urdu). 
}
    \label{tab:result-ser-few-shot-12lang}
    \centering
    \footnotesize
    \resizebox{\textwidth}{!}{
    \begin{tabular}{llcccccccccc|cccc}
\toprule
\multirow{2}{*}{\textbf{FM}} & 
        \multirow{2}{*}{\textbf{Model}} & \textbf{Pretrain} & \textbf{Fine-tune} & \multirow{2}{*}{\textbf{Aug. Space}} & \multicolumn{7}{c}{\textbf{Per emotion recall (\%)}} & 
        \textbf{UA} & \textbf{WA} & \textbf{Macro-F1} & \textbf{UA w/o}\\ 
        \cline{6-12}
        ~ & ~ & \textbf{Dataset} & \textbf{Dataset} & ~ &\textbf{Angry} & \textbf{Disgust} & \textbf{Fear} & \textbf{Happy} & \textbf{Neutral} & \textbf{Sad} & \multicolumn{1}{c}{\textbf{Surprise}} & \textbf{(\%)} & \textbf{(\%)} & \textbf{(\%)} & \textbf{ Neutral (\%)}\\
        \midrule
\multirow{8}{*}{WavLM-large} & Pretrained & $\mathcal{D}$ & -- & -- & 83.00 & 65.00 & 64.00 & 75.00 & 72.00 & 73.00 & 65.00 & 72.37 & 73.28 & 72.20 & 70.83 \\ 
~ & Fine-tuned & $\mathcal{D}$ & $\mathcal{D}^{(\kappa)}$ & -- & 83.92 & 65.47 & 64.93 & 75.11 & 73.80 & 74.35 & 66.32 & 73.47 & 74.39 & 73.04 & 71.68 \\ 
\cline{2-16}
~ & ~ & ~ & ~ &$\mathcal{Z}^{(0)}$ & 85.22 & 70.78 & 65.76 & 78.05 & 77.03 & 75.55 & 68.29 & 76.08 & 77.40 & 75.64 & 73.94 \\ 
~ & Latent Filling & $\mathcal{D}$ & $\mathcal{D}^{(\kappa)} \bigcup\bar{\mathcal{D}}^{(\kappa)}$ & $\mathcal{Z}^{(1)}$ & 83.52 & 66.59 & 66.50 & 74.63 & 76.76 & 73.90 & 68.89 & 74.19 & 74.91 & 73.56 & 72.34 \\ 
~ & ~ & ~ & ~ &$\mathcal{Z}^{(2)}$ & 83.76 & 65.77 & 64.81 & 75.35 & 73.19 & 73.89 & 66.31 & 73.26 & 74.14 & 72.79 & 71.65 \\ 
\cline{2-16}
~ & ~ & ~ & ~ & $\mathcal{Z}^{(0)}$ & 85.21 & 69.04 & 71.29 & 79.55 & 77.80 & 78.72 & 71.29 & 77.55 & 77.93 & 76.48 & 75.85 \\ 
~ & GeLDA (Ours) &  $\mathcal{D}$ & $\mathcal{D}^{(\kappa)} \bigcup\bar{\mathcal{D}}^{(\kappa)}$ & $\mathcal{Z}^{(1)}$ & 83.25 & 68.39 & 68.70 & 78.92 & 74.93 & 76.16 & 69.96 & 75.88 & 76.00 & 74.62 & 74.23 \\ 
~ & ~ & ~ & ~ & $\mathcal{Z}^{(2)}$ & 82.93 & 64.42 & 65.84 & 73.69 & 71.41 & 73.86 & 65.28 & 72.45 & 72.87 & 71.33 & 71.00 \\ 
\hline
\multirow{8}{*}{emotion2vec-base} & Pretrained & $\mathcal{D}$ & -- & -- & 72.00 & 51.00 & 52.00 & 55.00 & 68.00 & 61.00 & 50.00 & 60.63 & 61.98 & 60.80 & 56.83 \\ 
~ & Fine-tuned & $\mathcal{D}$ & $\mathcal{D}^{(\kappa)}$ & -- & 74.13 & 55.49 & 53.89 & 57.07 & 68.76 & 63.52 & 52.04 & 62.84 & 64.10 & 62.70 & 59.36 \\ 
\cline{2-16}
~ & ~ & ~ & ~ &$\mathcal{Z}^{(0)}$ & 75.47 & 57.28 & 55.82 & 61.89 & 73.51 & 69.59 & 57.61 & 66.79 & 68.05 & 66.57 & 62.94 \\ 
~ & Latent Filling & $\mathcal{D}$ & $\mathcal{D}^{(\kappa)} \bigcup\bar{\mathcal{D}}^{(\kappa)}$ & $\mathcal{Z}^{(1)}$ & 74.24 & 50.81 & 51.93 & 56.52 & 72.87 & 67.66 & 56.12 & 63.62 & 64.85 & 63.14 & 59.55 \\ 
~ & ~ & ~ & ~ & $\mathcal{Z}^{(2)}$ & 73.69 & 52.25 & 52.42 & 56.70 & 68.82 & 63.00 & 51.59 & 62.10 & 63.43 & 61.94 & 58.27 \\ 
\cline{2-16}
~ & ~ & ~ & ~ &$\mathcal{Z}^{(0)}$ & 74.20 & 59.42 & 60.71 & 66.60 & 72.82 & 66.97 & 60.42 & 67.75 & 68.25 & 66.79 & 64.72 \\ 
~ & GeLDA (Ours) & $\mathcal{D}$ & $\mathcal{D}^{(\kappa)} \bigcup\bar{\mathcal{D}}^{(\kappa)}$ & $\mathcal{Z}^{(1)}$ & 71.29 & 55.78 & 58.23 & 61.34 & 69.55 & 66.30 & 57.37 & 65.01 & 65.26 & 63.95 & 61.72 \\ 
~ & ~ & ~ & ~ & $\mathcal{Z}^{(2)}$ & 70.90 & 50.79 & 49.97 & 58.53 & 66.84 & 61.74 & 52.02 & 60.82 & 61.71 & 60.14 & 57.32 \\ 
\hline
\multirow{8}{*}{Whisper-large} & Pretrained & $\mathcal{D}$ & -- & -- & 88.00 & 66.00 & 68.00 & 79.00 & 81.00 & 75.00 & 72.00 & 76.93 & 77.84 & 76.46 & 74.67 \\ 
~ & Fine-tuned & $\mathcal{D}$ & $\mathcal{D}^{(\kappa)}$ & -- & 88.30 & 69.03 & 66.96 & 79.12 & 80.87 & 76.22 & 71.84 & 77.29 & 78.29 & 76.75 & 75.25 \\ 
\cline{2-16}
~ & ~ & ~ & ~ & $\mathcal{Z}^{(0)}$ & 88.87 & 69.29 & 64.72 & 81.20 & 83.04 & 76.54 & 73.57 & 78.27 & 79.51 & 77.65 & 75.70 \\ 
~ & Latent Filling & $\mathcal{D}$ & $\mathcal{D}^{(\kappa)} \bigcup\bar{\mathcal{D}}^{(\kappa)}$ & $\mathcal{Z}^{(1)}$ & 88.26 & 68.38 & 69.17 & 80.10 & 80.86 & 74.32 & 72.89 & 77.54 & 78.57 & 76.94 & 75.52 \\ 
~ & ~ & ~ & ~ & $\mathcal{Z}^{(2)}$ & 88.18 & 67.01 & 68.28 & 79.28 & 80.62 & 75.76 & 72.31 & 77.25 & 78.16 & 76.61 & 75.14 \\ 
\cline{2-16}
~ & ~ & ~ & ~ & $\mathcal{Z}^{(0)}$ & 88.65 & 71.74 & 72.81 & 82.71 & 83.69 & 78.90 & 72.26 & 80.23 & 80.74 & 79.27 & 77.84 \\ 
~ & GeLDA (Ours) & $\mathcal{D}$ & $\mathcal{D}^{(\kappa)} \bigcup\bar{\mathcal{D}}^{(\kappa)}$ & $\mathcal{Z}^{(1)}$ & 85.86 & 67.83 & 72.90 & 82.28 & 81.54 & 76.42 & 71.97 & 78.45 & 78.80 & 77.20 & 76.21 \\ 
~ & ~ & ~ & ~ & $\mathcal{Z}^{(2)}$ & 88.12 & 67.56 & 66.85 & 78.81 & 79.78 & 75.39 & 72.61 & 76.96 & 77.76 & 76.09 & 74.89 \\ 
\bottomrule
    \end{tabular}
}
\end{table*}

The EmoBox dataset~\cite{emobox}, denoted as $\mathcal{D}$, is a low-resource and highly imbalanced multilingual dataset: English and Mandarin contain most of the data, while the remaining languages have substantially less. For example, English and Mandarin provide 50.48 and 15.74 hours of speech, whereas the Polish and Spanish subsets contain only 0.07 and 0.11 hours. This imbalance makes building SER models for low-resource languages challenging even under $\mathcal{D}$. We refer to building a language-specific SER model for a low-resource language $\kappa$ using the full dataset $\mathcal{D}$ as \emph{few-shot SER}, and we evaluate GeLDA alongside comparison models. In the few-shot setting, we augment the dataset to 500 samples per emotion and report results in two setups: (i) four selected target languages (French, German, Italian, and Spanish), as in Sec.~\ref{sec:exp_ser}, and (ii) 12 target languages covering all low-resource languages in the dataset.

\paragraph{Comparison Models:} For the baselines, the \textsc{Pretrained} model is trained on the full dataset $\mathcal{D}$, and the \textsc{Fine-tuned} model is further fine-tuned on the target low-resource dataset $\mathcal{D}^{(\kappa)}$, yielding a language-$\kappa$-specific SER model. We also include \textsc{Latent Filling}~\cite{latent_filling}, which applies statistical latent-space DA techniques (interpolation and noise injection) to speech. Note that this approach cannot be applied to the zero-shot SER experiments in Sec.~\ref{sec:exp_ser}, since interpolation between emotions in the low-resource language $\kappa$ is not possible in a zero-shot setting.

\paragraph{Results:} 
The average results of few-shot SER experiments on four selected languages and on all 12 low-resource languages are presented in Tables~\ref{tab:result-ser-few-shot-4lang} and~\ref{tab:result-ser-few-shot-12lang}. Additionally, per language breakdown results are in Table~\ref{tab:result-ser-few-shot-per-lang-1} and~\ref{tab:result-ser-few-shot-per-lang-2}. In this few-shot setup, the \textsc{Fine-tuned} model consistently improves upon the \textsc{Pretrained} model. However, the input-space generative DA methods (\textsc{Seamless M4T} and \textsc{Seamless Expressive}) degrade performance compared to both \textsc{Pretrained} and \textsc{Fine-tuned}. \textsc{Latent Filling} improves performance overall, demonstrating the benefit of performing DA in latent space. GeLDA achieves the best performance across all backbone FMs, indicating the strong generalization of our method. Specifically, averaged over the 12 languages, GeLDA improves UA by 4.08\%, 4.91\%, and 2.94\% for \textsc{WavLM-large}, \textsc{emotion2vec-base}, and \textsc{Whisper-large}, respectively. Overall, these results show that GeLDA is effective for building SER models for low-resource target languages in highly imbalanced real-world settings, highlighting its broader impact on practical SER applications.

\subsection{Per Language Breakdown of Few-shot SER Experiments}

\begin{table*}[ht]
\caption{Per language breakdown of few-shot SER experiments with \textsc{Whisper-large} as a backbone FM.}
    \label{tab:result-ser-few-shot-per-lang-1}
    \centering
    \footnotesize
    \resizebox{\textwidth}{!}{
    \begin{tabular}{lcccccccccc|ccc}

    \toprule
        \multirow{2}{*}{\textbf{Model}} & \textbf{Pretrain} & \textbf{Fine-tune} & \multirow{2}{*}{\textbf{Aug. Space}} & \multicolumn{7}{c}{\textbf{Per emotion recall (\%)}} & 
        \multirow{2}{*}{\textbf{UA (\%)}} & \multirow{2}{*}{\textbf{WA (\%)}} & 
        \multirow{2}{*}{\textbf{Macro-F1 (\%)}}\\ 
        \cline{5-11}
        ~ & \textbf{Dataset} & \textbf{Dataset} & ~ &\textbf{Angry} & \textbf{Disgust} & \textbf{Fear} & \textbf{Happy} & \textbf{Neutral} & \textbf{Sad} & \multicolumn{1}{c}{\textbf{Surprise}} & ~ & ~ & ~\\
        \midrule
        \hline
\rowcolor{gray!20}\multicolumn{14}{c}{Language: Amharic} \\
Pretrained & $\mathcal{D}$ & -- & -- & 98.00 & -- & 93.00 & 97.00 & 95.00 & 93.00 & -- & 95.46 & 95.45 & 95.44 \\
Fine-tuned & $\mathcal{D}$ & $\mathcal{D}^{(\kappa)}$ & -- & 98.35 & -- & 95.54 & 97.25 & 96.41 & 95.44 & -- & 96.60 & 96.59 & 96.57 \\
\hline
~ & \multirow{3}{*}{$\mathcal{D}$} & \multirow{3}{*}{$\mathcal{D}^{(\kappa)} \bigcup \bar{\mathcal{D}}^{(\kappa)}$} & $\mathcal{Z}^{(0)}$ & 98.62 & -- & 95.01 & 97.80 & 97.18 & 94.02 & -- & 96.53 & 96.54 & 96.51 \\
Latent Filling & ~ & ~ & $\mathcal{Z}^{(1)}$ & 98.35 & -- & 94.75 & 98.90 & 96.67 & 93.73 & -- & 96.48 & 96.48 & 96.46 \\
~ &   ~ & ~ & $\mathcal{Z}^{(2)}$ & 97.80 & -- & 93.18 & 97.80 & 96.92 & 94.30 & -- & 96.00 & 95.99 & 95.97 \\
\hline
~ & \multirow{3}{*}{$\mathcal{D}$} & \multirow{3}{*}{$\mathcal{D}^{(\kappa)} \bigcup \bar{\mathcal{D}}^{(\kappa)}$} & $\mathcal{Z}^{(0)}$ & 98.90 & -- & 95.01 & 99.17 & 96.67 & 95.16 & -- & 96.98 & 96.97 & 96.96 \\
GeLDA (Ours) & ~ & ~ & $\mathcal{Z}^{(1)}$ & 98.35 & -- & 94.23 & 98.35 & 96.67 & 94.02 & -- & 96.32 & 96.32 & 96.30 \\
~ & ~ & ~ & $\mathcal{Z}^{(2)}$ & 98.62 & -- & 93.18 & 97.80 & 94.87 & 95.16 & -- & 95.92 & 95.89 & 95.88 \\
\hline
\rowcolor{gray!20}\multicolumn{14}{c}{Language: Bangla} \\
Pretrained & $\mathcal{D}$ & -- & -- & 76.00 & 60.00 & 73.00 & 78.00 & 90.00 & 78.00 & 76.00 & 75.81 & 75.81 & 75.79 \\
Fine-tuned & $\mathcal{D}$ & $\mathcal{D}^{(\kappa)}$ & -- & 76.83 & 60.00 & 72.17 & 77.83 & 89.50 & 76.50 & 75.67 & 75.50 & 75.50 & 75.44 \\
\hline
~ & \multirow{3}{*}{$\mathcal{D}$} & \multirow{3}{*}{$\mathcal{D}^{(\kappa)} \bigcup \bar{\mathcal{D}}^{(\kappa)}$} & $\mathcal{Z}^{(0)}$ & 73.33 & 63.67 & 63.17 & 75.50 & 87.83 & 77.67 & 75.67 & 73.84 & 73.84 & 74.08 \\
Latent Filling & ~ & ~ & $\mathcal{Z}^{(1)}$ & 78.33 & 59.17 & 71.83 & 78.67 & 91.33 & 76.67 & 74.33 & 75.76 & 75.76 & 75.68 \\
~ &   ~ & ~ & $\mathcal{Z}^{(2)}$ & 76.00 & 60.00 & 71.83 & 76.67 & 89.83 & 78.50 & 75.33 & 75.45 & 75.45 & 75.46 \\
\hline
~ & \multirow{3}{*}{$\mathcal{D}$} & \multirow{3}{*}{$\mathcal{D}^{(\kappa)} \bigcup \bar{\mathcal{D}}^{(\kappa)}$} & $\mathcal{Z}^{(0)}$ & 79.83 & 61.50 & 73.50 & 80.67 & 90.83 & 77.00 & 73.83 & 76.74 & 76.74 & 76.66 \\
GeLDA (Ours) & ~ & ~ & $\mathcal{Z}^{(1)}$ & 76.33 & 58.83 & 72.67 & 79.17 & 89.33 & 78.00 & 74.83 & 75.60 & 75.60 & 75.55 \\
~ & ~ & ~ & $\mathcal{Z}^{(2)}$ & 75.67 & 61.17 & 72.50 & 75.83 & 89.00 & 76.50 & 75.50 & 75.17 & 75.17 & 75.18 \\
\hline
\rowcolor{gray!20}\multicolumn{14}{c}{Language: French} \\
Pretrained & $\mathcal{D}$ & -- & -- & 82.00 & 67.00 & 74.00 & 75.00 & 82.00 & 70.00 & 70.00 & 74.29 & 74.00 & 73.80 \\
Fine-tuned & $\mathcal{D}$ & $\mathcal{D}^{(\kappa)}$ & -- & 83.13 & 71.05 & 71.21 & 74.12 & 85.71 & 71.31 & 69.96 & 75.21 & 74.81 & 74.57 \\
\hline
Seamless M4T & \multirow{2}{*}{$\mathcal{D}$} & \multirow{2}{*}{$\mathcal{D}^{(\kappa)} \bigcup \bar{\mathcal{D}}^{(\kappa)}$} & $\mathcal{X}$ & 73.09 & 69.30 & 62.12 & 76.47 & 80.36 & 72.15 & 73.63 & 72.45 & 72.45 & 72.16 \\
Seamless Expressive & ~ & ~ & $\mathcal{X}$ & 81.53 & 65.79 & 66.16 & 71.76 & 79.17 & 73.00 & 73.26 & 72.95 & 72.95 & 72.66 \\
\hline
~ & \multirow{3}{*}{$\mathcal{D}$} & \multirow{3}{*}{$\mathcal{D}^{(\kappa)} \bigcup \bar{\mathcal{D}}^{(\kappa)}$} & $\mathcal{Z}^{(0)}$ & 83.94 & 73.68 & 66.16 & 73.73 & 86.31 & 69.62 & 73.26 & 75.24 & 75.00 & 74.64 \\
Latent Filling & ~ & ~ & $\mathcal{Z}^{(1)}$ & 79.52 & 71.49 & 73.74 & 72.94 & 84.52 & 68.35 & 76.19 & 75.25 & 74.94 & 74.74 \\
~ &   ~ & ~ & $\mathcal{Z}^{(2)}$ & 81.53 & 67.98 & 71.21 & 74.51 & 84.52 & 70.89 & 70.70 & 74.48 & 74.13 & 73.90 \\
\hline
~ & \multirow{3}{*}{$\mathcal{D}$} & \multirow{3}{*}{$\mathcal{D}^{(\kappa)} \bigcup \bar{\mathcal{D}}^{(\kappa)}$} & $\mathcal{Z}^{(0)}$ & 85.54 & 72.37 & 70.20 & 72.55 & 90.48 & 70.04 & 71.79 & 76.14 & 75.62 & 75.33 \\
GeLDA (Ours) & ~ & ~ & $\mathcal{Z}^{(1)}$ & 76.31 & 66.23 & 66.67 & 72.55 & 91.67 & 65.40 & 69.60 & 72.63 & 71.95 & 71.73 \\
~ & ~ & ~ & $\mathcal{Z}^{(2)}$ & 81.12 & 68.86 & 68.18 & 72.55 & 81.55 & 68.78 & 69.60 & 72.95 & 72.70 & 72.40 \\
\hline
\rowcolor{gray!20}\multicolumn{14}{c}{Language: German} \\
Pretrained & $\mathcal{D}$ & -- & -- & 99.00 & 73.00 & 86.00 & 93.00 & 98.00 & 97.00 & -- & 91.10 & 92.73 & 91.79 \\
Fine-tuned & $\mathcal{D}$ & $\mathcal{D}^{(\kappa)}$ & -- & 98.96 & 80.30 & 88.54 & 95.56 & 97.78 & 100.00 & -- & 93.52 & 94.56 & 93.91 \\
\hline
Seamless M4T & \multirow{2}{*}{$\mathcal{D}$} & \multirow{2}{*}{$\mathcal{D}^{(\kappa)} \bigcup \bar{\mathcal{D}}^{(\kappa)}$} & $\mathcal{X}$ & 97.62 & 73.11 & 76.27 & 90.77 & 95.83 & 100.00 & -- & 88.93 & 90.26 & 88.66 \\
Seamless Expressive & ~ & ~ & $\mathcal{X}$ & 98.96 & 76.14 & 88.20 & 85.98 & 95.56 & 97.44 & -- & 90.38 & 91.98 & 91.17 \\
\hline
~ & \multirow{3}{*}{$\mathcal{D}$} & \multirow{3}{*}{$\mathcal{D}^{(\kappa)} \bigcup \bar{\mathcal{D}}^{(\kappa)}$} & $\mathcal{Z}^{(0)}$ & 100.00 & 76.14 & 85.97 & 97.78 & 97.78 & 100.00 & -- & 92.94 & 94.56 & 93.60 \\
Latent Filling & ~ & ~ & $\mathcal{Z}^{(1)}$ & 98.96 & 76.14 & 87.73 & 95.56 & 97.78 & 100.00 & -- & 92.69 & 94.16 & 93.25 \\
~ &   ~ & ~ & $\mathcal{Z}^{(2)}$ & 98.96 & 76.14 & 85.97 & 95.56 & 97.78 & 97.44 & -- & 91.97 & 93.42 & 92.58 \\
\hline
~ & \multirow{3}{*}{$\mathcal{D}$} & \multirow{3}{*}{$\mathcal{D}^{(\kappa)} \bigcup \bar{\mathcal{D}}^{(\kappa)}$} & $\mathcal{Z}^{(0)}$ & 100.00 & 80.30 & 97.78 & 97.78 & 100.00 & 100.00 & -- & 95.98 & 97.38 & 96.26 \\
GeLDA (Ours) & ~ & ~ & $\mathcal{Z}^{(1)}$ & 97.62 & 80.30 & 93.46 & 95.56 & 100.00 & 100.00 & -- & 94.49 & 95.69 & 94.84 \\
~ & ~ & ~ & $\mathcal{Z}^{(2)}$ & 98.96 & 79.17 & 88.20 & 97.78 & 95.69 & 100.00 & -- & 93.30 & 94.56 & 93.77 \\
\hline
\rowcolor{gray!20}\multicolumn{14}{c}{Language: Greek} \\
Pretrained & $\mathcal{D}$ & -- & -- & 90.00 & 72.00 & 91.00 & 85.00 & -- & 88.00 & -- & 85.23 & 85.23 & 85.33 \\
Fine-tuned & $\mathcal{D}$ & $\mathcal{D}^{(\kappa)}$ & -- & 90.00 & 75.56 & 90.00 & 80.46 & -- & 88.89 & -- & 84.98 & 85.01 & 85.07 \\
\hline
~ & \multirow{3}{*}{$\mathcal{D}$} & \multirow{3}{*}{$\mathcal{D}^{(\kappa)} \bigcup \bar{\mathcal{D}}^{(\kappa)}$} & $\mathcal{Z}^{(0)}$ & 93.33 & 74.44 & 90.00 & 85.06 & -- & 85.56 & -- & 85.68 & 85.68 & 85.61 \\
Latent Filling & ~ & ~ & $\mathcal{Z}^{(1)}$ & 90.00 & 74.44 & 95.56 & 78.16 & -- & 88.89 & -- & 85.41 & 85.46 & 85.47 \\
~ &   ~ & ~ & $\mathcal{Z}^{(2)}$ & 90.00 & 71.11 & 93.33 & 83.91 & -- & 87.78 & -- & 85.22 & 85.23 & 85.18 \\
\hline
~ & \multirow{3}{*}{$\mathcal{D}$} & \multirow{3}{*}{$\mathcal{D}^{(\kappa)} \bigcup \bar{\mathcal{D}}^{(\kappa)}$} & $\mathcal{Z}^{(0)}$ & 91.11 & 81.11 & 92.22 & 88.51 & -- & 92.22 & -- & 89.03 & 89.04 & 89.04 \\
GeLDA (Ours) & ~ & ~ & $\mathcal{Z}^{(1)}$ & 83.33 & 77.78 & 97.78 & 90.80 & -- & 88.89 & -- & 87.72 & 87.70 & 87.63 \\
~ & ~ & ~ & $\mathcal{Z}^{(2)}$ & 90.00 & 72.22 & 90.00 & 87.36 & -- & 84.44 & -- & 84.81 & 84.79 & 84.84 \\

\bottomrule
    \end{tabular}
}
\end{table*}

\begin{table*}[ht]
\caption{Per language breakdown of few-shot SER experiments with \textsc{Whisper-large} as a backbone FM.}
    \label{tab:result-ser-few-shot-per-lang-2}
    \centering
    \footnotesize
    \resizebox{\textwidth}{!}{
    \begin{tabular}{lcccccccccc|ccc}
    \toprule
        \multirow{2}{*}{\textbf{Model}} & \textbf{Pretrain} & \textbf{Fine-tune} & \multirow{2}{*}{\textbf{Aug. Space}} & \multicolumn{7}{c}{\textbf{Per emotion recall (\%)}} & 
        \multirow{2}{*}{\textbf{UA (\%)}} & \multirow{2}{*}{\textbf{WA (\%)}} & 
        \multirow{2}{*}{\textbf{Macro-F1 (\%)}}\\ 
        \cline{5-11}
        ~ & \textbf{Dataset} & \textbf{Dataset} & ~ &\textbf{Angry} & \textbf{Disgust} & \textbf{Fear} & \textbf{Happy} & \textbf{Neutral} & \textbf{Sad} & \multicolumn{1}{c}{\textbf{Surprise}} & ~ & ~ & ~\\
        \midrule
        \hline
\rowcolor{gray!20}\multicolumn{14}{c}{Language: Italian} \\
Pretrained & $\mathcal{D}$ & -- & -- & 78.00 & 73.00 & 72.00 & 76.00 & 87.00 & 75.00 & 71.00 & 75.88 & 75.88 & 75.84 \\
Fine-tuned & $\mathcal{D}$ & $\mathcal{D}^{(\kappa)}$ & -- & 78.08 & 74.49 & 71.03 & 76.41 & 87.95 & 75.77 & 71.67 & 76.48 & 76.48 & 76.45 \\
\hline
Seamless M4T & \multirow{2}{*}{$\mathcal{D}$} & \multirow{2}{*}{$\mathcal{D}^{(\kappa)} \bigcup \bar{\mathcal{D}}^{(\kappa)}$} & $\mathcal{X}$ & 79.23 & 75.51 & 75.64 & 74.74 & 87.82 & 71.54 & 66.92 & 75.91 & 75.91 & 75.84 \\
Seamless Expressive & ~ & ~ & $\mathcal{X}$ & 77.56 & 73.97 & 72.31 & 73.97 & 86.67 & 73.97 & 71.03 & 75.64 & 75.64 & 75.62 \\
\hline
~ & \multirow{3}{*}{$\mathcal{D}$} & \multirow{3}{*}{$\mathcal{D}^{(\kappa)} \bigcup \bar{\mathcal{D}}^{(\kappa)}$} & $\mathcal{Z}^{(0)}$ & 79.36 & 73.46 & 73.72 & 74.87 & 88.72 & 77.18 & 72.56 & 77.13 & 77.13 & 77.08 \\
Latent Filling & ~ & ~ & $\mathcal{Z}^{(1)}$ & 78.21 & 71.15 & 73.85 & 76.03 & 86.41 & 73.33 & 69.62 & 75.51 & 75.51 & 75.48 \\
~ &   ~ & ~ & $\mathcal{Z}^{(2)}$ & 76.92 & 72.95 & 72.31 & 74.87 & 87.05 & 75.51 & 71.79 & 75.92 & 75.92 & 75.90 \\
\hline
~ & \multirow{3}{*}{$\mathcal{D}$} & \multirow{3}{*}{$\mathcal{D}^{(\kappa)} \bigcup \bar{\mathcal{D}}^{(\kappa)}$} & $\mathcal{Z}^{(0)}$ & 80.26 & 73.33 & 73.97 & 77.31 & 87.82 & 76.92 & 70.64 & 77.18 & 77.18 & 77.14 \\
GeLDA (Ours) & ~ & ~ & $\mathcal{Z}^{(1)}$ & 77.31 & 72.56 & 70.38 & 76.92 & 85.64 & 73.97 & 67.95 & 74.96 & 74.96 & 74.94 \\
~ & ~ & ~ & $\mathcal{Z}^{(2)}$ & 77.82 & 71.54 & 70.00 & 73.21 & 84.74 & 73.72 & 70.51 & 74.51 & 74.51 & 74.48 \\
        \hline
\rowcolor{gray!20}\multicolumn{14}{c}{Language: Persian} \\
Pretrained & $\mathcal{D}$ & -- & -- & 95.00 & -- & 57.00 & 88.00 & 92.00 & 80.00 & 70.00 & 80.22 & 89.17 & 79.10 \\
Fine-tuned & $\mathcal{D}$ & $\mathcal{D}^{(\kappa)}$ & -- & 94.32 & -- & 47.62 & 86.39 & 92.06 & 78.52 & 70.07 & 78.16 & 88.79 & 77.55 \\
\hline
~ & \multirow{3}{*}{$\mathcal{D}$} & \multirow{3}{*}{$\mathcal{D}^{(\kappa)} \bigcup \bar{\mathcal{D}}^{(\kappa)}$} & $\mathcal{Z}^{(0)}$ & 94.96 & -- & 33.33 & 87.76 & 92.99 & 80.37 & 72.79 & 77.03 & 89.74 & 77.43 \\
Latent Filling & ~ & ~ & $\mathcal{Z}^{(1)}$ & 94.57 & -- & 47.62 & 90.48 & 92.20 & 82.59 & 71.43 & 79.81 & 89.83 & 78.96 \\
~ &   ~ & ~ & $\mathcal{Z}^{(2)}$ & 94.70 & -- & 57.14 & 87.07 & 91.80 & 79.26 & 71.43 & 80.24 & 89.17 & 78.72 \\
\hline
~ & \multirow{3}{*}{$\mathcal{D}$} & \multirow{3}{*}{$\mathcal{D}^{(\kappa)} \bigcup \bar{\mathcal{D}}^{(\kappa)}$} & $\mathcal{Z}^{(0)}$ & 94.44 & -- & 76.19 & 89.12 & 89.81 & 82.22 & 72.79 & 84.10 & 89.17 & 79.06 \\
GeLDA (Ours) & ~ & ~ & $\mathcal{Z}^{(1)}$ & 92.51 & -- & 85.71 & 88.44 & 88.49 & 75.19 & 75.51 & 84.31 & 87.33 & 76.52 \\
~ & ~ & ~ & $\mathcal{Z}^{(2)}$ & 93.15 & -- & 57.14 & 85.03 & 89.15 & 78.15 & 74.83 & 79.58 & 87.61 & 75.86 \\
        \hline
\rowcolor{gray!20}\multicolumn{14}{c}{Language: Polish} \\
Pretrained & $\mathcal{D}$ & -- & -- & 89.00 & -- & 35.00 & -- & 51.00 & -- & -- & 58.52 & 58.52 & 52.07 \\
Fine-tuned & $\mathcal{D}$ & $\mathcal{D}^{(\kappa)}$ & -- & 88.89 & -- & 34.44 & -- & 51.11 & -- & -- & 58.15 & 58.15 & 51.80 \\
\hline
~ & \multirow{3}{*}{$\mathcal{D}$} & \multirow{3}{*}{$\mathcal{D}^{(\kappa)} \bigcup \bar{\mathcal{D}}^{(\kappa)}$} & $\mathcal{Z}^{(0)}$ & 93.33 & -- & 34.44 & -- & 54.44 & -- & -- & 60.74 & 60.74 & 53.56 \\
Latent Filling & ~ & ~ & $\mathcal{Z}^{(1)}$ & 92.22 & -- & 33.33 & -- & 55.56 & -- & -- & 60.37 & 60.37 & 54.24 \\
~ &   ~ & ~ & $\mathcal{Z}^{(2)}$ & 90.00 & -- & 35.56 & -- & 51.11 & -- & -- & 58.89 & 58.89 & 52.46 \\
\hline
~ & \multirow{3}{*}{$\mathcal{D}$} & \multirow{3}{*}{$\mathcal{D}^{(\kappa)} \bigcup \bar{\mathcal{D}}^{(\kappa)}$} & $\mathcal{Z}^{(0)}$ & 91.11 & -- & 34.44 & -- & 55.56 & -- & -- & 60.37 & 60.37 & 53.76 \\
GeLDA (Ours) & ~ & ~ & $\mathcal{Z}^{(1)}$ & 87.78 & -- & 38.89 & -- & 44.44 & -- & -- & 57.04 & 57.04 & 50.51 \\
~ & ~ & ~ & $\mathcal{Z}^{(2)}$ & 94.44 & -- & 32.22 & -- & 54.44 & -- & -- & 60.37 & 60.37 & 54.14 \\
\hline
\rowcolor{gray!20}\multicolumn{14}{c}{Language: Russian} \\
Pretrained & $\mathcal{D}$ & -- & -- & 78.00 & 47.00 & 50.00 & 59.00 & 63.00 & 41.00 & -- & 56.08 & 56.95 & 56.17 \\
Fine-tuned & $\mathcal{D}$ & $\mathcal{D}^{(\kappa)}$ & -- & 77.27 & 48.65 & 51.85 & 56.82 & 59.65 & 40.62 & -- & 55.81 & 56.66 & 55.88 \\
\hline
~ & \multirow{3}{*}{$\mathcal{D}$} & \multirow{3}{*}{$\mathcal{D}^{(\kappa)} \bigcup \bar{\mathcal{D}}^{(\kappa)}$} & $\mathcal{Z}^{(0)}$ & 78.03 & 48.65 & 56.30 & 59.09 & 57.89 & 38.54 & -- & 56.42 & 57.50 & 56.03 \\
Latent Filling & ~ & ~ & $\mathcal{Z}^{(1)}$ & 76.52 & 54.95 & 57.78 & 55.30 & 57.89 & 31.25 & -- & 55.62 & 56.81 & 55.42 \\
~ &   ~ & ~ & $\mathcal{Z}^{(2)}$ & 79.55 & 46.85 & 54.07 & 58.33 & 60.53 & 42.71 & -- & 57.01 & 57.92 & 56.89 \\
\hline
~ & \multirow{3}{*}{$\mathcal{D}$} & \multirow{3}{*}{$\mathcal{D}^{(\kappa)} \bigcup \bar{\mathcal{D}}^{(\kappa)}$} & $\mathcal{Z}^{(0)}$ & 74.24 & 58.56 & 59.26 & 58.33 & 57.89 & 58.33 & -- & 61.10 & 61.39 & 61.20 \\
GeLDA (Ours) & ~ & ~ & $\mathcal{Z}^{(1)}$ & 71.21 & 46.85 & 59.26 & 62.88 & 57.02 & 45.83 & -- & 57.17 & 58.06 & 57.22 \\
~ & ~ & ~ & $\mathcal{Z}^{(2)}$ & 76.52 & 47.75 & 52.59 & 52.27 & 57.89 & 41.67 & -- & 54.78 & 55.56 & 54.54 \\
\hline
\rowcolor{gray!20}\multicolumn{14}{c}{Language: Spanish} \\
Pretrained & $\mathcal{D}$ & -- & -- & 81.00 & 72.00 & 48.00 & 58.00 & 87.00 & 56.00 & -- & 66.98 & 66.82 & 67.29 \\
Fine-tuned & $\mathcal{D}$ & $\mathcal{D}^{(\kappa)}$ & -- & 84.76 & 73.15 & 47.22 & 61.11 & 88.57 & 59.26 & -- & 69.01 & 68.85 & 69.10 \\
\hline
Seamless M4T & \multirow{2}{*}{$\mathcal{D}$} & \multirow{2}{*}{$\mathcal{D}^{(\kappa)} \bigcup \bar{\mathcal{D}}^{(\kappa)}$} & $\mathcal{X}$ & 78.10 & 75.00 & 39.81 & 63.89 & 85.71 & 61.11 & -- & 67.27 & 67.13 & 66.93 \\
Seamless Expressive & ~ & ~ & $\mathcal{X}$ & 79.05 & 75.93 & 41.67 & 69.44 & 88.57 & 60.19 & -- & 69.14 & 69.00 & 68.58 \\
\hline
~ & \multirow{3}{*}{$\mathcal{D}$} & \multirow{3}{*}{$\mathcal{D}^{(\kappa)} \bigcup \bar{\mathcal{D}}^{(\kappa)}$} & $\mathcal{Z}^{(0)}$ & 80.00 & 75.00 & 49.07 & 74.07 & 88.57 & 62.96 & -- & 71.61 & 71.50 & 71.57 \\
Latent Filling & ~ & ~ & $\mathcal{Z}^{(1)}$ & 83.81 & 71.30 & 55.56 & 70.37 & 87.62 & 53.70 & -- & 70.39 & 70.25 & 70.36 \\
~ &   ~ & ~ & $\mathcal{Z}^{(2)}$ & 82.86 & 74.07 & 48.15 & 61.11 & 86.67 & 55.56 & -- & 68.07 & 67.91 & 68.08 \\
\hline
~ & \multirow{3}{*}{$\mathcal{D}$} & \multirow{3}{*}{$\mathcal{D}^{(\kappa)} \bigcup \bar{\mathcal{D}}^{(\kappa)}$} & $\mathcal{Z}^{(0)}$ & 77.14 & 75.00 & 55.56 & 73.15 & 90.48 & 54.63 & -- & 70.99 & 70.87 & 71.00 \\
GeLDA (Ours) & ~ & ~ & $\mathcal{Z}^{(1)}$ & 81.90 & 72.22 & 50.00 & 77.78 & 84.76 & 57.41 & -- & 70.68 & 70.56 & 70.53 \\
~ & ~ & ~ & $\mathcal{Z}^{(2)}$ & 82.86 & 72.22 & 44.44 & 65.74 & 83.81 & 61.11 & -- & 68.36 & 68.22 & 68.10 \\
\hline
\rowcolor{gray!20}\multicolumn{14}{c}{Language: Turkish} \\
Pretrained & $\mathcal{D}$ & -- & -- & 91.00 & -- & -- & 91.00 & -- & 88.00 & -- & 89.96 & 89.90 & 91.36 \\
Fine-tuned & $\mathcal{D}$ & $\mathcal{D}^{(\kappa)}$ & -- & 91.74 & -- & -- & 91.01 & -- & 89.44 & -- & 90.73 & 90.71 & 91.65 \\
\hline
~ & \multirow{3}{*}{$\mathcal{D}$} & \multirow{3}{*}{$\mathcal{D}^{(\kappa)} \bigcup \bar{\mathcal{D}}^{(\kappa)}$} & $\mathcal{Z}^{(0)}$ & 94.21 & -- & -- & 95.51 & -- & 93.33 & -- & 94.35 & 94.24 & 94.22 \\
Latent Filling & ~ & ~ & $\mathcal{Z}^{(1)}$ & 93.94 & -- & -- & 91.39 & -- & 90.28 & -- & 91.87 & 91.92 & 92.53 \\
~ &   ~ & ~ & $\mathcal{Z}^{(2)}$ & 92.56 & -- & -- & 90.26 & -- & 91.39 & -- & 91.40 & 91.51 & 92.10 \\
\hline
~ & \multirow{3}{*}{$\mathcal{D}$} & \multirow{3}{*}{$\mathcal{D}^{(\kappa)} \bigcup \bar{\mathcal{D}}^{(\kappa)}$} & $\mathcal{Z}^{(0)}$ & 93.94 & -- & -- & 93.26 & -- & 93.33 & -- & 93.51 & 93.53 & 93.92 \\
GeLDA (Ours) & ~ & ~ & $\mathcal{Z}^{(1)}$ & 90.36 & -- & -- & 93.26 & -- & 93.89 & -- & 92.50 & 92.42 & 92.74 \\
~ & ~ & ~ & $\mathcal{Z}^{(2)}$ & 90.91 & -- & -- & 91.39 & -- & 91.11 & -- & 91.13 & 91.11 & 91.55 \\
\hline
\rowcolor{gray!20}\multicolumn{14}{c}{Language: Urdu} \\
Pretrained & $\mathcal{D}$ & -- & -- & 96.00 & -- & -- & 73.00 & 68.00 & 57.00 & -- & 73.67 & 73.67 & 73.58 \\
Fine-tuned & $\mathcal{D}$ & $\mathcal{D}^{(\kappa)}$ & -- & 97.33 & -- & -- & 73.33 & 60.00 & 62.67 & -- & 73.33 & 73.33 & 73.07 \\
\hline
~ & \multirow{3}{*}{$\mathcal{D}$} & \multirow{3}{*}{$\mathcal{D}^{(\kappa)} \bigcup \bar{\mathcal{D}}^{(\kappa)}$} & $\mathcal{Z}^{(0)}$ & 97.33 & -- & -- & 72.00 & 78.67 & 62.67 & -- & 77.67 & 77.67 & 77.47 \\
Latent Filling & ~ & ~ & $\mathcal{Z}^{(1)}$ & 94.67 & -- & -- & 73.33 & 58.67 & 58.67 & -- & 71.33 & 71.33 & 70.65 \\
~ &   ~ & ~ & $\mathcal{Z}^{(2)}$ & 97.33 & -- & -- & 72.00 & 60.00 & 60.00 & -- & 72.33 & 72.33 & 72.07 \\
\hline
~ & \multirow{3}{*}{$\mathcal{D}$} & \multirow{3}{*}{$\mathcal{D}^{(\kappa)} \bigcup \bar{\mathcal{D}}^{(\kappa)}$} & $\mathcal{Z}^{(0)}$ & 97.33 & -- & -- & 80.00 & 77.33 & 68.00 & -- & 80.67 & 80.67 & 80.91 \\
GeLDA (Ours) & ~ & ~ & $\mathcal{Z}^{(1)}$ & 97.33 & -- & -- & 69.33 & 77.33 & 68.00 & -- & 78.00 & 78.00 & 77.87 \\
~ & ~ & ~ & $\mathcal{Z}^{(2)}$ & 97.33 & -- & -- & 68.00 & 66.67 & 58.67 & -- & 72.67 & 72.67 & 72.32 \\
\bottomrule
    \end{tabular}
}
\end{table*}

\newpage
\clearpage
\section{Input-Space Image Generation}
\label{apdx:image_generation}
\begin{lstlisting}[caption={Prompt template used for text prompt generation.}, label={lst:prompt_generation}]
You are an expert prompt engineer for Stable Diffusion.
Given an image label, generate 100 diverse, high-quality prompts.
Each prompt must be visually descriptive and vary in art style,
camera angle, lighting, mood, era, medium, and environment.
Avoid repeating prompt structures.
All prompts must be written in English.

Image label: {label_name}
\end{lstlisting}

\begin{lstlisting}[caption={Prompt examples of image label "white stork."}, label={lst:prompt_examples}]
 A serene white stork standing on a sunlit marshland at sunrise.
 An ancient white stork perched gracefully on a crumbling Roman ruin.
 A detailed oil painting of a white stork against a vivid sunset sky.
 A white stork soaring through a dense forest canopy with golden leaves.
 A surreal scene of a white stork flying over a futuristic city skyline.
\end{lstlisting}

\begin{figure}[ht]
    \centering
    \begin{subfigure}{0.19\textwidth}
        \centering
        \includegraphics[width=\linewidth]{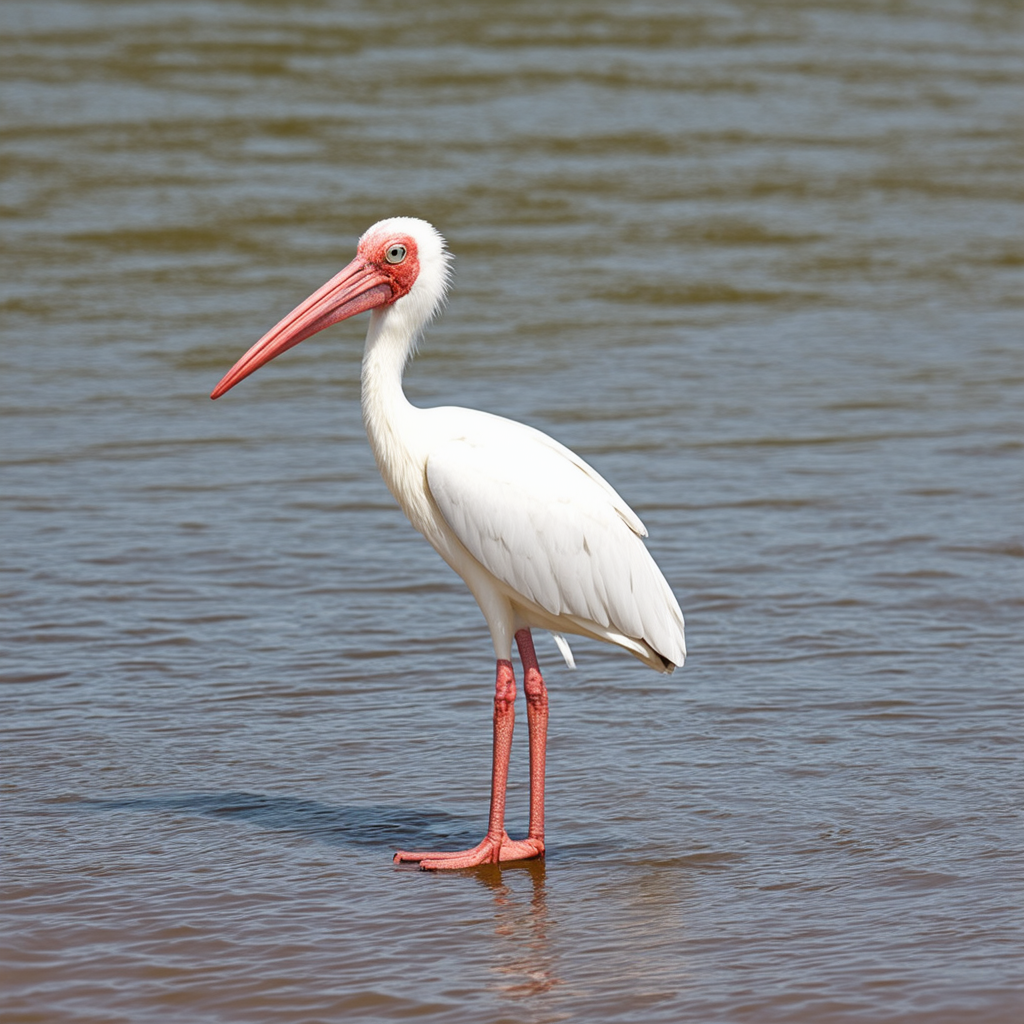}
    \end{subfigure}
    \begin{subfigure}{0.19\textwidth}
        \centering
        \includegraphics[width=\linewidth]{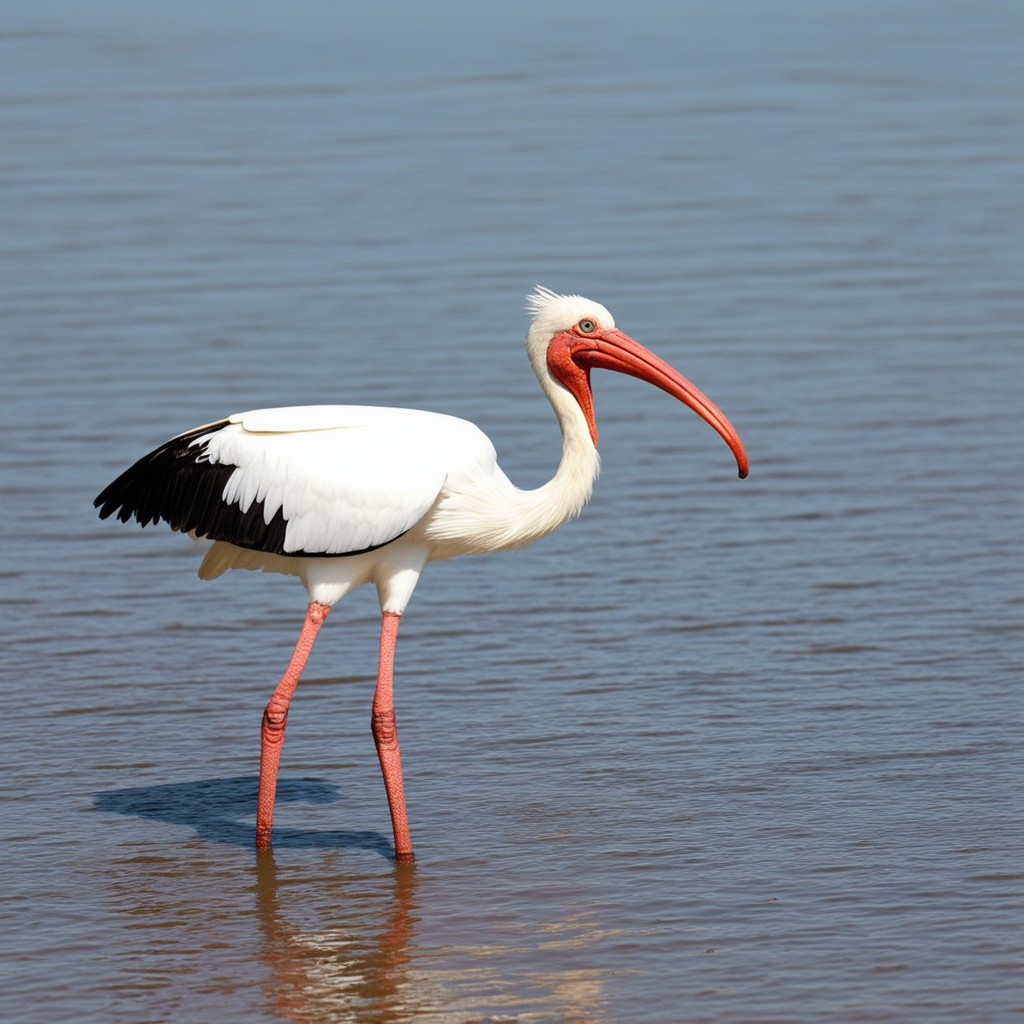}
    \end{subfigure}
    \begin{subfigure}{0.19\textwidth}
        \centering
        \includegraphics[width=\linewidth]{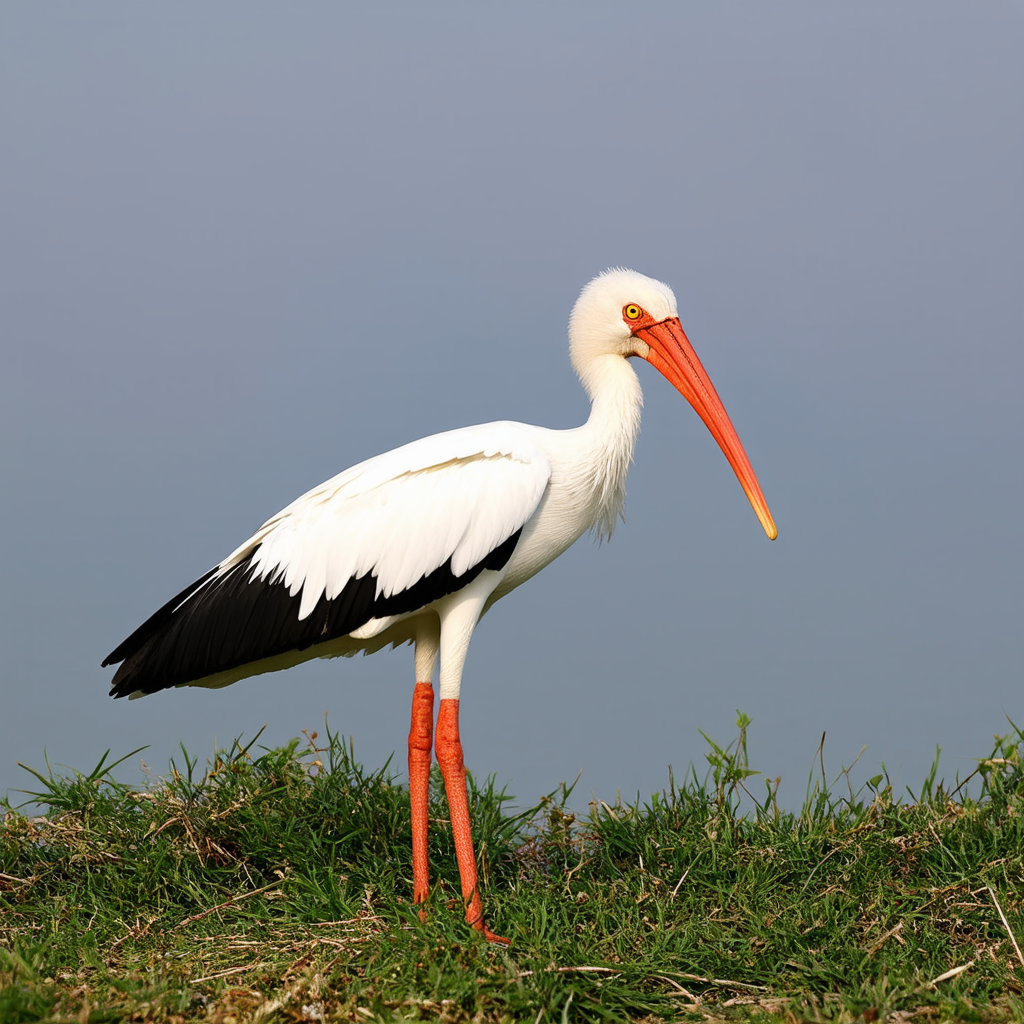}
    \end{subfigure}
    \begin{subfigure}{0.19\textwidth}
        \centering
        \includegraphics[width=\linewidth]{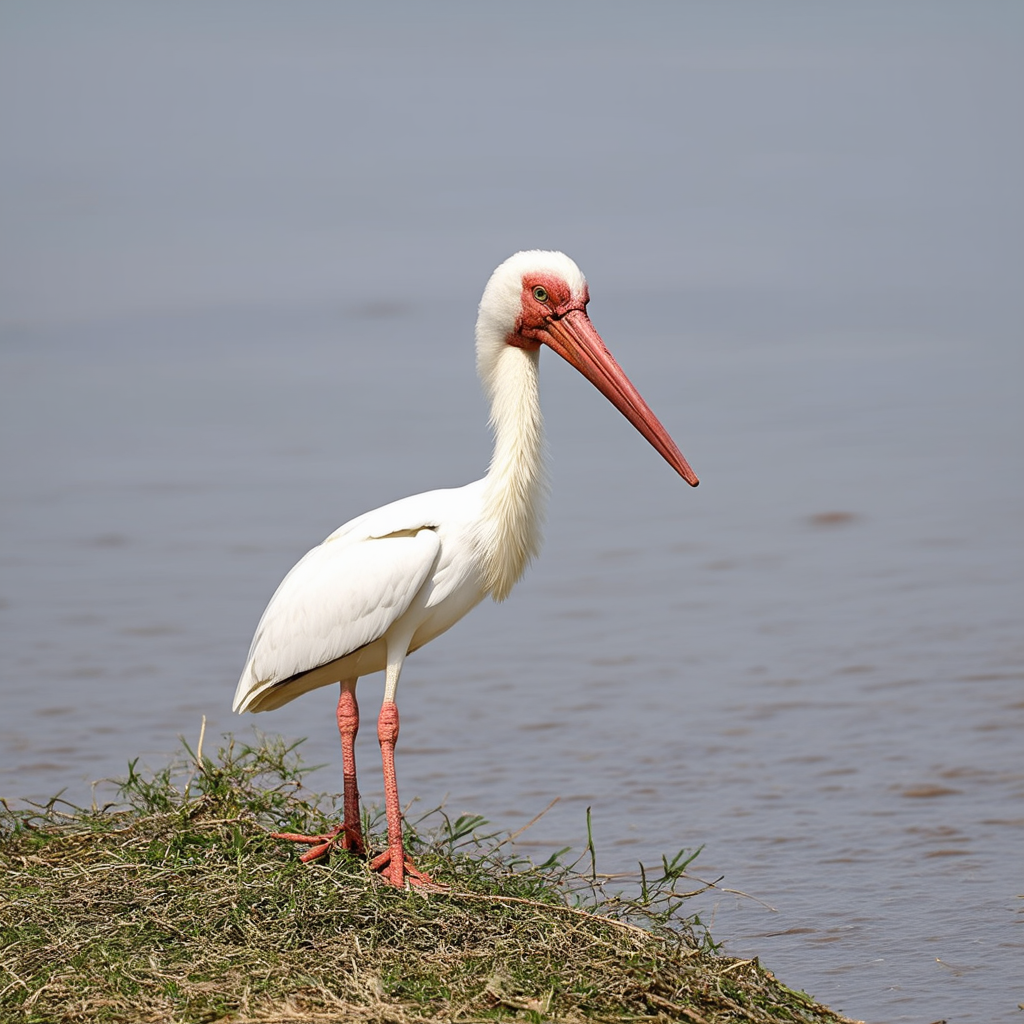}
    \end{subfigure}
    \begin{subfigure}{0.19\textwidth}
        \centering
        \includegraphics[width=\linewidth]{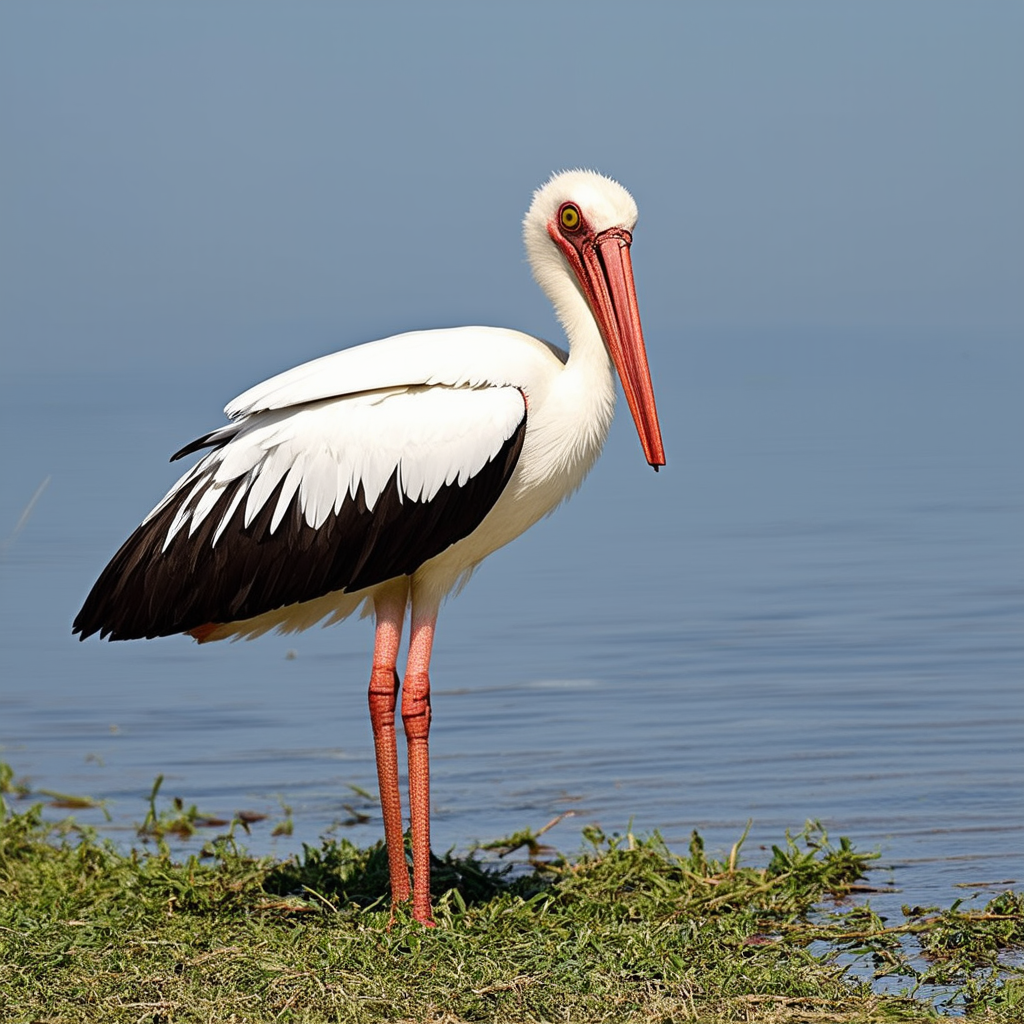}
    \end{subfigure}

    \caption{Examples of images generated from Stable Diffusion 3 with generated text prompts of ``A photo of a white stork''.}
    \label{fig:img_examples}
\end{figure}

\begin{figure}[ht]
    \centering
    \begin{subfigure}{0.19\textwidth}
        \centering
        \includegraphics[width=\linewidth]{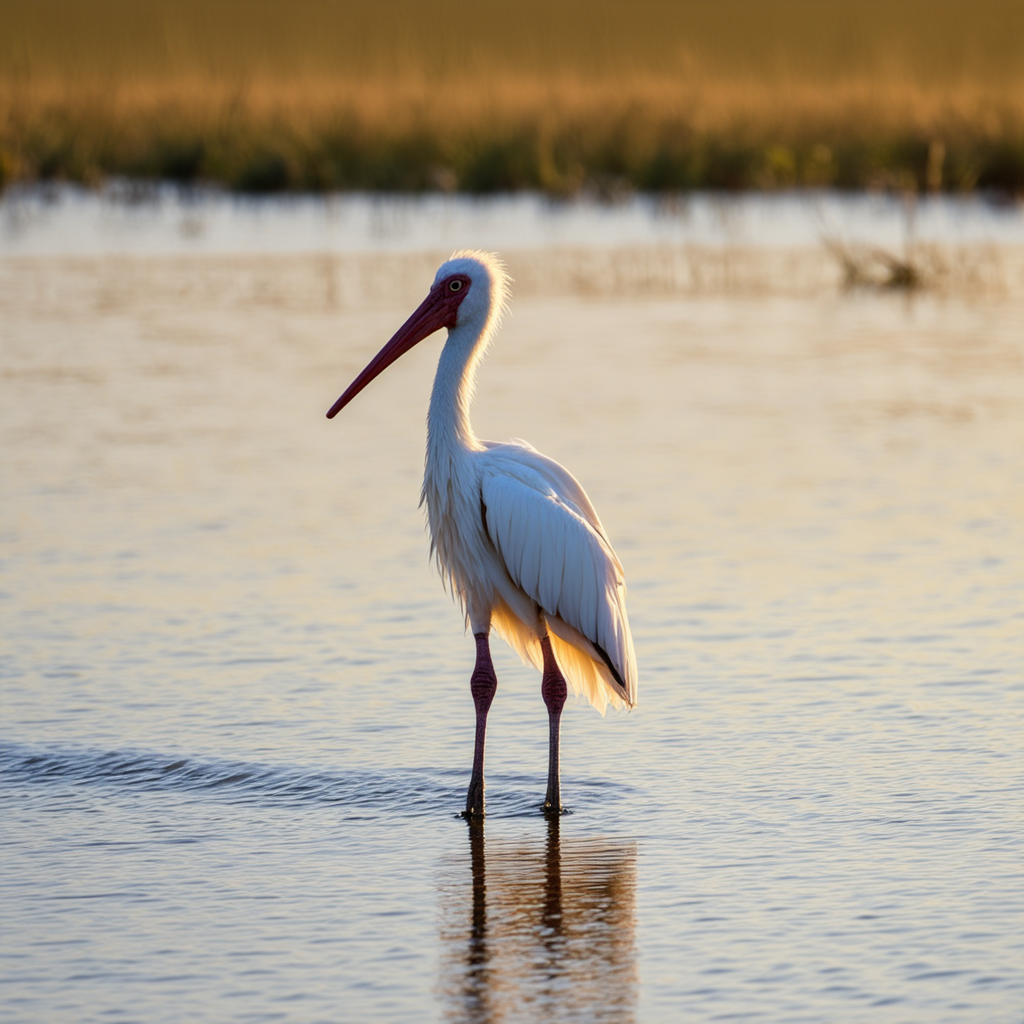}
    \end{subfigure}
    \begin{subfigure}{0.19\textwidth}
        \centering
        \includegraphics[width=\linewidth]{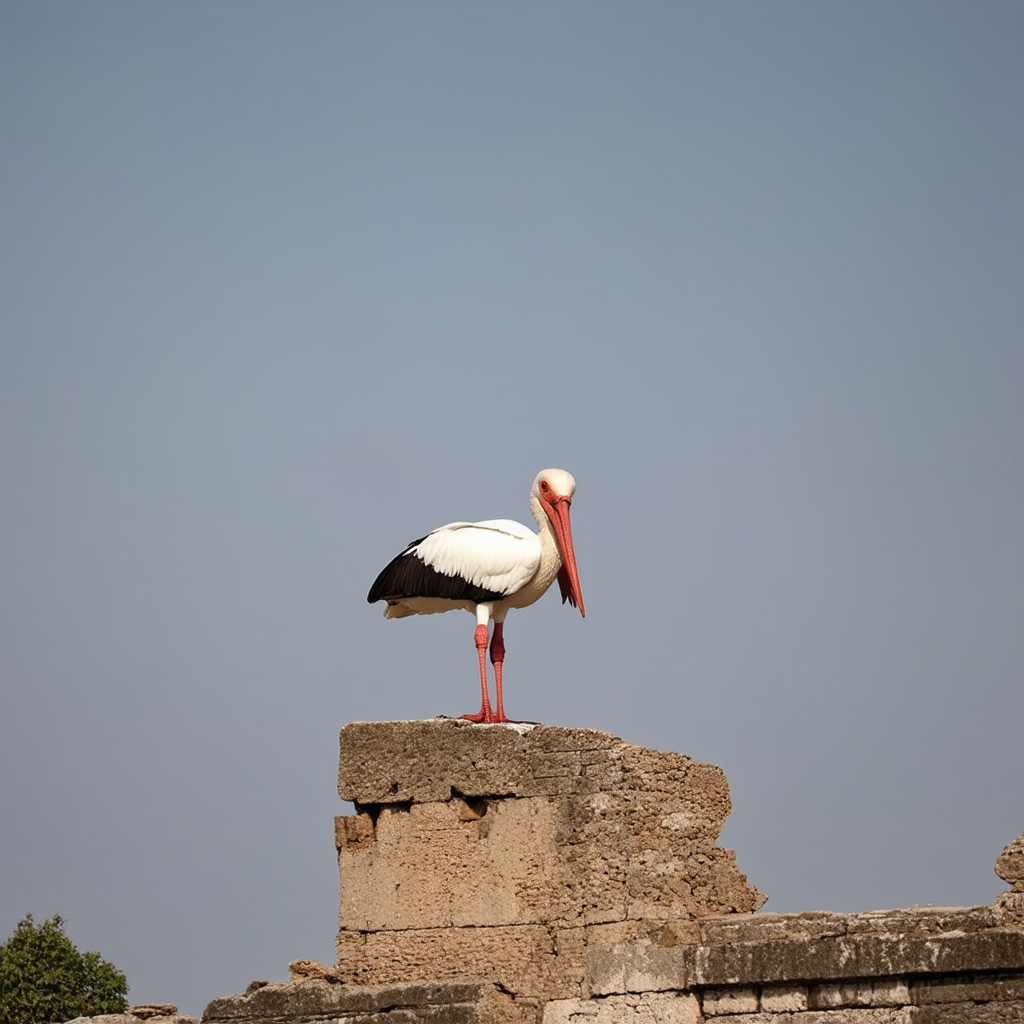}
    \end{subfigure}
    \begin{subfigure}{0.19\textwidth}
        \centering
        \includegraphics[width=\linewidth]{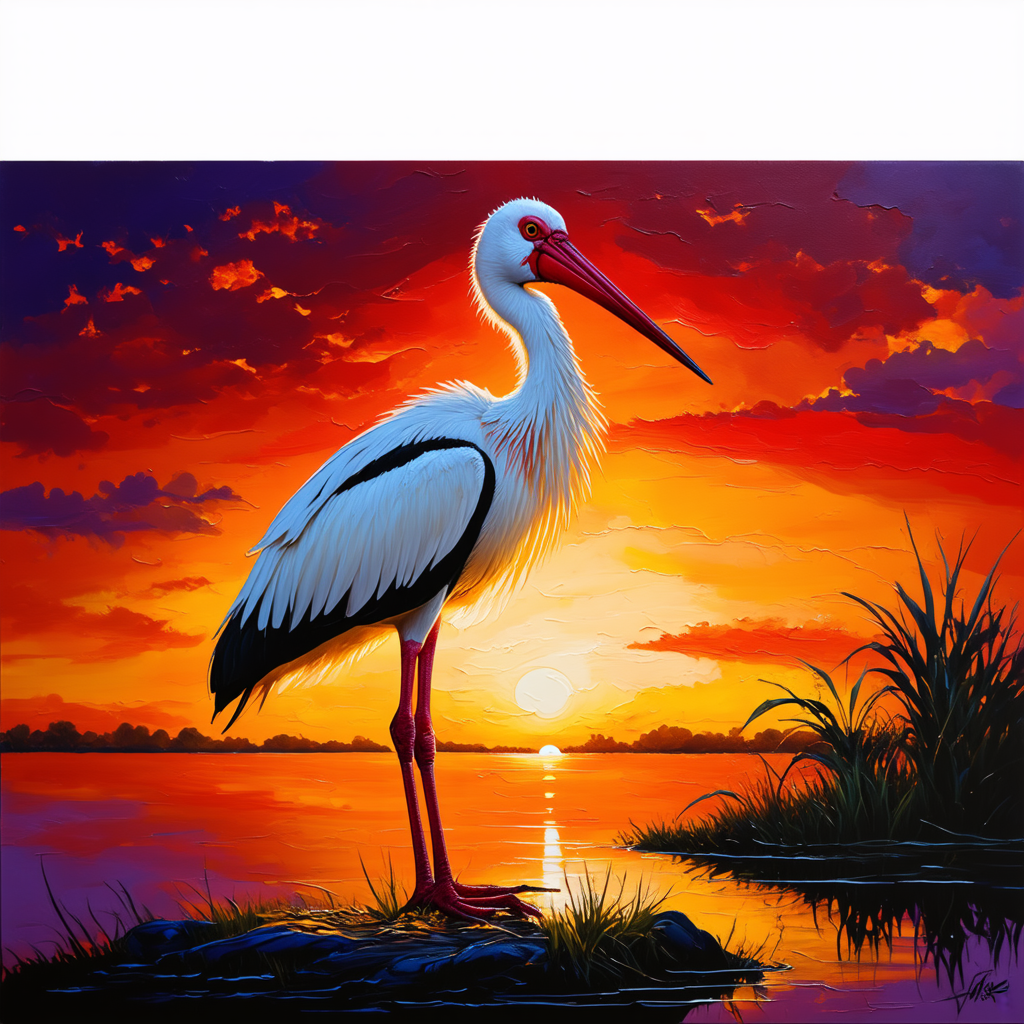}
    \end{subfigure}
    \begin{subfigure}{0.19\textwidth}
        \centering
        \includegraphics[width=\linewidth]{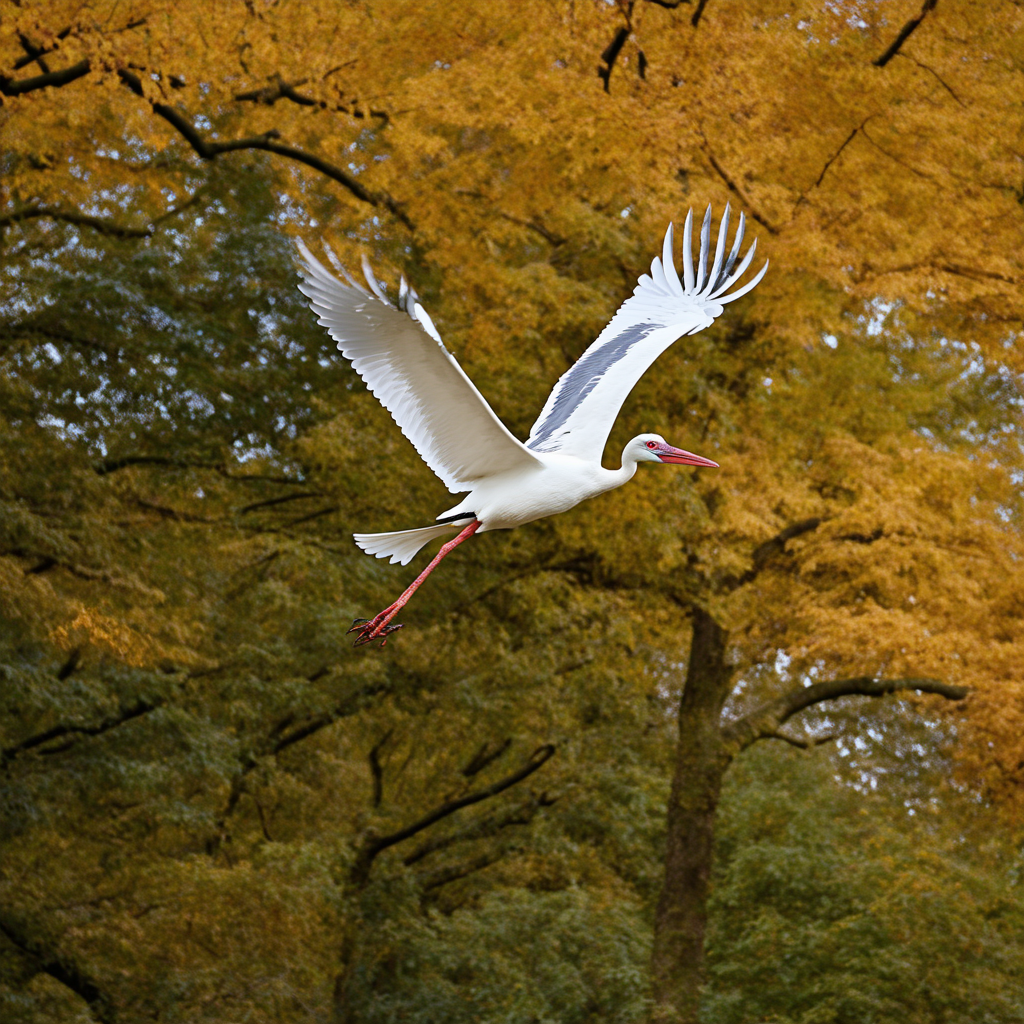}
    \end{subfigure}
    \begin{subfigure}{0.19\textwidth}
        \centering
        \includegraphics[width=\linewidth]{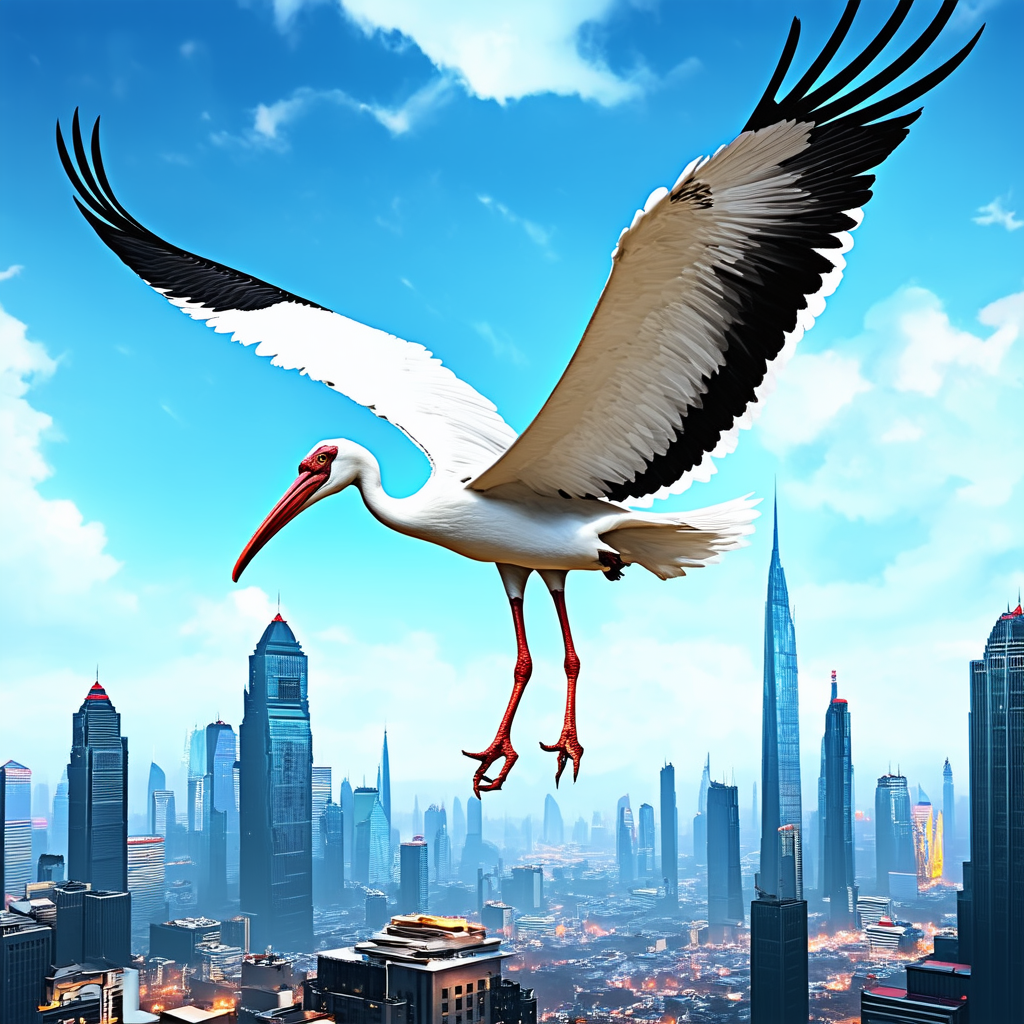}
    \end{subfigure}

    \caption{Examples of images generated from Stable Diffusion 3 with generated text prompts from Qwen2.5 with a label of ``white stork'' in Listing~\ref{lst:prompt_examples}.}
    \label{fig:img_examples_qwen}
\end{figure}

To generate a diverse set of images for each image label, we first construct a collection of text prompts that span a wide range of visual styles and attributes. Specifically, we use the Qwen~2.5~\cite{qwen2.5} large language model to generate 100 distinct text prompts per image label. These prompts are designed for input to the Stable Diffusion~3 image generation model~\cite{sd3}.

The prompt provided to Qwen~2.5 instructs the model to generate visually descriptive text prompts that vary across multiple dimensions, including art style, camera angle, lighting, mood, historical era, medium, and environment, while avoiding repetitive structures. The exact prompt template used for text prompt generation is shown in Listing~\ref{lst:prompt_generation}. Representative examples of the generated prompts are provided in Listing~\ref{lst:prompt_examples}, and the corresponding generated images are shown in Fig.~\ref{fig:img_examples_qwen}. Compared to the images in Fig.~\ref{fig:img_examples}, which are generated from a simple prompt (``A photo of a white stork''), the Qwen~2.5-generated prompts produce substantially more diverse images. This increased diversity is expected to be beneficial for training more robust image classifiers.




\end{document}